\documentclass[11pt]{article}
\usepackage[margin=2.5cm]{geometry}
\usepackage{amsmath,amssymb,amsthm,mathtools}
\usepackage{booktabs}
\usepackage{array}
\usepackage{longtable}
\usepackage{enumitem}
\usepackage{tikz}
\usepackage[hidelinks]{hyperref}
\DeclareMathOperator{\tr}{tr}
\DeclareMathOperator{\rank}{rank}
\DeclareMathOperator{\diag}{diag}
\DeclareMathOperator{\adj}{adj}
\DeclareMathOperator{\ReLU}{ReLU}
\DeclareMathOperator{\ch}{ch}
\newcommand{\R}{\mathbb{R}}
\newcommand{\op}{\mathrm{op}}
\newcommand{\FRR}{\mathrm{FRR}}
\newcommand{\Rec}{\mathrm{Rec}}
\newcommand{\Lost}{\mathrm{Lost}}
\newcommand{\Gain}{\mathrm{Gain}}
\newcommand{\deff}{d_{\mathrm{eff}}}
\newcommand{\Dtot}{D_{\mathrm{tot}}}
\newcommand{\Dt}{\tilde D}
\newcommand{\Dttot}{\tilde D_{\mathrm{tot}}}
\newcommand{\talign}{\tau_{\mathrm{align}}}
\newcommand{\cA}{\mathcal A}
\newcommand{\lsc}{\ell^{\mathrm{sc}}_\lambda}
\newcommand{\Fsc}{F_{\mathrm{sc}}}
\newcommand{\Lsc}{L_{\mathrm{sc}}}
\newcommand{\nphi}{\mathbf n_\phi}
\newcommand{\nphip}{\mathbf n_\phi^{\perp}}
\newcommand{\Waug}{W_{\mathrm{aug}}}
\newcommand{\Rb}{\bar R}
\newcommand{\qrad}{q_{\mathrm{rad}}}
\newcommand{\qtan}{q_{\mathrm{tan}}}
\newcommand{\qcpl}{q_{\mathrm{cpl}}}
\newcommand{\Pcpl}{\Pi_{\mathrm{cpl}}}
\newcommand{\arad}{\alpha_{\mathrm{rad}}}
\newcommand{\atan}{\alpha_{\mathrm{tan}}}
\newcommand{\acpl}{\alpha_{\mathrm{cpl}}}
\newcommand{\Numseg}{\mathrm{Num}_{\mathrm{seg}}}
\newcommand{\Denseg}{\mathrm{Den}_{\mathrm{seg}}}
\newcommand{\Numfam}{\mathrm{Num}_{\mathrm{fam}}}
\newcommand{\Denfam}{\mathrm{Den}_{\mathrm{fam}}}
\newcommand{\one}{\mathbf 1}

\newcommand{\mk}[1]{\fbox{\footnotesize\textsf{#1}}\,}
\newcommand{\MkDef}{\mk{definition (ours)}}
\newcommand{\MkDefS}[1]{\mk{definition (#1)}}
\newcommand{\MkOurs}{\mk{ours $\circ$}}

\newcommand{\MkOursChk}{\mk{ours $\checkmark$}}
\newcommand{\MkRepOursChk}{\mk{source (unreviewed) $\checkmark$}}
\newcommand{\MkRep}{\mk{source (unreviewed)}}

\newcommand{\MkTool}{\mk{tool}}
\newcommand{\MkConj}{\mk{conjecture (ours)}}
\newcommand{\MkSpec}{\mk{speculation}}
\newcommand{\MkMeas}{\mk{measured}}

\theoremstyle{definition}
\newtheorem{stmt}{Statement}[section]
\newtheorem{definition}[stmt]{Definition}
\newtheorem{theorem}[stmt]{Theorem}
\newtheorem{lemma}[stmt]{Lemma}
\newtheorem{proposition}[stmt]{Proposition}
\newtheorem{corollary}[stmt]{Corollary}
\newtheorem{remark}[stmt]{Remark}
\newtheorem{derivation}[stmt]{Derivation}
\newtheorem{empirical}[stmt]{Empirical finding}
\newtheorem{heuristic}[stmt]{Heuristic}
\newtheorem{conjecture}[stmt]{Conjecture}
\newtheorem{toolstmt}[stmt]{Tool}

\newenvironment{steps}{\begin{enumerate}[label=\textit{Step~\arabic*.},ref=\arabic*,leftmargin=*,itemsep=1pt,topsep=2pt]}{\end{enumerate}}

\title{When can an interpretability claim be trusted?\\ Stability and certification of feature dimensionality,\\ ridge leverage and the feature recovery rate}
\author{Ali Tajfar \qquad Behnam Tajfar\\[2pt] \normalsize Independent researchers}
\date{31 October 2026\\[2pt] Revised 11 November 2026: DOI corrected\\[4pt] Preprint v1.1\\[2pt] Licence: CC BY 4.0\\[2pt] DOI (all versions): \href{https://doi.org/10.5281/zenodo.23842568}{10.5281/zenodo.23842568}}

\begin{document}
\maketitle

\section*{Abstract}

Interpretability tools increasingly support safety claims: that a model contains a feature for a given concept, or that a sparse dictionary has recovered most of a model's features. Such claims rest on measurements with arbitrary thresholds. Small changes to the model weights, a different training seed, or a different threshold can reverse them. In earlier work in this program, a norm threshold alone produced a $24.8\%$ apparent failure of a well-known regularity. We ask when such measurements can be trusted.

We study two widely used measurements in toy models and synthetic dictionaries whose true features are known: the feature dimensionality of Elhage et al.\ \cite{elhage2022} and the feature recovery rate (FRR) with a cosine threshold. We characterize exactly where feature dimensionality is discontinuous: where a feature's weight vector vanishes while some direction is orthogonal to all the other feature vectors. In the earlier work, the apparent failure came from a threshold applied to small feature vectors. We replace feature dimensionality with a regularized (ridge) version and prove sharp stability bounds. The ridge leverage score changes by at most $3\sqrt3/(8\sqrt\lambda)$ times the operator-norm change of the weights, and this constant cannot be improved. For the regularized dimensionality, the stability constant grows like $\rho/4$ in the scale $\rho=R/\sqrt\lambda$, with explicit upper and lower bounds. We also prove certificates that state when a claim ``feature $i$ is recovered'' must survive a given change of the dictionary. Every proof was checked by an independent blind reviewer, with a few minor points still open. A novelty audit found no earlier statement of the stability bounds or the certificates; parts of the discontinuity result appear in earlier work.

\paragraph{Limitations.} Our experiments use toy models, synthetic dictionaries and small dictionaries trained with three seeds; we have not tested real language models. The certificates held in every test and are informative for small perturbations, but they certify no feature non-trivially across independently trained dictionaries, or even between two checkpoints of one training run. At the standard threshold, whether a feature counts as recovered changed with the seed for $55\%$ of the features. Making such certificates useful for trained models, and closing one remaining gap in the sharp constant for the regularized dimensionality, are left open.

\paragraph{Keywords:} mechanistic interpretability; superposition; sparse autoencoders; feature recovery; ridge leverage scores; stability bounds; Lipschitz constants; certificates; AI safety.

\paragraph{AI use statement.} This work was carried out with substantial help from AI systems. The literature search, the statement of research targets, the novelty audits, the proofs, the experiment code and the text were drafted by a pipeline of AI agents built on Anthropic's Claude models (the Opus- and Sonnet-tier models available in Claude Code in October 2026, including Claude Opus 5.5), run in Claude Code and directed by the authors. The agents had separate roles: a literature scout, a gap analyst, a novelty auditor that saw only the claims, a mathematician, an experimenter, a blind reviewer that saw only each statement and its proof, and an editor that changed presentation only. Claude (Anthropic), in a chat interface, was also used for planning, review and the design of the accompanying website. Every proof was checked by the blind reviewer and by numerical tests. No human mathematician other than the authors has verified the proofs. AI systems are not authors of this work: the authors chose the research questions, made the decisions recorded in the audit trail, and take full responsibility for the content, including any errors.

\paragraph{Data and code availability:} the scripts that produced every measured number and figure will be released in a public repository linked from the record at doi:10.5281/zenodo.23842568.

\paragraph{Competing interests:} the authors declare no competing interests.

\clearpage
\section*{Executive summary}

\paragraph{Question.} Interpretability claims about superposition rest on numbers computed from weights: Elhage's feature dimensionality $D_i$ and its total (the ``sum rule''), and, for sparse dictionaries, the feature recovery rate $\FRR_t$ at a cosine threshold $t$. We ask when such a number can support a claim. Where is the metric discontinuous? Which threshold-free variants are Lipschitz, and with which constants? Under which checkable condition does the verdict ``feature $i$ is recovered'' survive a change of dictionary? All settings are toy models and synthetic dictionaries with known ground truth.

\paragraph{Sources.} The metrics come from Elhage et al.\ \cite{elhage2022} ($D_i$) and Chen et al.\ \cite{chen2026} ($\FRR$, $\talign$). Ridge leverage is that of \cite{cohen2017,elalaoui2015}. We credit Ivanov et al.\ \cite{ivanov2026} (Theorem~8) for the defect identity, Mirsky \cite{mirsky1960} for the singular-value inequality, Tsuzuku et al.\ \cite{tsuzuku2018} for the margin mechanism and Pali and Schnass \cite{pali2023} for the one-sided dictionary distance. Linear-algebra pointers are to Strang \cite{strang2016,strang2023}.

\paragraph{Main results.} All proofs are ours. An independent reviewer confirmed each of them (marker ``ours $\checkmark$''), except for eight statements that keep the marker ``ours $\circ$'' because a point the reviewer raised on them is not yet resolved (Section~\ref{sec:open} and Appendix~\ref{app:labels}).
\begin{enumerate}[leftmargin=*,itemsep=1pt]
\item \emph{Diagnosis} (Theorem~\ref{thm:C1}). $D_i$ is discontinuous exactly on the set $Z_i$ where $W_i=0$ and some unit vector is orthogonal to all other columns; the jump of $D_i$ and of $\Dtot$ is $1$. Away from $Z_i$ we give a Lipschitz modulus of order $R/\|W_i\|^2$, and the order is necessary.
\item \emph{Ridge leverage} (Theorem~\ref{thm:C4b}). For every shape, $|\ell_i(W;\lambda)-\ell_i(W';\lambda)|\le\frac{3\sqrt3}{8\sqrt\lambda}\|W-W'\|_{\op}$, and the constant is sharp. A second, independent proof is in Appendix~\ref{app:B4}. The effective dimension has the same sharp constant in the nuclear norm (Theorem~\ref{thm:C5}).
\item \emph{Regularised dimensionality} $\Dt_i$ (Theorems~\ref{thm:A76} and \ref{thm:A78}). $|\Dt_i(W;\lambda)-\Dt_i(W';\lambda)|\le\big(\frac{3\sqrt3}{8\sqrt\lambda}+\frac{R}{4\lambda}\big)\|W-W'\|_{\op}$. The coefficient $1/4$ of $R/\lambda$ is optimal, and the asymptotic additive constant is $1/(4\sqrt2)$; the dependence on $R$ cannot be removed. A second, independent proof of the coefficient $1/4$ is in Appendix~\ref{app:B7}. The optimal additive constant equals its scalar value on part of the range and is open on $(\rho_1,\rho_2)$, with $\rho_1\approx0.556$ and $\rho_2\approx3.326$ (\S\ref{ssec:C7pp}).
\item \emph{Certification} (\S\ref{sec:T4}). Feature scores are Lipschitz in a line-Hausdorff distance; above the sharp threshold $\cos(\theta_V/2)$ each atom is aligned with at most one feature; a global certificate (C11) and feature-local certificates (C11-loc) bound the change of $\FRR_t$ between two dictionaries. We also show that the uniqueness sentence of Chen et al.\ can fail when signed and absolute cosines are mixed (Remark~\ref{rem:chen}).
\item \emph{Measurements} (\S\ref{sec:T5}). On 12 trained toy models, the regularised sum rule gave the same verdict at all 21 tested values of its smoothing parameter in the range fixed in advance, while the version that ignores short feature vectors gave the opposite verdict for two of the models. In all searches no pair exceeded the proved constants. The certificates were never wrong, but across training seeds, and between two checkpoints of one run, they certify no feature non-trivially. At $t=\talign$, $55.47\%$ of the features change status across three seeds while the area $\bar c$ under the FRR curve moves by at most $0.0054$.
\end{enumerate}

\paragraph{What was verified.} The proofs are complete as written, with every step justified. An independent reviewer confirmed every proof (marker $\checkmark$), except on eight statements (marker $\circ$; Appendix~\ref{app:labels}). All numerical checks passed.

\paragraph{What failed or is open.} The scalar value of the additive constant for $\Dt_i$ is not proved on $(\rho_1,\rho_2)$ for $d,n\ge2$; the remaining two-variable inequality is a candidate for a computer-assisted check. The certificates are uninformative on trained dictionaries. The optimal coefficient in the local modulus of $D_i$ lies in $[1/2,4]$ and is not determined.

\newpage
\section*{Roadmap}

Section~\ref{sec:notation} fixes one notation. Section~\ref{sec:defs} gives the definitions and Section~\ref{sec:prelim} the tools, elementary facts and a $2\times2$ family used three times. Section~\ref{sec:T1} diagnoses $D_i$. Section~\ref{sec:T2} proves the stability results for $\ell_i$, $\deff$ and $\Dt_i$. Section~\ref{sec:T4} proves the certificates for $\FRR_t$. Section~\ref{sec:T5} reports our measurements, and Section~\ref{sec:open} lists our contributions with their status and novelty verdicts, followed by the open problems. Appendix~\ref{app:labels} explains the markers; Appendices~\ref{app:B4} and \ref{app:B7} give independent second proofs; Appendix~\ref{app:glossary} is the glossary.

\begin{center}
\begin{tikzpicture}[x=1cm,y=1cm,every node/.style={draw,rounded corners,align=center,font=\footnotesize,inner sep=3pt}]
\node (bg) at (0,0) {Definitions D0--D7, tools,\\ P1--P4, $2\times2$ family};
\node (c1) at (-5.2,-2) {C1: discontinuity\\ of $D_i$ (Thm \ref{thm:C1})};
\node (c4a) at (-1.9,-2) {C4(a) range,\\ scalar case};
\node (c6) at (1.4,-2) {C6: $\Dt_i$, defect\\ identity};
\node (prof) at (4.7,-2) {profile $\Phi$:\\ lower bounds};
\node (c4b) at (-3.5,-4) {C4(b): sharp $\frac{3\sqrt3}{8\sqrt\lambda}$\\ (Thm \ref{thm:C4b}; App.\ \ref{app:B4})};
\node (c5) at (-3.5,-6) {C5: $\deff$ in $\|\cdot\|_*$\\ (Thm \ref{thm:C5})};
\node (c7) at (2.6,-4) {C7$'$: $\frac{3\sqrt3}{8\sqrt\lambda}+\frac{R}{4\lambda}$,\\ limits $\frac14$, $\frac1{4\sqrt2}$ (App.\ \ref{app:B7})};
\node (c7pp) at (2.6,-6) {C7$''$: partial;\\ open on $(\rho_1,\rho_2)$};
\node (ang) at (-5.2,-8) {line angle,\\ chord lemmas};
\node (c8) at (-2.2,-8) {C8: scores\\ Lipschitz in $H$};
\node (c9) at (0.8,-8) {C9(a): uniqueness\\ above $\cos(\theta_V/2)$};
\node (c11) at (4.0,-8) {C11, C11-loc:\\ certificates};
\node (t5) at (0,-10) {T5: measurements (\S\ref{sec:T5})};
\draw[->] (bg) -- (c1);
\draw[->] (bg) -- (c4a);
\draw[->] (bg) -- (c6);
\draw[->] (bg) -- (prof);
\draw[->] (c4a) -- (c4b);
\draw[->] (c4a) -- (c5);
\draw[->] (c6) -- (c7);
\draw[->] (prof) -- (c7);
\draw[->] (c7) -- (c7pp);
\draw[->] (ang) -- (c8);
\draw[->] (ang) -- (c9);
\draw[->] (c8) -- (c11);
\draw[->] (c9) -- (c11);
\draw[->] (c6) .. controls (6.8,-6) and (6.8,-9) .. (t5);
\draw[->] (c11) -- (t5);
\end{tikzpicture}
\end{center}

\noindent The certification part (bottom row) depends only on the definitions D0 and D7 and on the line-angle lemmas of \S\ref{ssec:angle}; it does not use the stability results for $\Dt_i$.

\section{Notation and conventions}\label{sec:notation}

This section fixes the notation for the whole document. Every symbol is defined once; the glossary (Appendix~\ref{app:glossary}) lists them all.

\begin{definition}[D0: conventions]\label{def:D0}\MkDef
\begin{itemize}[itemsep=1pt]
\item $d,n,M,M'$ are positive integers, and $[n]=\{1,\dots,n\}$.
\item On $\R^p$, $\langle x,y\rangle=x^\top y$ and $\|x\|=\langle x,x\rangle^{1/2}$. For $x\ne0$, $\hat x=x/\|x\|$. For $x,y\ne0$, $\cos(x,y)=\langle x,y\rangle/(\|x\|\,\|y\|)$.
\item For a real matrix $B$: $\|B\|_{\op}$ is its largest singular value, $\|B\|_F=\big(\sum_{j,k}B_{jk}^2\big)^{1/2}$ its Frobenius norm and $\|B\|_*$ its nuclear norm (the sum of its singular values); $\sigma_1(B)\ge\sigma_2(B)\ge\dots$ are its singular values.
\item For a symmetric $B\in\R^{p\times p}$: $\nu_{\min}(B)$ and $\nu_{\max}(B)$ are its smallest and largest eigenvalues, and $\nu_1(B)\ge\dots\ge\nu_p(B)$ all its eigenvalues.
\item $I_p$ is the $p\times p$ identity, $O(d)$ the group of orthogonal $d\times d$ matrices, $\#\mathcal U$ the number of elements of a finite set $\mathcal U$, and $\one[\cdot]$ the indicator of a condition. $\mathsf e_1,\dots,\mathsf e_p$ are the standard basis vectors of $\R^p$ ($p$ is clear from the context).
\end{itemize}
\end{definition}

\paragraph{Dummy letters.} The letters $u,x,y,z$ (with primes, stars or subscripts), $B,B'$ (matrices), $\mathsf f$ (functions) and $\vartheta$ (with subscripts or primes; angles) are bound variables: each is introduced, with its range, inside the statement or proof that uses it, and has no meaning outside it. Sans-serif letters other than $\mathsf e_j$, $\mathsf a,\mathsf a',\mathsf b,\mathsf b',\mathsf q_i,\mathsf h$ (Definition~\ref{def:D1}) are local abbreviations: each is defined at the start of the one proof that uses it. Every other symbol has one meaning throughout.

\paragraph{Notation of the sources.} The table maps the notation of the main sources to ours.

\begin{center}
\begin{tabular}{@{}p{0.27\textwidth}p{0.36\textwidth}p{0.29\textwidth}@{}}
\toprule
Source & Their object (in words) & Ours \\
\midrule
Elhage et al.\ \cite{elhage2022} & feature dimensionality & $D_i$ (Definition~\ref{def:D2})\\
Scherlis et al.\ \cite{scherlis2022} & capacity of a feature & $D_i$\\
Ivanov et al.\ \cite{ivanov2026} & frame operator; leverage with the pseudo-inverse & $S$; $\ell^0_j$ (Proposition~\ref{prop:I8})\\
Cohen et al.\ \cite{cohen2017} & ridge leverage score of a column & $\ell_i(W;\lambda)$ (Definition~\ref{def:D4})\\
Chen et al.\ \cite{chen2026} & alignment threshold, feature recovery rate & $\talign(V)$, $\FRR_t$ (Definition~\ref{def:D7})\\
Pali and Schnass \cite{pali2023} & one-sided distance between dictionaries of unit atoms & $\max_m e_m(A\to A')$ on normalised atoms\\
\bottomrule
\end{tabular}
\end{center}

\section{Definitions}\label{sec:defs}

These are the objects the theorems are about: the weight matrix and its frame operator, the four dimensionality-type metrics, the range of the ridge parameter, and the dictionary setting for feature recovery.

\begin{definition}[D1: weights and frame operator]\label{def:D1}\MkDef
\begin{itemize}[itemsep=1pt]
\item $W\in\R^{d\times n}$ has columns $W_1,\dots,W_n\in\R^d$. Column $W_i$ is the direction of feature $i$. $W_{-i}\in\R^{d\times(n-1)}$ is the matrix of the columns $W_j$, $j\ne i$, in increasing order of $j$; for $n=1$ it has no columns, and we set $W_{-i}^\top y=0$ for every $y$, all norms of these being $0$ (empty sums are $0$).
\item $S=S(W)=WW^\top=\sum_{j=1}^nW_jW_j^\top$, $S_{-i}=S_{-i}(W)=\sum_{j\ne i}W_jW_j^\top=W_{-i}W_{-i}^\top$, and $r=\rank S$.
\item $\sigma_1\ge\dots\ge\sigma_{\min(d,n)}\ge0$ are the singular values of $W$, so the nonzero eigenvalues of $S$ are $\sigma_1^2,\dots,\sigma_r^2$ (Lemma~\ref{lem:P2}). $\omega(W)=\tr(S)/d$ is the mean eigenvalue of $S$.
\item For $\lambda>0$: $S_\lambda=S+\lambda I_d$, $S_{-i,\lambda}=S_{-i}+\lambda I_d$ and $K_\lambda=W^\top W+\lambda I_n$.
\item For a second matrix $W'\in\R^{d\times n}$ (with $S'=S(W')$): $E=W'-W$, $\mathsf h=\|W'-W\|_{\op}$, $R=\max(\|W\|_{\op},\|W'\|_{\op})$, $R_{\min}=\min(\|W\|_{\op},\|W'\|_{\op})$, and $\Rb$ denotes any number with $\Rb\ge R$ and $\Rb>0$. For $\eta\in\R$, $W(\eta)=W+\eta E$, with columns $W_j(\eta)$; $S(\eta)$, $S_{-i}(\eta)$, $S_\lambda(\eta)$, $K_\lambda(\eta)$ are formed from $W(\eta)$.
\item For fixed $i$: $\mathsf a=\|W_i\|$ and $\mathsf q_i=W_i^\top SW_i$, and, if $W_i\ne0$, $\mathsf b=\|W^\top\hat W_i\|$; $\mathsf a'$, $\mathsf q_i'$, $\mathsf b'$ are the same quantities for $W'$.
\end{itemize}
\end{definition}

\begin{definition}[D2: feature dimensionality]\label{def:D2}\MkDefS{source \cite{elhage2022}; case $W_i=0$ ours}
For $W_i\ne0$,
\begin{equation}\label{eq:D2}
D_i(W)=\frac{\|W_i\|^2}{\sum_{j=1}^n\langle\hat W_i,W_j\rangle^2}=\frac{\|W_i\|^4}{W_i^\top SW_i}\qquad(\text{the sum includes }j=i),
\end{equation}
and $D_i(W)=0$ if $W_i=0$. The total is $\Dtot(W)=\sum_{i=1}^nD_i(W)$.
\end{definition}

\begin{definition}[D3: norm-thresholded dimensionality]\label{def:D3}\MkDef
For $\tau\ge0$, $D_i^{(\tau)}(W)=D_i(W)$ if $\|W_i\|>\tau$ and $D_i^{(\tau)}(W)=0$ otherwise; $\Dtot^{(\tau)}(W)=\sum_iD_i^{(\tau)}(W)$. So $D_i^{(0)}=D_i$. In the runs of \S\ref{sec:T5}, $\tau=0.2$ is the thresholded version and $\tau=0$ the unthresholded one; the test is computed as $\|W_i\|^2>\tau^2$, which is equivalent.
\end{definition}

\begin{definition}[D4: ridge leverage and effective dimension]\label{def:D4}\MkDefS{source \cite{cohen2017,elalaoui2015}}
For $\lambda>0$,
\begin{equation}\label{eq:D4}
\ell_i(W;\lambda)=W_i^\top S_\lambda^{-1}W_i,\qquad \deff(W;\lambda)=\sum_{i=1}^n\ell_i(W;\lambda).
\end{equation}
($S_\lambda$ is invertible by Lemma~\ref{lem:P2}; the closed forms of $\deff$ are Lemma~\ref{lem:P3}.)
\end{definition}

\begin{definition}[D5: regularised dimensionality]\label{def:D5}\MkDef
For $\lambda>0$ and $W_i\ne0$,
\begin{equation}\label{eq:D5}
\Dt_i(W;\lambda)=\frac{\|W_i\|^4}{W_i^\top SW_i+\lambda\|W_i\|^2}=\frac{\|W_i\|^2}{\sum_{j=1}^n\langle\hat W_i,W_j\rangle^2+\lambda},
\end{equation}
and $\Dt_i(W;\lambda)=0$ if $W_i=0$. The total is $\Dttot(W;\lambda)=\sum_i\Dt_i(W;\lambda)$.
\end{definition}

\begin{definition}[D6: range of the ridge parameter]\label{def:D6}\MkDef
For $W\ne0$,
\begin{equation}\label{eq:D6}
\Lambda(W)=\big[10^{-3}\,\omega(W),\ 10^{-1}\,\omega(W)\big].
\end{equation}
Then $\omega(W)>0$, so $\Lambda(W)\subset(0,\infty)$; $\Lambda(W)$ is not defined for $W=0$. A statement ``for $\lambda\in\Lambda(W)$'' fixes the number $\lambda$ from the unperturbed $W$; when $W$ is perturbed, $\lambda$ stays fixed. All results of \S\ref{sec:T2} hold for every $\lambda>0$, so in particular for every $\lambda\in\Lambda(W)$.
\end{definition}

\begin{definition}[D7: dictionary setting]\label{def:D7}\MkDefS{ours, except where a source is named}
\begin{itemize}[itemsep=1pt]
\item $V\in\R^{d\times n}$, with $n\ge2$, has nonzero columns $v_1,\dots,v_n$ (the ground-truth features), and $|\cos(v_i,v_j)|<1$ for $i\ne j$. We write $\mu_{ij}=\cos(v_i,v_j)$.
\item $\cA_M$ is the set of dictionaries $A\in\R^{d\times M}$ whose columns (atoms) $a_1,\dots,a_M$ are all nonzero; $A'\in\cA_{M'}$ has atoms $a'_1,\dots,a'_{M'}$. Indices $m$ run over $[M]$ and $m'$ over $[M']$.
\item \emph{Line separation:} $\theta_V=\min_{i\ne j}\arccos|\cos(v_i,v_j)|\in(0,\pi/2]$.
\item \emph{Signed coherence and alignment threshold} (source: Chen et al.\ \cite{chen2026}): $\mu(V)=\max_{i\ne j}\mu_{ij}$ and $\talign(V)=\cos\big(\tfrac13\arccos\mu(V)\big)$.
\item \emph{Per-feature score:} $c_i(A)=\max_{m\in[M]}|\cos(a_m,v_i)|\in[0,1]$. \emph{Per-atom score:} $g_m(A)=\max_{i\in[n]}|\cos(a_m,v_i)|$; $c_i(A')$ and $g_{m'}(A')=\max_{i\in[n]}|\cos(a'_{m'},v_i)|$ are defined in the same way.
\item \emph{Feature recovery rate at threshold $t\in(0,1]$} (source: Chen et al.\ \cite{chen2026}, with the absolute cosine):
\begin{equation}\label{eq:FRR}
\FRR_t(A)=\frac1n\#\{i\in[n]:c_i(A)\ge t\}.
\end{equation}
\emph{Feature $i$ is recovered by $A$ at threshold $t$} if $c_i(A)\ge t$.
\item \emph{Area under the FRR curve:} $\bar c(A)=\frac1n\sum_ic_i(A)$.
\item \emph{Smoothed FRR:} for $\beta\in(0,1)$, $\FRR_{t,\beta}(A)=\frac1n\sum_i\psi_{t,\beta}(c_i(A))$, where $\psi_{t,\beta}(u)=0$ for $u\le t-\beta/2$, $\psi_{t,\beta}(u)=1$ for $u\ge t+\beta/2$, and $\psi_{t,\beta}(u)=(u-t+\beta/2)/\beta$ for $t-\beta/2<u<t+\beta/2$ (the affine interpolation between the values $0$ and $1$).
\item \emph{Line distances:} $e_m(A\to A')=\min_{m'\in[M']}\min_{s\in\{\pm1\}}\|\hat a_m-s\,\hat a'_{m'}\|$, and $e_{m'}(A'\to A)$ in the same way. The line-Hausdorff distance is $H(A,A')=\max\big(\max_me_m(A\to A'),\ \max_{m'}e_{m'}(A'\to A)\big)$.
\item \emph{Margin count:} for $\gamma>0$, $N_\gamma(A;t)=\#\{i:|c_i(A)-t|<\gamma\}$.
\end{itemize}
\end{definition}

\section{Tools and preliminary facts}\label{sec:prelim}

We first list the standard facts we use without proof, then prove a few elementary facts (P1--P4) and compute the slopes of one $2\times2$ family that serves three later results.

\subsection{Tools}\label{ssec:tools}

Each tool is a standard fact that we use and do not prove. Where a pointer is given, ``not verified'' means that we have not checked the section number in the book.

\begin{toolstmt}[spectral theorem]\label{tool:spec}\MkTool
Every real symmetric $B\in\R^{p\times p}$ can be written $B=U\diag(\nu_1,\dots,\nu_p)U^\top$ with $U\in O(p)$ and real $\nu_1\ge\dots\ge\nu_p$, the eigenvalues of $B$; the columns of $U$ are orthonormal eigenvectors. Consequently, for every $x\in\R^p$:
\begin{equation}\label{eq:specbounds}
\nu_{\min}(B)\|x\|^2\le x^\top Bx\le\nu_{\max}(B)\|x\|^2;
\end{equation}
if $\nu_{\min}(B)>0$, then $B^{-1}$ is symmetric and its eigenvalues are the reciprocals of those of $B$; and if $B$ is positive semidefinite, then
\begin{equation}\label{eq:specpsd}
\|Bx\|^2=x^\top B^2x\le\nu_{\max}(B)\,x^\top Bx.
\end{equation}
(Each consequence follows by expanding $x$ in the eigenbasis.) Pointer: Strang \cite{strang2016}, Ch.~6 (symmetric matrices); section number not verified.
\end{toolstmt}

\begin{toolstmt}[Rayleigh characterisation]\label{tool:rayleigh}\MkTool
For real symmetric $B$, $\nu_{\min}(B)=\min_{\|z\|=1}z^\top Bz$, and the minimum is attained at a unit eigenvector for $\nu_{\min}(B)$. (It follows from Tool~\ref{tool:spec}: with $z=Uz'$, $z^\top Bz=\sum_k\nu_kz_k'^2$ and $\|z'\|=\|z\|$.) Pointer: Strang \cite{strang2016}, Ch.~6; not verified.
\end{toolstmt}

\begin{toolstmt}[singular values]\label{tool:sv}\MkTool
For $B\in\R^{d\times n}$ with singular values $\sigma_1(B)\ge\dots\ge\sigma_{\min(d,n)}(B)\ge0$:
(i) $\|B\|_{\op}=\sigma_1(B)=\max_{\|z\|=1}\|Bz\|$;
(ii) the eigenvalues of $BB^\top$ are $\sigma_1(B)^2,\dots,\sigma_{\min(d,n)}(B)^2$ together with $d-\min(d,n)$ zeros;
(iii) the number of nonzero singular values is $\rank B=\rank BB^\top\le\min(d,n)$;
(iv) $\|B\|_F^2=\tr(B^\top B)=\sum_k\sigma_k(B)^2$.
Pointers: Strang \cite{strang2016}, Ch.~7 (singular value decomposition), not verified; $\rank AA^\top=\rank A$ is boxed fact~3 of Appendix~A1 of the 6th edition, p.~400 \cite{strang2023} (verified).
\end{toolstmt}

\begin{toolstmt}[norm facts]\label{tool:op}\MkTool
For real matrices $B$ and vectors $x$: $\|Bx\|\le\|B\|_{\op}\|x\|$; $\|B^\top\|_{\op}=\|B\|_{\op}$; $\|B\|_{\op}^2=\nu_{\max}(BB^\top)$; $\|\cdot\|_{\op}$ satisfies the triangle inequality and $\|cB\|_{\op}=|c|\,\|B\|_{\op}$. The Euclidean norm satisfies the triangle inequality and the reverse triangle inequality $|\|y\|-\|y'\||\le\|y-y'\|$; for real numbers, $\big||y|-|y'|\big|\le|y-y'|$.
\end{toolstmt}

\begin{toolstmt}[Cauchy--Schwarz]\label{tool:cs}\MkTool
For an inner product $\langle\cdot,\cdot\rangle_\ast$ on $\R^p$, $\langle z,z'\rangle_\ast^2\le\langle z,z\rangle_\ast\langle z',z'\rangle_\ast$, with equality if and only if $z,z'$ are linearly dependent. In particular $|\langle x,y\rangle|\le\|x\|\,\|y\|$. Pointer: Strang \cite{strang2016}, Ch.~1 (dot products, lengths and angles), for the standard inner product; section number not verified.
\end{toolstmt}

\begin{toolstmt}[Minkowski in $\R^2$]\label{tool:mink}\MkTool
For real $y_1,y_2,y_3$: $\sqrt{y_1^2+(y_2+y_3)^2}\le\sqrt{y_1^2+y_2^2}+|y_3|$.
\end{toolstmt}

\begin{toolstmt}[positive definite matrices]\label{tool:pd}\MkTool
If $B$ is a real symmetric positive semidefinite matrix and $\lambda>0$, then $B+\lambda I$ is symmetric positive definite. A symmetric positive definite matrix $K$ is invertible ($Kz=0$ gives $z^\top Kz=0$, hence $z=0$; Tool~\ref{tool:dim}). Its inverse is symmetric positive definite, so its diagonal entries are $>0$. For such $K$, $\langle y,y'\rangle_K=y^\top Ky'$ is an inner product.
\end{toolstmt}

\begin{toolstmt}[trace]\label{tool:trace}\MkTool
$\tr(BB')=\tr(B'B)$ whenever both products are defined; the trace of a symmetric matrix is the sum of its eigenvalues. Pointer: Strang \cite{strang2016}, Ch.~6; not verified.
\end{toolstmt}

\begin{toolstmt}[dimension count]\label{tool:dim}\MkTool
The nullspace of a $p\times d$ matrix of rank $r'$ has dimension $d-r'$, and a square matrix with trivial nullspace is invertible. Pointer: Strang \cite{strang2016}, 5th ed., \S3.5, p.~181 box and the Counting Theorem, p.~185 (verified for the dimension of the nullspace; the invertibility clause is not on these pages).
\end{toolstmt}

\begin{toolstmt}[Cramer's rule]\label{tool:cramer}\MkTool
An invertible square matrix $B$ satisfies $B^{-1}=\adj(B)/\det B$, where the entries of $\adj(B)$ are polynomials in the entries of $B$.
\end{toolstmt}

\begin{toolstmt}[one-variable calculus]\label{tool:calc}\MkTool
(a) The product rule (also for products of matrix-valued functions, entrywise) and the quotient rule. (b) Mean value theorem: if $h$ is continuous on $[u_1,u_2]$ and differentiable on $(u_1,u_2)$, then $h(u_2)-h(u_1)=h'(u)(u_2-u_1)$ for some $u\in(u_1,u_2)$; in particular, if $|h'|\le L$ on $(u_1,u_2)$, then $|h(u_2)-h(u_1)|\le L(u_2-u_1)$. (c) Fundamental theorem of calculus: if $h$ is continuously differentiable on $[u_1,u_2]$, then $h(u_2)-h(u_1)=\int_{u_1}^{u_2}h'(u)\,\mathrm du$; hence if $h'\le L$ there, $h(u_2)-h(u_1)\le L(u_2-u_1)$. (d) A function with positive (negative) derivative on an open interval is strictly increasing (decreasing) on its closure. (e) Rational functions are continuous, and differentiable, where the denominator does not vanish. (f) A continuous function $\mathsf f\ge0$ on $[u_1,u_2]$, $u_1<u_2$, with $\mathsf f>0$ at some point has $\int_{u_1}^{u_2}\mathsf f>0$. (g) For continuous $\mathsf f_1,\mathsf f_2$ on $[u_1,u_2]$, $u_1<u_2$: the integral is linear; $\mathsf f_1\le\mathsf f_2$ implies $\int_{u_1}^{u_2}\mathsf f_1\le\int_{u_1}^{u_2}\mathsf f_2$; hence $\big|\int_{u_1}^{u_2}\mathsf f_1\big|\le\int_{u_1}^{u_2}|\mathsf f_1|$; and $\int_{u_1}^{u_2}L\,\mathrm du=L(u_2-u_1)$ for a constant $L$. (h) Intermediate value theorem: a continuous function on $[u_1,u_2]$ takes every value between its values at $u_1$ and $u_2$. No source in our background is cited for these.
\end{toolstmt}

\begin{toolstmt}[pseudo-inverse]\label{tool:pinv}\MkTool
The pseudo-inverse of an invertible matrix is its inverse.
\end{toolstmt}

\begin{toolstmt}[Mirsky's inequality, nuclear-norm case]\label{tool:mirsky}\MkTool
For all $B,B'\in\R^{d\times n}$, with singular values in non-increasing order,
\begin{equation}\label{eq:mirsky}
\sum_{k=1}^{\min(d,n)}\big|\sigma_k(B)-\sigma_k(B')\big|\le\|B-B'\|_* .
\end{equation}
Source: Mirsky \cite{mirsky1960}, full text not seen. Secondary sources seen only as abstracts or search snippets attribute to Mirsky the general inequality $\|\diag(\sigma(B)-\sigma(B'))\|\le\|B-B'\|$ for every unitarily invariant norm $\|\cdot\|$. Inequality \eqref{eq:mirsky} is its case for the Schatten-1 (nuclear) norm, whose symmetric gauge function is the $\ell^1$ norm; we have not seen that specialisation stated in any source we read. We credit \eqref{eq:mirsky} to Mirsky and claim no novelty for its use.
\end{toolstmt}

\begin{toolstmt}[trigonometry]\label{tool:trig}\MkTool
$\cos$ is continuous and strictly decreasing on $[0,\pi]$; $\arccos:[-1,1]\to[0,\pi]$ is its inverse and is strictly decreasing; $\arccos(-u)=\pi-\arccos u$ for $u\in[-1,1]$; $\sin\ge0$ on $[0,\pi]$, $\sin$ is strictly increasing on $[0,\pi/2]$, $\sin\le1$, and $\sin^2+\cos^2=1$. Values: $\cos0=1$, $\cos(\pi/4)=1/\sqrt2$, $\cos(\pi/2)=0$; hence $\arccos1=0$ and $\arccos0=\pi/2$. For real $\vartheta_1,\vartheta_2,\vartheta$: $\cos(\vartheta_1+\vartheta_2)=\cos\vartheta_1\cos\vartheta_2-\sin\vartheta_1\sin\vartheta_2$; $\cos\vartheta_1-\cos\vartheta_2=2\sin\frac{\vartheta_1+\vartheta_2}2\,\sin\frac{\vartheta_2-\vartheta_1}2$; $1-\cos\vartheta=2\sin^2(\vartheta/2)$ and $1+\cos\vartheta=2\cos^2(\vartheta/2)$.
\end{toolstmt}

\begin{toolstmt}[monotone maps and extrema]\label{tool:mono}\MkTool
If $\mathsf f$ is non-decreasing on a set containing the finite nonempty set $\{u_1,\dots,u_k\}$, then $\mathsf f(\min_ju_j)=\min_j\mathsf f(u_j)$ and $\mathsf f(\max_ju_j)=\max_j\mathsf f(u_j)$; if $\mathsf f$ is non-increasing, the same holds with $\min$ and $\max$ exchanged on one side. The square root is strictly increasing on $[0,\infty)$.
\end{toolstmt}

\begin{toolstmt}[continuity]\label{tool:cont}\MkTool
Equip $\R^{d\times n}$ with the distance $\|B-B'\|_{\op}$ (a metric by Tool~\ref{tool:op}). A function $\mathsf f:\R^{d\times n}\to\R$ with $|\mathsf f(B)-\mathsf f(B')|\le\|B-B'\|_{\op}$ for all $B,B'$ is continuous. Constants are continuous; sums and products of continuous functions are continuous; a quotient of continuous functions is continuous on the set where its denominator does not vanish; and if $\mathsf f$ is continuous, the set $\{B:\mathsf f(B)\ne0\}$ is open.
\end{toolstmt}

\begin{toolstmt}[Lipschitz compositions]\label{tool:lip}\MkTool
If $\mathsf f_1,\mathsf f_2:\R\to\R$ are Lipschitz with constants $L_1,L_2$, then $\mathsf f_1\circ\mathsf f_2$ is Lipschitz with constant $L_1L_2$; the maps $u\mapsto\max(0,u)$ and $u\mapsto\min(1,u)$ are $1$-Lipschitz.
\end{toolstmt}

\begin{toolstmt}[Riemann integral]\label{tool:riemann}\MkTool
The Riemann integral is linear, and $\int_0^1\one[u\le y]\,\mathrm du=y$ for $y\in[0,1]$ (the integrand is the indicator of the interval $[0,y]$, of length $y$); changing an integrand at one point does not change its integral.
\end{toolstmt}

\subsection{Preliminary facts P1--P4}\label{ssec:P}

These four facts are classical; we prove them for completeness.

\begin{lemma}[P1: norm facts]\label{lem:P1}\MkOursChk
Let $B,B'\in\R^{d\times n}$.
\begin{enumerate}[label=P1.\arabic*,leftmargin=*]
\item $\|B\mathsf e_i\|\le\|B\|_{\op}$, so $\|B_i-B'_i\|\le\|B-B'\|_{\op}$ for every column $i$. (By Tool~\ref{tool:sv}(i), since $\|\mathsf e_i\|=1$ and $B\mathsf e_i=B_i$.)
\item $|B_{jk}|=|\mathsf e_j^\top B\mathsf e_k|\le\|\mathsf e_j\|\,\|B\mathsf e_k\|\le\|B\|_{\op}$ (Tool~\ref{tool:cs} and P1.1). Applied to $B-B'$, this gives $|B_{jk}-B'_{jk}|\le\|B-B'\|_{\op}$, so each entry map $B\mapsto B_{jk}$ is continuous (Tool~\ref{tool:cont}). Hence, by Tool~\ref{tool:cont}, every polynomial in the entries of $B$ is continuous, and every quotient of two such polynomials is continuous on the set where its denominator does not vanish. All continuity statements below refer to $\|\cdot\|_{\op}$.
\item $\|B\|_{\op}\le\|B\|_F$: for $\|z\|=1$, $\|Bz\|=\|\sum_kz_kB_k\|\le\sum_k|z_k|\,\|B_k\|\le\big(\sum_kz_k^2\big)^{1/2}\big(\sum_k\|B_k\|^2\big)^{1/2}=\|B\|_F$ (triangle inequality, then Tool~\ref{tool:cs} in $\R^n$; the last equality is $\sum_k\|B_k\|^2=\sum_{j,k}B_{jk}^2$ and D0); take the maximum over $z$ (Tool~\ref{tool:sv}(i)).
\item $\|B^\top z\|\le\|B\|_{\op}\|z\|$ for $z\in\R^d$: $\|B^\top z\|^2=\langle BB^\top z,z\rangle\le\|B(B^\top z)\|\,\|z\|\le\|B\|_{\op}\|B^\top z\|\,\|z\|$ (Tools~\ref{tool:cs} and \ref{tool:sv}(i)); divide by $\|B^\top z\|$ if it is nonzero.
\item For $B\in\R^{d\times d}$ and $\|z\|=1$ (symmetry of $B$ is not needed; below it is used only for symmetric $B$): $|z^\top Bz|\le\|z\|\,\|Bz\|\le\|B\|_{\op}$ (Tools~\ref{tool:cs} and \ref{tool:sv}(i)).
\end{enumerate}
\end{lemma}

\begin{lemma}[P2: positivity]\label{lem:P2}\MkOursChk
For $z\in\R^d$, $z^\top Sz=\sum_{j=1}^n\langle W_j,z\rangle^2\ge0$ and $z^\top S_{-i}z=\sum_{j\ne i}\langle W_j,z\rangle^2\ge0$. For $\lambda>0$, $z^\top S_\lambda z\ge\lambda\|z\|^2$ and $z^\top S_{-i,\lambda}z\ge\lambda\|z\|^2$; $S_\lambda$ is invertible and $S_\lambda^{-1}$ is symmetric. The eigenvalues of $S$ are $\nu_k(S)=\sigma_k^2$ for $k\le\min(d,n)$ and $\nu_k(S)=0$ for $k>\min(d,n)$, and exactly $r$ of them are nonzero, namely $\sigma_1^2,\dots,\sigma_r^2$.
\end{lemma}
\begin{proof}
\begin{steps}
\item By D1, $z^\top Sz=\sum_jz^\top W_jW_j^\top z=\sum_j\langle W_j,z\rangle^2\ge0$, and the same with $j\ne i$ for $S_{-i}$.
\item By D1, $z^\top S_\lambda z=z^\top Sz+\lambda\|z\|^2\ge\lambda\|z\|^2$ and $z^\top S_{-i,\lambda}z=z^\top S_{-i}z+\lambda\|z\|^2\ge\lambda\|z\|^2$ (Step~1).
\item If $S_\lambda z=0$, then $0=z^\top S_\lambda z\ge\lambda\|z\|^2$ (Step~2), so $z=0$; hence $S_\lambda$ is invertible (Tool~\ref{tool:dim}). From $S_\lambda^\top=S_\lambda$, $(S_\lambda^{-1})^\top=(S_\lambda^\top)^{-1}=S_\lambda^{-1}$.
\item $S=WW^\top$ is symmetric; by Tool~\ref{tool:sv}(ii) its eigenvalues are $\sigma_1^2,\dots,\sigma_{\min(d,n)}^2$ and $d-\min(d,n)$ zeros, which in non-increasing order (Tool~\ref{tool:spec}) gives $\nu_k(S)$ as stated.
\item By Tool~\ref{tool:sv}(iii), exactly $r=\rank WW^\top$ singular values are nonzero; with the ordering, these are $\sigma_1,\dots,\sigma_r$, so the nonzero eigenvalues are $\sigma_1^2,\dots,\sigma_r^2$.
\end{steps}
\end{proof}

\begin{lemma}[P3: formula for $\deff$]\label{lem:P3}\MkOursChk
For $\lambda>0$,
\begin{equation}\label{eq:deff}
\begin{gathered}
\deff(W;\lambda)=\tr\big(SS_\lambda^{-1}\big)=\sum_{k=1}^{\min(d,n)}\frac{\sigma_k^2}{\sigma_k^2+\lambda}=\sum_{k=1}^r\frac{\sigma_k^2}{\sigma_k^2+\lambda},\\
d-\deff(W;\lambda)=\lambda\tr\big(S_\lambda^{-1}\big)=\sum_{k=1}^d\frac{\lambda}{\nu_k(S)+\lambda}.
\end{gathered}
\end{equation}
\end{lemma}
\begin{proof}
\begin{steps}
\item $\sum_i\ell_i=\sum_iW_i^\top S_\lambda^{-1}W_i=\tr(W^\top S_\lambda^{-1}W)=\tr(S_\lambda^{-1}WW^\top)=\tr(S_\lambda^{-1}S)$ (D4; the diagonal of $W^\top S_\lambda^{-1}W$; Tool~\ref{tool:trace}).
\item With $S=U\diag(\nu_k(S))U^\top$ (Tool~\ref{tool:spec}), $S_\lambda=U\diag(\nu_k(S)+\lambda)U^\top$ and $S_\lambda^{-1}=U\diag((\nu_k(S)+\lambda)^{-1})U^\top$ (multiply out, $U^\top U=I_d$). So $S_\lambda^{-1}S=U\diag(\nu_k(S)/(\nu_k(S)+\lambda))U^\top$, whose trace is $\sum_k\nu_k(S)/(\nu_k(S)+\lambda)$ (Tool~\ref{tool:trace}). By Lemma~\ref{lem:P2} this is the middle and right expressions in \eqref{eq:deff} (zero eigenvalues contribute $0$).
\item $d-\sum_k\frac{\nu_k}{\nu_k+\lambda}=\sum_{k=1}^d\frac{\lambda}{\nu_k+\lambda}=\lambda\tr(S_\lambda^{-1})$ (Step~2 and Tool~\ref{tool:trace}).
\end{steps}
\end{proof}

\begin{lemma}[P4: the Rayleigh form of $D_i$]\label{lem:P4}\MkOursChk
For $W_i\ne0$: $W_i^\top SW_i=\sum_j\langle W_i,W_j\rangle^2=\|W_i\|^4+W_i^\top S_{-i}W_i\ge\|W_i\|^4>0$, $W_i^\top SW_i=\|W_i\|^2\,\|W^\top\hat W_i\|^2$, and
\begin{equation}\label{eq:Dform}
D_i(W)=\frac{\|W_i\|^2}{\|W^\top\hat W_i\|^2},\qquad \|W^\top\hat W_i\|\ge\|W_i\|,\qquad 0<D_i(W)\le1 .
\end{equation}
\end{lemma}
\begin{proof}
\begin{steps}
\item By Lemma~\ref{lem:P2}, Step~1, with $z=W_i$: $W_i^\top SW_i=\sum_j\langle W_j,W_i\rangle^2$. The term $j=i$ is $\|W_i\|^4$, and the others sum to $W_i^\top S_{-i}W_i\ge0$ (Lemma~\ref{lem:P2}). So $W_i^\top SW_i\ge\|W_i\|^4>0$.
\item $(W^\top\hat W_i)_j=\langle W_j,\hat W_i\rangle=\langle W_j,W_i\rangle/\|W_i\|$, so $\|W^\top\hat W_i\|^2=W_i^\top SW_i/\|W_i\|^2$ (Step~1).
\item By D2 and Step~2, $D_i(W)=\|W_i\|^4/(\|W_i\|^2\|W^\top\hat W_i\|^2)=\|W_i\|^2/\|W^\top\hat W_i\|^2$.
\item By Steps~1--2, $\|W^\top\hat W_i\|^2\ge\|W_i\|^4/\|W_i\|^2=\|W_i\|^2$, so $\|W^\top\hat W_i\|\ge\|W_i\|>0$ and, by Step~3, $0<D_i(W)\le1$.
\end{steps}
\end{proof}

\subsection{A $2\times2$ family and its slopes}\label{ssec:family}

One explicit family of $2\times2$ matrices shows that the local modulus of $D_i$ (Theorem~\ref{thm:C1}(d)) and the Lipschitz constant of $\Dt_i$ (Proposition~\ref{prop:C7b}, Proposition~\ref{prop:profile}) must grow with $R$. It also gives the lower bounds that match the coefficients of \S\ref{ssec:C7p}.

\begin{definition}[the family; F1 and F2]\label{def:family}\MkDef
Let $d=n=2$, $0<\alpha_0<R_0$, $\phi=\arccos(\alpha_0/R_0)\in(0,\pi/2)$, $\nphi=(\cos\phi,\sin\phi)^\top$, $\nphip=(-\sin\phi,\cos\phi)^\top$, and
\[
W^\star=[\alpha_0\nphi,\ R_0\mathsf e_1],\qquad E^{(\iota)}=[\nphip,\ -\iota\,\nphi],\quad\iota\in\{0,1\}.
\]
For $\lambda>0$, the direction $E^{(0)}$ is called \textsf{F1} and $E^{(1)}$ is called \textsf{F2} when $\alpha_0=\sqrt{\lambda/2}$ (so $R_0^2>\lambda/2$). For $\lambda\ge0$ and $\eta\in\R$ let $W^\star(\eta)=W^\star+\eta E^{(\iota)}$ (columns $W^\star_1(\eta),W^\star_2(\eta)$; this curve is not the segment $W(\eta)$ of D1),
\[
\pi_{12}(\eta)=\langle W^\star_1(\eta),W^\star_2(\eta)\rangle,\qquad
\mathcal C_\lambda(\eta)=\frac{\|W^\star_1(\eta)\|^4}{\|W^\star_1(\eta)\|^4+\pi_{12}(\eta)^2+\lambda\|W^\star_1(\eta)\|^2}.
\]
(The dependence on $\alpha_0,R_0,\iota$ is not shown.) The parameters of the later statements are: Theorem~\ref{thm:C1}(d) is $\lambda=0$, $\iota=0$; Proposition~\ref{prop:C7b} is $\lambda>0$, $\alpha_0=\sqrt\lambda$, $\iota=0$; \textsf{F1} and \textsf{F2} are $\lambda>0$, $\alpha_0=\sqrt{\lambda/2}$, $\iota=0,1$.
\end{definition}

\begin{lemma}[slopes of the family]\label{lem:family}\MkOursChk
\begin{enumerate}[label=(\roman*)]
\item $\mathcal C_0(\eta)=D_1(W^\star(\eta))$ and, for $\lambda>0$, $\mathcal C_\lambda(\eta)=\Dt_1(W^\star(\eta);\lambda)$, for all $\eta$.
\item $\|E^{(\iota)}\|_{\op}=1$ and $R_0\le\|W^\star\|_{\op}\le(R_0^2+\alpha_0^2)^{1/2}$.
\item $\mathcal C_\lambda$ is differentiable and
\begin{equation}\label{eq:slope}
\mathcal C_\lambda'(0)=\frac{2\alpha_0^2\big(R_0\sin\phi+\iota\alpha_0\big)}{(2\alpha_0^2+\lambda)^2},\qquad R_0\sin\phi=\sqrt{R_0^2-\alpha_0^2}>0 .
\end{equation}
\end{enumerate}
\end{lemma}
\begin{proof}
Tools used: Tools~\ref{tool:sv}(i), \ref{tool:calc}, \ref{tool:trig}; Lemmas~\ref{lem:P1} and \ref{lem:P4}.
\begin{steps}
\item $\langle\nphi,\nphip\rangle=0$, $\|\nphi\|=\|\nphip\|=1$, $\langle\nphi,\mathsf e_1\rangle=\cos\phi=\alpha_0/R_0$, $\langle\nphip,\mathsf e_1\rangle=-\sin\phi$, and $\sin\phi=\sqrt{1-\alpha_0^2/R_0^2}>0$ since $\phi\in(0,\pi/2)$ (Tool~\ref{tool:trig}).
\item $W^\star_1(\eta)=\alpha_0\nphi+\eta\nphip$, so $\|W^\star_1(\eta)\|^2=\alpha_0^2+\eta^2>0$ (Step~1).
\item $W^\star_2(\eta)=R_0\mathsf e_1-\iota\eta\nphi$. Expanding with Step~1, $\pi_{12}(\eta)=\alpha_0R_0\cos\phi-\iota\alpha_0\eta-\eta R_0\sin\phi-0=\alpha_0^2-\eta\big(R_0\sin\phi+\iota\alpha_0\big)$.
\item (i): for $n=2$ and a matrix $B$ with columns $B_1\ne0$, $B_2$, $B_1^\top S(B)B_1=\|B_1\|^4+\langle B_1,B_2\rangle^2$ (Lemma~\ref{lem:P4}, Step~1). With $B=W^\star(\eta)$, whose first column is nonzero (Step~2), $\mathcal C_0(\eta)$ is D2 and $\mathcal C_\lambda(\eta)$ is D5.
\item (ii): for $z\in\R^2$, $\|E^{(0)}z\|=|z_1|\le\|z\|$ with equality at $z=\mathsf e_1$; $\|E^{(1)}z\|^2=\|z_1\nphip-z_2\nphi\|^2=z_1^2+z_2^2$ (Step~1). By Tool~\ref{tool:sv}(i), both norms are $1$. $\|W^\star\mathsf e_2\|=R_0$ gives the lower bound, and P1.3 gives $\|W^\star\|_{\op}\le\|W^\star\|_F=(\alpha_0^2+R_0^2)^{1/2}$.
\item (iii): write $\mathcal C_\lambda=\Numfam/\Denfam$ with $\Numfam(\eta)=(\alpha_0^2+\eta^2)^2$ and $\Denfam(\eta)=(\alpha_0^2+\eta^2)^2+\pi_{12}(\eta)^2+\lambda(\alpha_0^2+\eta^2)$. Both are polynomials, and $\Denfam(\eta)\ge(\alpha_0^2+\eta^2)^2>0$, so $\mathcal C_\lambda$ is differentiable (Tool~\ref{tool:calc}(e)).
\item At $\eta=0$: $\Numfam(0)=\alpha_0^4$, $\Numfam'(0)=4\eta(\alpha_0^2+\eta^2)|_{\eta=0}=0$, $\Denfam(0)=2\alpha_0^4+\lambda\alpha_0^2=\alpha_0^2(2\alpha_0^2+\lambda)$, and $\Denfam'(0)=0+2\pi_{12}(0)\pi_{12}'(0)+0=-2\alpha_0^2(R_0\sin\phi+\iota\alpha_0)$.
\item Quotient rule: $\mathcal C_\lambda'(0)=\frac{\Numfam'(0)\Denfam(0)-\Numfam(0)\Denfam'(0)}{\Denfam(0)^2}=\frac{2\alpha_0^6(R_0\sin\phi+\iota\alpha_0)}{\alpha_0^4(2\alpha_0^2+\lambda)^2}$, which is \eqref{eq:slope}. Finally $R_0\sin\phi=\sqrt{R_0^2-\alpha_0^2}$ by Step~1.
\end{steps}
\end{proof}

\begin{corollary}[specialisations]\label{cor:special}\MkOursChk
In the family of Definition~\ref{def:family} (so $0<\alpha_0<R_0$):
\begin{enumerate}[label=(\arabic*)]
\item ($\lambda=0$, $\iota=0$): $\mathcal C_0'(0)=\dfrac{R_0\sin\phi}{2\alpha_0^2}=\dfrac{R_0\sqrt{1-\alpha_0^2/R_0^2}}{2\alpha_0^2}$.
\item ($\lambda>0$, $\alpha_0=\sqrt\lambda$, $\iota=0$, $R_0>\sqrt\lambda$): $\mathcal C_\lambda'(0)=\dfrac{2\lambda R_0\sin\phi}{9\lambda^2}=\dfrac{2R_0}{9\lambda}\sqrt{1-\lambda/R_0^2}$; and $\|W^\star\|_{\op}\le(R_0^2+\lambda)^{1/2}$.
\item (\textsf{F1}: $\lambda>0$, $\alpha_0^2=\lambda/2$, $\iota=0$, $R_0^2>\lambda/2$): $\mathcal C_\lambda'(0)=\dfrac{\lambda R_0\sin\phi}{4\lambda^2}=\dfrac{R_0}{4\lambda}\sqrt{1-\dfrac{\lambda}{2R_0^2}}$.
\item (\textsf{F2}: $\lambda>0$, $\alpha_0^2=\lambda/2$, $\iota=1$, $R_0^2>\lambda/2$): $\mathcal C_\lambda'(0)=\dfrac{R_0}{4\lambda}\sqrt{1-\dfrac{\lambda}{2R_0^2}}+\dfrac{\lambda\sqrt{\lambda/2}}{4\lambda^2}=\dfrac{R_0}{4\lambda}\sqrt{1-\dfrac{\lambda}{2R_0^2}}+\dfrac{1}{4\sqrt2\sqrt\lambda}$.
\end{enumerate}
\end{corollary}
\begin{proof}
\begin{steps}
\item In each case the hypotheses give $0<\alpha_0<R_0$, so Lemma~\ref{lem:family} applies, and $R_0\sin\phi=R_0\sqrt{1-\alpha_0^2/R_0^2}$ by \eqref{eq:slope}.
\item (1): put $\lambda=0$, $\iota=0$ in \eqref{eq:slope}: $2\alpha_0^2R_0\sin\phi/(2\alpha_0^2)^2=R_0\sin\phi/(2\alpha_0^2)$.
\item (2): put $\alpha_0^2=\lambda$, $\iota=0$: $2\lambda R_0\sin\phi/(3\lambda)^2$, and $\sqrt{1-\alpha_0^2/R_0^2}=\sqrt{1-\lambda/R_0^2}$. The norm bound is Lemma~\ref{lem:family}(ii) with $\alpha_0^2=\lambda$.
\item (3), (4): put $\alpha_0^2=\lambda/2$, so $(2\alpha_0^2+\lambda)^2=4\lambda^2$ and $2\alpha_0^2=\lambda$: $\mathcal C_\lambda'(0)=\lambda(R_0\sin\phi+\iota\sqrt{\lambda/2})/(4\lambda^2)$, with $\sqrt{1-\alpha_0^2/R_0^2}=\sqrt{1-\lambda/(2R_0^2)}$. For $\iota=1$ the extra term is $\lambda\sqrt{\lambda/2}/(4\lambda^2)=1/(4\sqrt2\sqrt\lambda)$.
\end{steps}
\end{proof}

\section{T1: where feature dimensionality is discontinuous}\label{sec:T1}

Elhage's $D_i$ sets an unlearned feature ($W_i=0$) to $0$. The theorem below says that this convention makes $D_i$ and the sum rule $\Dtot$ jump by exactly $1$ at every zero column that has a direction orthogonal to all other columns, and nowhere else. Away from such points $D_i$ is locally Lipschitz, but its modulus blows up like $R/\|W_i\|^2$ as the column shrinks, and this order cannot be improved. So a small feature column can carry a measurement error of order one.

\begin{theorem}[C1: continuity and discontinuity of $D_i$]\label{thm:C1}\MkOursChk
Let $d\ge1$, $n\ge2$, $i\in[n]$, and
\[
Z_i=\{W\in\R^{d\times n}:\ W_i=0\ \text{and}\ \nu_{\min}(S_{-i}(W))=0\}.
\]
\begin{enumerate}[label=(\alph*)]
\item $D_i$ is continuous at every $W\notin Z_i$.
\item For every $W\in Z_i$ there are $W^{(\eta)}\to W$ ($\eta\downarrow0$) with $D_i(W^{(\eta)})=1$ and $\Dtot(W^{(\eta)})=\Dtot(W)+1$, while $D_i(W)=0$. So $D_i$ and $\Dtot$ are discontinuous at every point of $Z_i$. If $n-1<d$, then $Z_i=\{W:W_i=0\}$.
\item Let $W_i\ne0$, let $W'$ be arbitrary and $R=\max(\|W\|_{\op},\|W'\|_{\op})$. Then
\begin{equation}\label{eq:C1c}
|D_i(W')-D_i(W)|\le\Big(\frac{4}{\|W_i\|}+\frac{4R}{\|W_i\|^2}\Big)\|W'-W\|_{\op} .
\end{equation}
\item In the family of Definition~\ref{def:family} with $\lambda=0$ and $\iota=0$ (so $W^\star_1=\alpha_0(\cos\phi,\sin\phi)^\top$, $W^\star_2=(R_0,0)^\top$, $E^{(0)}_1=(-\sin\phi,\cos\phi)^\top$, $E^{(0)}_2=0$, $\|E^{(0)}\|_{\op}=1$), $\|W^\star\|_{\op}\le(R_0^2+\alpha_0^2)^{1/2}$ and
\[
\Big|\tfrac{\mathrm d}{\mathrm d\eta}D_1(W^\star+\eta E^{(0)})\big|_{\eta=0}\Big|=\frac{R_0\sqrt{1-\alpha_0^2/R_0^2}}{2\alpha_0^2}.
\]
Moreover, let $L_a,L_c\in\R$, and suppose that
\begin{equation}\label{eq:C1dform}
|D_1(W')-D_1(W)|\le\Big(\frac{L_a}{\|W_1\|}+\frac{L_cR}{\|W_1\|^2}\Big)\|W'-W\|_{\op}
\end{equation}
for all $W,W'\in\R^{2\times2}$ with $W_1\ne0$ and $\|W'-W\|_{\op}\le\|W_1\|/2$ ($L_a$ and $L_c$ do not depend on $W,W'$). Then $L_c\ge1/2$. In particular, every bound of the form \eqref{eq:C1dform} for column $i$ that is required to hold, with constants independent of $W,W'$, on a class of $(d,n,i,W,W')$ containing these $2\times2$ pairs with $i=1$, has $L_c\ge1/2$.
\end{enumerate}
\end{theorem}

\begin{proof}[Proof of (a)]
Tool used: Tool~\ref{tool:rayleigh}; Lemmas~\ref{lem:P1}, \ref{lem:P2}, \ref{lem:P4}.
\begin{steps}
\item \emph{Case $W_i\ne0$.} On the open set $\{W':W'_i\ne0\}$ (P1.1), $D_i=\|W'_i\|^4/(W_i'^\top S'W'_i)$ is a quotient of polynomials in the entries, with denominator $\ge\|W'_i\|^4>0$ (Lemma~\ref{lem:P4}); it is continuous there (P1.2).
\item \emph{Case $W_i=0$, $\nu_0:=\nu_{\min}(S_{-i}(W))\ne0$.} Then $\nu_0>0$ (Lemma~\ref{lem:P2} and Tool~\ref{tool:rayleigh}), and $D_i(W)=0$.
\item For symmetric $B,B'$: $\nu_{\min}(B')\ge\nu_{\min}(B)-\|B'-B\|_{\op}$. Indeed for $\|z\|=1$, $z^\top B'z=z^\top Bz+z^\top(B'-B)z\ge\nu_{\min}(B)-\|B'-B\|_{\op}$ (Tool~\ref{tool:rayleigh}, P1.5); minimise over $z$ (Tool~\ref{tool:rayleigh}).
\item The entries of $S_{-i}(W')$ are polynomials in the entries of $W'$, so $\|S_{-i}(W')-S_{-i}(W)\|_{\op}\le\|S_{-i}(W')-S_{-i}(W)\|_F\to0$ as $W'\to W$ (P1.2, P1.3). Hence there is $\varepsilon_0>0$ with $\nu_{\min}(S_{-i}(W'))\ge\nu_0/2$ whenever $\|W'-W\|_{\op}<\varepsilon_0$ (Step~3).
\item For such $W'$ with $W'_i\ne0$: $D_i(W')=\frac{\|W'_i\|^4}{\|W'_i\|^4+W_i'^\top S_{-i}(W')W'_i}\le\frac{\|W'_i\|^4}{(\nu_0/2)\|W'_i\|^2}=\frac{2\|W'_i\|^2}{\nu_0}$ (Lemma~\ref{lem:P4}; Tool~\ref{tool:rayleigh} applied to the unit vector $W'_i/\|W'_i\|$ and multiplied by $\|W'_i\|^2$; Step~4). For $W'_i=0$, $D_i(W')=0$.
\item By P1.1, $\|W'_i\|=\|W'_i-W_i\|\le\|W'-W\|_{\op}$, so $|D_i(W')-D_i(W)|\le2\|W'-W\|_{\op}^2/\nu_0\to0$.
\end{steps}
\end{proof}

\begin{proof}[Proof of (b)]
\begin{steps}
\item Let $W\in Z_i$. By Tool~\ref{tool:rayleigh} there is a unit $z$ with $z^\top S_{-i}(W)z=\nu_{\min}=0$, i.e.\ $\sum_{j\ne i}\langle W_j,z\rangle^2=0$ (Lemma~\ref{lem:P2}), so $\langle W_j,z\rangle=0$ for all $j\ne i$.
\item For $\eta>0$ let $W^{(\eta)}$ be $W$ with column $i$ replaced by $\eta z$. Then $\|W^{(\eta)}-W\|_{\op}\le\|W^{(\eta)}-W\|_F=\eta\to0$ (P1.3).
\item $D_i(W^{(\eta)})$: the column $\eta z\ne0$ has $\langle\widehat{\eta z},W^{(\eta)}_j\rangle=\langle z,W_j\rangle=0$ for $j\ne i$ and $\langle\widehat{\eta z},\eta z\rangle^2=\eta^2$, so by D2 $D_i(W^{(\eta)})=\eta^2/\eta^2=1$. And $D_i(W)=0$ (D2, $W_i=0$).
\item For $j\ne i$ with $W_j\ne0$: in the denominator of D2 for $D_j(W^{(\eta)})$, the term with index $i$ is $\langle\hat W_j,\eta z\rangle^2=0=\langle\hat W_j,W_i\rangle^2$, and all other terms are those of $D_j(W)$. So $D_j(W^{(\eta)})=D_j(W)$; for $W_j=0$ both are $0$.
\item Hence $\Dtot(W^{(\eta)})=1+\sum_{j\ne i}D_j(W)=\Dtot(W)+1$ (using $D_i(W)=0$). Since $D_i(W^{(\eta)})=1\not\to0=D_i(W)$, $D_i$ and $\Dtot$ are discontinuous at $W$.
\item If $n-1<d$: the $(n-1)\times d$ matrix with rows $W_j^\top$, $j\ne i$, has rank $\le n-1$, so its nullspace has dimension $\ge d-(n-1)\ge1$ (Tool~\ref{tool:dim}); take a unit $z$ in it. Then $z^\top S_{-i}z=0$, so $\nu_{\min}(S_{-i})\le0$ (Tool~\ref{tool:rayleigh}), and $\ge0$ (Lemma~\ref{lem:P2}, Tool~\ref{tool:rayleigh}). Hence every $W$ with $W_i=0$ lies in $Z_i$; the reverse inclusion is the definition.
\end{steps}
\end{proof}

\begin{proof}[Proof of (c)]
Here $\mathsf h=\|W'-W\|_{\op}$ and $\mathsf a=\|W_i\|>0$ (Definition~\ref{def:D1}).
\begin{steps}
\item \emph{Large perturbations.} If $\mathsf h\ge\mathsf a$, the right side of \eqref{eq:C1c} is $\ge4$, while $D_i(W),D_i(W')\in[0,1]$ (D2, \eqref{eq:Dform}); done. Assume $\mathsf h<\mathsf a$.
\item Then $\|W'_i\|\ge\mathsf a-\|W'_i-W_i\|\ge\mathsf a-\mathsf h>0$ (P1.1). With $\mathsf b=\|W^\top\hat W_i\|$, $\mathsf a'=\|W'_i\|$, $\mathsf b'=\|W'^\top\hat W'_i\|$, \eqref{eq:Dform} gives $D_i(W)=(\mathsf a/\mathsf b)^2$, $D_i(W')=(\mathsf a'/\mathsf b')^2$, $\mathsf b\ge\mathsf a$, $\mathsf b'\ge\mathsf a'$.
\item $|D_i(W)-D_i(W')|=\big|\tfrac{\mathsf a}{\mathsf b}-\tfrac{\mathsf a'}{\mathsf b'}\big|\,\big(\tfrac{\mathsf a}{\mathsf b}+\tfrac{\mathsf a'}{\mathsf b'}\big)\le2\big|\tfrac{\mathsf a}{\mathsf b}-\tfrac{\mathsf a'}{\mathsf b'}\big|$ (both ratios in $(0,1]$).
\item $\tfrac{\mathsf a}{\mathsf b}-\tfrac{\mathsf a'}{\mathsf b'}=\tfrac{\mathsf a-\mathsf a'}{\mathsf b}+\tfrac{\mathsf a'}{\mathsf b'}\cdot\tfrac{\mathsf b'-\mathsf b}{\mathsf b}$ (algebra), so its absolute value is $\le(|\mathsf a-\mathsf a'|+|\mathsf b-\mathsf b'|)/\mathsf b\le(|\mathsf a-\mathsf a'|+|\mathsf b-\mathsf b'|)/\mathsf a$ ($\mathsf a'/\mathsf b'\le1$, $\mathsf b\ge\mathsf a$).
\item $|\mathsf a-\mathsf a'|\le\|W_i-W'_i\|\le\mathsf h$ (triangle inequality, P1.1).
\item $\|\hat W_i-\hat W'_i\|\le2\mathsf h/\mathsf a$: indeed $\hat W_i-\hat W'_i=\frac{W_i-W'_i}{\mathsf a}+W'_i\big(\frac1{\mathsf a}-\frac1{\mathsf a'}\big)$, and the second term has norm $|\mathsf a'-\mathsf a|/\mathsf a\le\mathsf h/\mathsf a$ (Step~5).
\item $|\mathsf b-\mathsf b'|\le\|W^\top\hat W_i-W'^\top\hat W'_i\|\le\|(W-W')^\top\hat W_i\|+\|W'^\top(\hat W_i-\hat W'_i)\|\le\mathsf h+R\cdot2\mathsf h/\mathsf a$ (triangle inequality, P1.4 twice, Step~6).
\item Steps~3--7: $|D_i(W)-D_i(W')|\le\frac{2}{\mathsf a}\big(2\mathsf h+\frac{2R\mathsf h}{\mathsf a}\big)=\big(\frac4{\mathsf a}+\frac{4R}{\mathsf a^2}\big)\mathsf h$.
\end{steps}
\end{proof}

\begin{proof}[Proof of (d)]
\begin{steps}
\item The example is the family of Definition~\ref{def:family} with $\lambda=0$, $\iota=0$. By Lemma~\ref{lem:family}(i) and Corollary~\ref{cor:special}(1), $\frac{\mathrm d}{\mathrm d\eta}D_1(W^\star+\eta E^{(0)})|_{\eta=0}=R_0\sqrt{1-\alpha_0^2/R_0^2}/(2\alpha_0^2)$, and $\|E^{(0)}\|_{\op}=1$, $\|W^\star\|_{\op}\le(R_0^2+\alpha_0^2)^{1/2}$ by Lemma~\ref{lem:family}(ii).
\item Suppose \eqref{eq:C1dform} holds on the stated class. Put $L_c^+:=\max(L_c,0)\ge L_c$. Since $R\ge0$ and $\|W_1\|>0$, the right side of \eqref{eq:C1dform} does not decrease when $L_c$ is replaced by $L_c^+$, so \eqref{eq:C1dform} also holds with $(L_a,L_c^+)$ on the same class.
\item Fix $0<\alpha_0<R_0$. For $0<|\eta|\le\alpha_0/2$, the pair $W=W^\star$, $W'=W^\star+\eta E^{(0)}$ is in the class: $\|W^\star_1\|=\alpha_0>0$ and $\|W'-W\|_{\op}=|\eta|\,\|E^{(0)}\|_{\op}=|\eta|\le\alpha_0/2$ (Step~1). Its $R$ satisfies $R\le\|W^\star\|_{\op}+|\eta|\le(R_0^2+\alpha_0^2)^{1/2}+|\eta|$ (Tool~\ref{tool:op}, Step~1). Since $L_c^+\ge0$, Step~2 gives
\[
\frac{|D_1(W^\star+\eta E^{(0)})-D_1(W^\star)|}{|\eta|}\le\frac{L_a}{\alpha_0}+\frac{L_c^+\big((R_0^2+\alpha_0^2)^{1/2}+|\eta|\big)}{\alpha_0^2}.
\]
\item Let $\eta\to0$. The left side tends to $|\mathcal C_0'(0)|=\sqrt{R_0^2-\alpha_0^2}/(2\alpha_0^2)$ (Step~1; $\mathcal C_0$ is differentiable by Lemma~\ref{lem:family}(iii)), and the right side tends to $L_a/\alpha_0+L_c^+(R_0^2+\alpha_0^2)^{1/2}/\alpha_0^2$. So $\frac{\sqrt{R_0^2-\alpha_0^2}}{2\alpha_0^2}\le\frac{L_a}{\alpha_0}+\frac{L_c^+(R_0^2+\alpha_0^2)^{1/2}}{\alpha_0^2}$.
\item Multiply by $\alpha_0^2/R_0>0$: $\frac12\sqrt{1-\alpha_0^2/R_0^2}\le\frac{L_a\alpha_0}{R_0}+L_c^+\sqrt{1+\alpha_0^2/R_0^2}$. The constants $L_a,L_c^+$ do not depend on $R_0$, so we may let $R_0\to\infty$ with $\alpha_0$ fixed: $\frac12\le L_c^+$.
\item Since $L_c^+\ge1/2>0$, $L_c^+=\max(L_c,0)$ equals $L_c$. Hence $L_c\ge1/2$. The last sentence of (d) follows, because such a bound holds in particular on the $2\times2$ class with $i=1$.
\end{steps}
\end{proof}

\begin{remark}[a weaker form of (c), and the bracket for the coefficient]\label{rem:C1c}\MkOursChk
Let $\mathcal N$ be the class of all $(d,n,i,W,W')$ with $d\ge1$, $n\ge2$, $i\in[n]$, $W,W'\in\R^{d\times n}$, $W_i\ne0$ and $\|W'-W\|_{\op}\le\|W_i\|/2$, and let
\begin{equation}\label{eq:Lcbest}
\begin{aligned}
L_c^{\mathrm{best}}:=\inf\Big\{L_c\in\R:\ &\exists L_a\in\R\ \text{such that}\\
&|D_i(W')-D_i(W)|\le\Big(\frac{L_a}{\|W_i\|}+\frac{L_cR}{\|W_i\|^2}\Big)\|W'-W\|_{\op}\ \text{on all of}\ \mathcal N\Big\}.
\end{aligned}
\end{equation}
Then (1) on $\mathcal N$, $|D_i(W')-D_i(W)|\le\big(\frac{6}{\|W_i\|}+\frac{4R}{\|W_i\|^2}\big)\|W'-W\|_{\op}$; and (2) $1/2\le L_c^{\mathrm{best}}\le4$. Whether the infimum is attained, and its value, are not determined here.
\end{remark}
\begin{proof}
\begin{steps}
\item (1): every element of $\mathcal N$ satisfies the hypotheses of (c), which hold for all $W'$, in particular when $\|W'-W\|_{\op}\le\|W_i\|/2$. So \eqref{eq:C1c} holds on $\mathcal N$, and $4/\|W_i\|\le6/\|W_i\|$.
\item Upper end: by Step~1, \eqref{eq:C1c} holds on $\mathcal N$ with $(L_a,L_c)=(4,4)$, so $4$ belongs to the set in the definition; hence the set is nonempty and $L_c^{\mathrm{best}}\le4$.
\item Lower end: $\mathcal N$ contains all $(2,2,1,W,W')$ with $W_1\ne0$ and $\|W'-W\|_{\op}\le\|W_1\|/2$. So every $L_c$ in the set, with its $L_a$, satisfies the hypothesis of (d), and (d) gives $L_c\ge1/2$. Hence $L_c^{\mathrm{best}}\ge1/2$.
\end{steps}
\end{proof}

\begin{remark}[a remark in the literature]\label{rem:scherlis}\MkOurs
Scherlis et al.\ \cite{scherlis2022} (Sec.~2.1) remark that $W_i\approx0$ gives $D_i\approx0$ ``in practice'' (only these two words are quoted). Their remark is informal; the precise statement that fails is the implication ``$W^{(\eta)}\to W$ in $\|\cdot\|_{\op}$ with $W_i=0$ implies $D_i(W^{(\eta)})\to0$''. It fails at every point of $Z_i$.
\end{remark}
\begin{proof}
\begin{steps}
\item Let $W\in Z_i$. By (b) there are $W^{(\eta)}\to W$ in $\|\cdot\|_{\op}$ ($\eta\downarrow0$) with $D_i(W^{(\eta)})=1$ for every $\eta>0$, while $W_i=0$.
\item By Step~2 of the proof of (b), the column $i$ of $W^{(\eta)}$ is $\eta z$ with $z$ a unit vector orthogonal to all other columns; its norm $\eta$ tends to $0$. So arbitrarily small columns, orthogonal to the others, have $D_i=1\not\to0$.
\end{steps}
\end{proof}

\section{T2: stable metrics}\label{sec:T2}

This section replaces the discontinuous $D_i$ by quantities that are Lipschitz in $W$ for every fixed ridge parameter $\lambda>0$. Ridge leverage $\ell_i$ and the effective dimension $\deff$ are Lipschitz with the sharp, scale-free constant $3\sqrt3/(8\sqrt\lambda)$. The regularised dimensionality $\Dt_i$, which is the closest analogue of $D_i$, is also Lipschitz, but its constant must grow like $R/(4\lambda)$, and we determine the coefficient $1/4$ exactly. By D6, every result holds for each fixed $\lambda$ in the range $\Lambda(W)$ used in \S\ref{sec:T5}.

\subsection{Ridge leverage: range}\label{ssec:C4a}

\begin{proposition}[C4(a), range]\label{prop:C4a}\MkOursChk
For all $d,n\ge1$, $\lambda>0$, $i\in[n]$ and $W\in\R^{d\times n}$,
\begin{equation}\label{eq:range}
0\le\ell_i(W;\lambda)<1,\qquad\text{and }\ell_i(W;\lambda)=0\iff W_i=0 .
\end{equation}
\end{proposition}
\begin{proof}
Used: Lemma~\ref{lem:P2}; D1, D4.
\begin{steps}
\item If $W_i=0$ then $\ell_i=0$ (D4), so \eqref{eq:range} holds. Assume $W_i\ne0$ from now on.
\item Put $z:=S_\lambda^{-1}W_i$ (exists by Lemma~\ref{lem:P2}). Then $z\ne0$, since $S_\lambda z=W_i\ne0$.
\item $\ell_i=W_i^\top z=(S_\lambda z)^\top z=z^\top S_\lambda z\ge\lambda\|z\|^2>0$ (D4, symmetry of $S_\lambda$, Lemma~\ref{lem:P2}, Step~2). This gives $\ell_i>0$, hence the ``$\iff$''.
\item By D1, $S_\lambda=S_{-i,\lambda}+W_iW_i^\top$, and $z^\top S_{-i,\lambda}z=z^\top S_{-i}z+\lambda\|z\|^2\ge\lambda\|z\|^2>0$ ($z^\top S_{-i}z\ge0$ by Lemma~\ref{lem:P2}; $z\ne0$ by Step~2).
\item From $S_\lambda z=W_i$ and Step~4: $S_{-i,\lambda}z+W_i(W_i^\top z)=W_i$, that is $S_{-i,\lambda}z=(1-\ell_i)W_i$ (Step~3: $W_i^\top z=\ell_i$).
\item Multiply Step~5 on the left by $z^\top$: $z^\top S_{-i,\lambda}z=(1-\ell_i)\,z^\top W_i=(1-\ell_i)\ell_i$.
\item By Steps~4 and 6, $(1-\ell_i)\ell_i>0$. Since $\ell_i>0$ (Step~3), $1-\ell_i>0$.
\end{steps}
\end{proof}

\subsection{The scalar case}\label{ssec:scalar}

For $d=n=1$ everything reduces to one function of one variable. Its maximal slope, $3\sqrt3/(8\sqrt\lambda)$ at $w^2=\lambda/3$, turns out to be the sharp constant in every dimension.

\begin{definition}[scalar leverage]\label{def:scalar}\MkDef
For $\lambda>0$ let $\lsc(w)=w^2/(w^2+\lambda)$ for $w\in\R$, $w_0=\sqrt{\lambda/3}$ and $\Lsc(\lambda)=3\sqrt3/(8\sqrt\lambda)$. For $u\ge0$ let $\Fsc(u)=2u/(1+u^2)^2$. If $d=n=1$ and $W=(w)$, then $S=w^2$, $\ell_1(W;\lambda)=w(w^2+\lambda)^{-1}w=\lsc(w)$, $\deff(W;\lambda)=\lsc(w)$ (D4, $n=1$), and $\|W-W'\|_{\op}=\|W-W'\|_*=\|W-W'\|_F=|w-w'|$ (the single singular value of a $1\times1$ matrix $(y)$ is $|y|$, and $\|(y)\|_F=|y|$ by D0).
\end{definition}

\begin{proposition}[C4(b) and C5 for $d=n=1$, and sharpness]\label{prop:scalar}\MkOurs
Let $\lambda>0$.
\begin{enumerate}[label=(\roman*)]
\item For all $w,w'\in\R$: $|\lsc(w)-\lsc(w')|\le\Lsc(\lambda)\,|w-w'|$.
\item For $w\ne w'$ the inequality in (i) is strict.
\item $\lim_{\eta\to0}|\lsc(w_0+\eta)-\lsc(w_0)|/|\eta|=\Lsc(\lambda)$, where the point $W=(w_0)$ has $\|W\|_{\op}/\sqrt\lambda=1/\sqrt3$.
\end{enumerate}
Consequently: the inequality of C4(b) holds for $d=n=1$ with least constant $\Lsc(\lambda)$, and the ratio $|\lsc(w)-\lsc(w')|/|w-w'|$ has supremum $\Lsc(\lambda)$, attained by no pair; the number $L^\star(\lambda)$ of Theorem~\ref{thm:C4b} satisfies $L^\star(\lambda)\ge\Lsc(\lambda)$; and the same statements hold for $\deff$ with $\|\cdot\|_*$ and with $\|\cdot\|_F$. This proposition proves only the sharpness clause of Theorem~\ref{thm:C5}, not its inequalities.
\end{proposition}
\begin{proof}
Tools used: Tool~\ref{tool:calc}; D4.
\begin{steps}
\item \emph{First derivative} (quotient rule): $(\lsc)'(w)=\dfrac{2w(w^2+\lambda)-w^2\cdot2w}{(w^2+\lambda)^2}=\dfrac{2\lambda w}{(w^2+\lambda)^2}$.
\item \emph{Second derivative} (quotient rule on Step~1): $(\lsc)''(w)=2\lambda\dfrac{(w^2+\lambda)^2-w\cdot2(w^2+\lambda)\cdot2w}{(w^2+\lambda)^4}=\dfrac{2\lambda(\lambda-3w^2)}{(w^2+\lambda)^3}$.
\item \emph{Shape.} By Step~1, $(\lsc)'$ is odd, so $|(\lsc)'|$ is even, and $(\lsc)'\ge0$ on $[0,\infty)$ with value $0$ at $0$. By Step~2, $(\lsc)''>0$ on $(0,w_0)$ and $<0$ on $(w_0,\infty)$; so $(\lsc)'$ is strictly increasing on $[0,w_0]$ and strictly decreasing on $[w_0,\infty)$ (Tool~\ref{tool:calc}(d)).
\item \emph{Maximum.} By Step~3, $|(\lsc)'(w)|\le(\lsc)'(w_0)$ for all $w$, with equality only at $w=\pm w_0$.
\item \emph{Value.} $w_0^2+\lambda=4\lambda/3$, so $(w_0^2+\lambda)^2=16\lambda^2/9$ and $(\lsc)'(w_0)=2\lambda w_0\cdot\frac{9}{16\lambda^2}=\frac{9w_0}{8\lambda}=\frac{9}{8\sqrt3\sqrt\lambda}=\frac{3\sqrt3}{8\sqrt\lambda}=\Lsc(\lambda)$ (using $w_0=\sqrt\lambda/\sqrt3$ and $9/\sqrt3=3\sqrt3$).
\item \emph{(i).} For $w\ne w'$, the mean value theorem (Tool~\ref{tool:calc}(b)) gives $\bar w$ between them with $\lsc(w)-\lsc(w')=(\lsc)'(\bar w)(w-w')$; by Steps~4--5, $|(\lsc)'(\bar w)|\le\Lsc(\lambda)$. For $w=w'$ there is nothing to prove.
\item \emph{(ii).} Let $w<w'$ (the other order is symmetric). By the fundamental theorem of calculus (Tool~\ref{tool:calc}(c); $(\lsc)'$ is continuous) and Tool~\ref{tool:calc}(g), $|\lsc(w')-\lsc(w)|=\big|\int_w^{w'}(\lsc)'(u)\,\mathrm du\big|\le\int_w^{w'}|(\lsc)'(u)|\,\mathrm du$. The function $\mathsf f(u):=\Lsc(\lambda)-|(\lsc)'(u)|$ is continuous, $\ge0$ (Step~4), and $>0$ except at $u=\pm w_0$; $[w,w']$ contains a point other than $\pm w_0$, so $\int_w^{w'}\mathsf f>0$ (Tool~\ref{tool:calc}(f)). By linearity and $\int_w^{w'}\Lsc(\lambda)\,\mathrm du=\Lsc(\lambda)(w'-w)$ (Tool~\ref{tool:calc}(g)), $\int_w^{w'}|(\lsc)'|<\Lsc(\lambda)(w'-w)$.
\item \emph{(iii).} By the definition of the derivative and Step~5, the difference quotient at $w_0$ tends to $(\lsc)'(w_0)=\Lsc(\lambda)>0$, so its absolute value tends to $\Lsc(\lambda)$.
\item \emph{Least constant for $d=n=1$.} By (i), $\Lsc(\lambda)$ works. If $L<\Lsc(\lambda)$, then by (iii) some $\eta\ne0$ gives a quotient $>L$, with $W=(w_0)$, $W'=(w_0+\eta)$; so $L$ fails. By (ii) no pair attains $\Lsc(\lambda)$.
\item \emph{$L^\star(\lambda)\ge\Lsc(\lambda)$.} $L^\star(\lambda)$ is the infimum of the constants valid for all $d,n\ge1$ (Theorem~\ref{thm:C4b}). Each such constant is valid for $d=n=1$, hence $\ge\Lsc(\lambda)$ by Step~9; so the infimum is $\ge\Lsc(\lambda)$ (also if the set of valid constants is empty, when the infimum is $+\infty$).
\item \emph{C5.} For $d=n=1$, $\deff=\lsc(w)$ and $\|W-W'\|_*=\|W-W'\|_F=|w-w'|$ (Definition~\ref{def:scalar}), so Steps~1--9 apply verbatim to $\deff$ with the nuclear or the Frobenius norm. Since $\sqrt{\min(d,n)}=1$ for $d=n=1$, the number $3\sqrt3/8$ cannot be replaced by a smaller one in either inequality of \eqref{eq:C5}.
\end{steps}
\end{proof}

\subsection{The sharp Lipschitz constant of ridge leverage}\label{ssec:C4b}

The next theorem says that the scalar slope is the worst case in every dimension: no matrix shape, no size of $W$ and no interaction between columns makes ridge leverage more sensitive than in the $1\times1$ case. The proof bounds the derivative along the segment from $W$ to $W'$ and reduces it to a maximisation over two numbers. Appendix~\ref{app:B4} gives an independent proof by a variational argument.

\begin{theorem}[C4(b): sharp Lipschitz constant of ridge leverage]\label{thm:C4b}\MkOursChk
For $\lambda>0$ let $L^\star(\lambda)\in[0,\infty]$ be the infimum of the set of constants $L\ge0$ such that $|\ell_i(W;\lambda)-\ell_i(W';\lambda)|\le L\|W-W'\|_{\op}$ for all $d,n\ge1$, $i\in[n]$ and $W,W'\in\R^{d\times n}$ (the infimum of the empty set being $+\infty$). For all $d,n\ge1$, $\lambda>0$, $i\in[n]$ and $W,W'\in\R^{d\times n}$,
\begin{equation}\label{eq:A4main}
|\ell_i(W;\lambda)-\ell_i(W';\lambda)|\le\frac{3\sqrt3}{8\sqrt\lambda}\,\|W-W'\|_{\op} .
\end{equation}
Together with the case $d=n=1$ (Proposition~\ref{prop:scalar}), $L^\star(\lambda)=3\sqrt3/(8\sqrt\lambda)$, and this infimum is itself a valid constant, i.e.\ the least one. No hypothesis on $W_i$ (for instance $W_i\ne0$) and no bound on $\|W\|_{\op}$ is needed.
\end{theorem}

\begin{proof}
Fix $W,W'$, and use $E$, $W(\eta)$, $S(\eta)$, $S_\lambda(\eta)$, $K_\lambda(\eta)$ of D1. Put $\Theta(\eta):=\lambda K_\lambda(\eta)^{-1}$ (the inverse exists by Step~3), $z(\eta):=S_\lambda(\eta)^{-1}W_i(\eta)\in\R^d$,
\[
\Gamma_1(\eta):=\mathsf e_i^\top\Theta(\eta)\mathsf e_i,\qquad \Gamma_2(\eta):=\|\Theta(\eta)\mathsf e_i\|^2 ,
\]
with $\mathsf e_i\in\R^n$, so that $W_i=W\mathsf e_i$. Where no confusion arises we drop the argument $\eta$. Tools used: Tools~\ref{tool:spec}, \ref{tool:op}, \ref{tool:cramer}, \ref{tool:calc}, \ref{tool:cs}.
\begin{steps}
\item \emph{$S_\lambda(\eta)^{-1}$ exists and is smooth in $\eta$.} For $x\in\R^d$, $x^\top S_\lambda(\eta)x=\|W(\eta)^\top x\|^2+\lambda\|x\|^2\ge\lambda\|x\|^2$. The matrix $S_\lambda(\eta)$ is symmetric, so by Tool~\ref{tool:spec} (with $x$ a unit eigenvector) all its eigenvalues are $\ge\lambda>0$; it is invertible and $S_\lambda(\eta)^{-1}$ is symmetric. Its entries are polynomials in $\eta$ (of degree at most 2), and its determinant is a polynomial in $\eta$ with no zero on $\R$. By Tool~\ref{tool:cramer} every entry of $S_\lambda(\eta)^{-1}$ is a quotient of polynomials in $\eta$ with a nowhere-vanishing denominator, hence infinitely differentiable in $\eta$.
\item \emph{Derivative of $S_\lambda^{-1}$.} Differentiate the identity $S_\lambda(\eta)S_\lambda(\eta)^{-1}=I_d$ with the product rule (Tool~\ref{tool:calc}(a); legitimate by Step~1): $S'(\eta)S_\lambda^{-1}+S_\lambda(S_\lambda^{-1})'=0$, so $(S_\lambda^{-1})'=-S_\lambda^{-1}S'S_\lambda^{-1}$. Here $S'(\eta)=\frac{\mathrm d}{\mathrm d\eta}\big(W(\eta)W(\eta)^\top\big)=EW(\eta)^\top+W(\eta)E^\top$.
\item \emph{Push-through identity.} For $x\in\R^n$, $x^\top K_\lambda(\eta)x=\|W(\eta)x\|^2+\lambda\|x\|^2\ge\lambda\|x\|^2$, so as in Step~1, $K_\lambda(\eta)$ is symmetric with all eigenvalues $\ge\lambda$, and invertible. Next, $S_\lambda W=WW^\top W+\lambda W=W(W^\top W+\lambda I_n)=WK_\lambda$. Multiplying on the left by $S_\lambda^{-1}$ and on the right by $K_\lambda^{-1}$ gives $S_\lambda^{-1}W=WK_\lambda^{-1}$.
\item \emph{Derivative of $\ell_i$.} Write $\ell_i(\eta)$ for $\ell_i(W(\eta);\lambda)=W_i(\eta)^\top S_\lambda(\eta)^{-1}W_i(\eta)$. By Step~1 it is continuously differentiable. By the product rule, Step~2, $W_i'(\eta)=E_i$ and the symmetry of $S_\lambda^{-1}$,
\begin{align*}
\ell_i'(\eta)&=2E_i^\top S_\lambda^{-1}W_i-W_i^\top S_\lambda^{-1}S'S_\lambda^{-1}W_i\\
&=2\langle E_i,z\rangle-z^\top\big(EW^\top+WE^\top\big)z=2\langle E_i,z\rangle-2\langle E^\top z,W^\top z\rangle .
\end{align*}
Since $E_i=E\mathsf e_i$, $\langle E_i,z\rangle=\langle\mathsf e_i,E^\top z\rangle$. Hence
\begin{equation}\label{eq:A4deriv}
\ell_i'(\eta)=2\,\big\langle E^\top z,\ \mathsf e_i-W^\top z\big\rangle .
\end{equation}
\item \emph{The second factor.} By Step~3, $z=S_\lambda^{-1}W\mathsf e_i=WK_\lambda^{-1}\mathsf e_i$, so $W^\top z=W^\top WK_\lambda^{-1}\mathsf e_i=(K_\lambda-\lambda I_n)K_\lambda^{-1}\mathsf e_i=\mathsf e_i-\lambda K_\lambda^{-1}\mathsf e_i$. Hence $\mathsf e_i-W^\top z=\Theta\mathsf e_i$.
\item \emph{First bound.} By Tools~\ref{tool:cs} and \ref{tool:op} applied to \eqref{eq:A4deriv} and Step~5,
\[
|\ell_i'(\eta)|\le2\|E^\top z\|\,\|\Theta\mathsf e_i\|\le2\|E\|_{\op}\,\|z\|\,\|\Theta\mathsf e_i\| .
\]
\item \emph{The two norms in terms of $\Gamma_1,\Gamma_2$.} $\Theta$ is symmetric (Step~3 and Tool~\ref{tool:spec}), so $\|\Theta\mathsf e_i\|^2=\Gamma_2$. For $\|z\|$: by Step~5, $z=WK_\lambda^{-1}\mathsf e_i$, so $\|z\|^2=\mathsf e_i^\top K_\lambda^{-1}W^\top WK_\lambda^{-1}\mathsf e_i$. Substitute $K_\lambda^{-1}=\Theta/\lambda$ and $W^\top W=K_\lambda-\lambda I_n=\lambda\Theta^{-1}-\lambda I_n$:
\[
K_\lambda^{-1}W^\top WK_\lambda^{-1}=\lambda^{-2}\,\Theta\big(\lambda\Theta^{-1}-\lambda I_n\big)\Theta=\lambda^{-1}\big(\Theta-\Theta^2\big),
\]
so $\|z\|^2=(\Gamma_1-\Gamma_2)/\lambda$ (using $\mathsf e_i^\top\Theta^2\mathsf e_i=\|\Theta\mathsf e_i\|^2=\Gamma_2$). With Step~6,
\begin{equation}\label{eq:A4gamma}
|\ell_i'(\eta)|\le\frac{2\|E\|_{\op}}{\sqrt\lambda}\,\sqrt{(\Gamma_1-\Gamma_2)\,\Gamma_2}\,.
\end{equation}
\item \emph{Constraints on $(\Gamma_1,\Gamma_2)$.} By Step~3 the eigenvalues of $K_\lambda$ are $\ge\lambda$, so by Tool~\ref{tool:spec} those of $K_\lambda^{-1}$ lie in $(0,1/\lambda]$ and those of $\Theta=\lambda K_\lambda^{-1}$ lie in $(0,1]$; $\Theta$ is positive semidefinite with $\nu_{\max}(\Theta)\le1$. Then, with $\|\mathsf e_i\|=1$:
 \begin{itemize}
 \item $0\le\Gamma_1\le\nu_{\max}(\Theta)\le1$ (Tool~\ref{tool:spec});
 \item $\Gamma_2=\|\Theta\mathsf e_i\|^2\le\nu_{\max}(\Theta)\,\mathsf e_i^\top\Theta\mathsf e_i\le\Gamma_1$ (Tool~\ref{tool:spec}, positive semidefinite case);
 \item $\Gamma_1=\langle\mathsf e_i,\Theta\mathsf e_i\rangle\le\|\Theta\mathsf e_i\|=\Gamma_2^{1/2}$ by Tool~\ref{tool:cs}, and $\Gamma_1\ge0$, so $\Gamma_1^2\le\Gamma_2$.
 \end{itemize}
 So $(\Gamma_1,\Gamma_2)$ lies in $\mathcal T:=\{(y_1,y_2)\in\R^2:\ y_1^2\le y_2\le y_1\le1\}$.
\item \emph{Maximisation of $(y_1-y_2)y_2$ over $\mathcal T$.} Every $(y_1,y_2)\in\mathcal T$ has $y_1^2\le y_2\le y_1$, so $y_1-y_1^2\ge0$, i.e.\ $y_1\in[0,1]$. Fix $y_1\in[0,1]$ and put $\mathcal P(y_2):=(y_1-y_2)y_2$ on $[y_1^2,y_1]$. $\mathcal P$ is a concave quadratic with $\mathcal P'(y_2)=y_1-2y_2$, which vanishes only at $y_2=y_1/2$.
 \begin{itemize}
 \item If $y_1\le1/2$: then $y_1^2\le y_1/2$, so $y_1/2\in[y_1^2,y_1]$ and $\max\mathcal P=\mathcal P(y_1/2)=y_1^2/4\le1/16$.
 \item If $y_1>1/2$: then $y_1^2>y_1/2$, so $\mathcal P'<0$ on $[y_1^2,y_1]$, and $\max\mathcal P=\mathcal P(y_1^2)=y_1^3-y_1^4$. On $[1/2,1]$, $\frac{\mathrm d}{\mathrm dy_1}(y_1^3-y_1^4)=y_1^2(3-4y_1)$ is positive for $y_1<3/4$ and negative for $y_1>3/4$, so the maximum is at $y_1=3/4$, with value $27/64-81/256=27/256$.
 \end{itemize}
 Since $1/16=16/256<27/256$, $\max_{\mathcal T}(y_1-y_2)y_2=27/256$, attained only at $(y_1,y_2)=(3/4,9/16)$.
\item \emph{Pointwise bound.} By \eqref{eq:A4gamma} and Steps~8--9, for every $\eta\in\R$,
\[
|\ell_i'(\eta)|\le\frac{2\|E\|_{\op}}{\sqrt\lambda}\sqrt{\frac{27}{256}}=\frac{2\cdot3\sqrt3}{16\sqrt\lambda}\|E\|_{\op}=\frac{3\sqrt3}{8\sqrt\lambda}\|E\|_{\op} .
\]
\item \emph{Integration along the segment.} By Step~4, $\eta\mapsto\ell_i(\eta)$ is continuously differentiable on $[0,1]$, with $\ell_i(0)=\ell_i(W;\lambda)$ and $\ell_i(1)=\ell_i(W';\lambda)$. By Tool~\ref{tool:calc}(c), Tool~\ref{tool:calc}(g) and Step~10 (with $\|E\|_{\op}=\|W'-W\|_{\op}$),
\[
|\ell_i(W';\lambda)-\ell_i(W;\lambda)|=\Big|\int_0^1\ell_i'(\eta)\,\mathrm d\eta\Big|\le\int_0^1|\ell_i'(\eta)|\,\mathrm d\eta\le\frac{3\sqrt3}{8\sqrt\lambda}\|W'-W\|_{\op} .
\]
This is \eqref{eq:A4main}.
\item \emph{Sharpness.} \eqref{eq:A4main} shows that $3\sqrt3/(8\sqrt\lambda)$ belongs to the set defining $L^\star(\lambda)$, so $L^\star(\lambda)\le3\sqrt3/(8\sqrt\lambda)$. Proposition~\ref{prop:scalar} shows $L^\star(\lambda)\ge3\sqrt3/(8\sqrt\lambda)$. Hence $L^\star(\lambda)=3\sqrt3/(8\sqrt\lambda)$, and it is the least valid constant.
\end{steps}
\end{proof}

\begin{remark}[where the bound of Step~9 is attained]\label{rem:A4tight}\MkOursChk
Fix $\lambda>0$ and a matrix $W\in\R^{d\times n}$ (in the proof, $W=W(\eta)$), and let $\Theta,\Gamma_1,\Gamma_2$ be formed from $W$ as in the proof of Theorem~\ref{thm:C4b}. Then $(\Gamma_1-\Gamma_2)\Gamma_2=27/256$ (equality in Step~9) if and only if $W^\top W\mathsf e_i=(\lambda/3)\mathsf e_i$, that is, if and only if $\langle W_j,W_i\rangle=0$ for every $j\ne i$ and $\|W_i\|^2=\lambda/3$. In particular, if $E\ne0$ and $|\ell_i'(\eta)|=\frac{3\sqrt3}{8\sqrt\lambda}\|E\|_{\op}$ at some $\eta$, then $W(\eta)$ has this form. This is the scalar extremum embedded in a larger matrix. (We do not characterise equality in Step~6.)
\end{remark}
\begin{proof}
\begin{steps}
\item \emph{Forward.} Assume $(\Gamma_1-\Gamma_2)\Gamma_2=27/256$. By Step~8 of the proof of Theorem~\ref{thm:C4b}, $(\Gamma_1,\Gamma_2)\in\mathcal T$, and by Step~9 the maximum $27/256$ over $\mathcal T$ is attained only at $(3/4,9/16)$. So $\Gamma_1=3/4$ and $\Gamma_2=9/16$; in particular $\Gamma_1=\Gamma_2^{1/2}$, as $\Gamma_1\ge0$.
\item The third bullet of Step~8 is $\Gamma_1=\langle\mathsf e_i,\Theta\mathsf e_i\rangle\le\|\mathsf e_i\|\,\|\Theta\mathsf e_i\|=\Gamma_2^{1/2}$. By Step~1 it holds with equality, so by the equality case of Tool~\ref{tool:cs} the vectors $\mathsf e_i$ and $\Theta\mathsf e_i$ are linearly dependent. As $\mathsf e_i\ne0$, $\Theta\mathsf e_i=c\,\mathsf e_i$ for some real $c$, and $c=\mathsf e_i^\top\Theta\mathsf e_i=\Gamma_1=3/4$.
\item $\Theta=\lambda K_\lambda^{-1}$, so $K_\lambda\Theta=\lambda I_n$. Applying $K_\lambda$ to Step~2: $\lambda\mathsf e_i=K_\lambda\Theta\mathsf e_i=\frac34K_\lambda\mathsf e_i$, so $K_\lambda\mathsf e_i=\frac{4\lambda}3\mathsf e_i$ and $W^\top W\mathsf e_i=(K_\lambda-\lambda I_n)\mathsf e_i=\frac\lambda3\mathsf e_i$ (D1).
\item The $j$-th entry of $W^\top W\mathsf e_i$ is $\langle W_j,W_i\rangle$. So Step~3 says $\langle W_j,W_i\rangle=0$ for $j\ne i$ and $\|W_i\|^2=\lambda/3$, and conversely these conditions give $W^\top W\mathsf e_i=(\lambda/3)\mathsf e_i$.
\item \emph{Converse.} If $W^\top W\mathsf e_i=(\lambda/3)\mathsf e_i$, then $K_\lambda\mathsf e_i=\frac{4\lambda}3\mathsf e_i$, so $K_\lambda^{-1}\mathsf e_i=\frac3{4\lambda}\mathsf e_i$ and $\Theta\mathsf e_i=\frac34\mathsf e_i$. Hence $\Gamma_1=3/4$, $\Gamma_2=\|\Theta\mathsf e_i\|^2=9/16$ and $(\Gamma_1-\Gamma_2)\Gamma_2=\frac3{16}\cdot\frac9{16}=\frac{27}{256}$.
\item \emph{The derivative.} By \eqref{eq:A4gamma} and Steps~8--10 of the proof of Theorem~\ref{thm:C4b},
\[
|\ell_i'(\eta)|\le\frac{2\|E\|_{\op}}{\sqrt\lambda}\sqrt{(\Gamma_1-\Gamma_2)\Gamma_2}\le\frac{2\|E\|_{\op}}{\sqrt\lambda}\sqrt{27/256}.
\]
If $E\ne0$ and the outer terms are equal, then, dividing by $2\|E\|_{\op}/\sqrt\lambda>0$ and squaring (Tool~\ref{tool:mono}), $(\Gamma_1-\Gamma_2)\Gamma_2=27/256$ at $\eta$, and Steps~1--4 apply to $W(\eta)$.
\end{steps}
\end{proof}

\begin{corollary}[C4(a), Lipschitz part]\label{cor:C4a}\MkOursChk
For all $d,n,\lambda,i,W,W'$: $|\ell_i(W;\lambda)-\ell_i(W';\lambda)|\le\|W-W'\|_{\op}/\sqrt\lambda$.
\end{corollary}
\begin{proof}
\begin{steps}
\item $3\sqrt3/8<1$: both sides are positive and $(3\sqrt3)^2=27<64=8^2$ (Tool~\ref{tool:mono}, square root).
\item By Theorem~\ref{thm:C4b} and Step~1, $|\ell_i(W;\lambda)-\ell_i(W';\lambda)|\le\frac{3\sqrt3}{8\sqrt\lambda}\|W-W'\|_{\op}\le\frac{1}{\sqrt\lambda}\|W-W'\|_{\op}$.
\end{steps}
\end{proof}

\begin{remark}[a related bound in the literature]\label{rem:fadnavis}\MkRep
For ridge leverage of the rows of a design matrix, Fadnavis \cite{fadnavis2026} (Preprint (arXiv:2602.12693), Prop.~D.15(ii), p.~33) states a bound of order $\varepsilon/(\sigma_p^2+\lambda)$ on the change of every leverage score under a perturbation of Frobenius norm at most $\varepsilon$, with $\sigma_p$ the smallest singular value there, an unspecified constant and a sketched proof. Theorem~\ref{thm:C4b} and Corollary~\ref{cor:C4a} give explicit, scale-free constants in the operator norm, with no condition on the singular values.
\end{remark}

\subsection{Effective dimension in the nuclear norm}\label{ssec:C5}

The effective dimension is a sum of scalar leverages of the singular values, so the scalar constant transfers once singular values are known to move by at most the nuclear-norm distance (Mirsky's inequality).

\begin{theorem}[C5: effective dimension]\label{thm:C5}\MkOursChk
For all $d,n\ge1$, $\lambda>0$ and $W,W'\in\R^{d\times n}$:
\begin{equation}\label{eq:C5}
|\deff(W;\lambda)-\deff(W';\lambda)|\le\frac{3\sqrt3}{8\sqrt\lambda}\|W-W'\|_*\le\frac{3\sqrt3}{8\sqrt\lambda}\sqrt{\min(d,n)}\,\|W-W'\|_F;
\end{equation}
the constant $3\sqrt3/8$ cannot be lowered in either inequality of \eqref{eq:C5} (for the class of all $d,n\ge1$; the witness has $d=n=1$); and
\begin{equation}\label{eq:rdeff}
0\le r-\deff(W;\lambda)=\sum_{k=1}^r\frac{\lambda}{\sigma_k^2+\lambda}\le\frac{r\lambda}{\sigma_r^2+\lambda}\quad(r\ge1),\qquad \deff(W;\lambda)=0\ \text{if}\ r=0 .
\end{equation}
The proof rests on the tool \eqref{eq:mirsky} (Mirsky, full text not seen).
\end{theorem}
\begin{proof}
Tools used: Tools~\ref{tool:sv}, \ref{tool:cs}, \ref{tool:mirsky}; Lemmas~\ref{lem:P2}, \ref{lem:P3}; Proposition~\ref{prop:scalar}(i).
\begin{steps}
\item By Lemma~\ref{lem:P3}, $\deff(W;\lambda)=\sum_{k=1}^{\min(d,n)}\lsc(\sigma_k(W))$, and the same for $W'$.
\item Triangle inequality: $|\deff(W)-\deff(W')|\le\sum_k|\lsc(\sigma_k(W))-\lsc(\sigma_k(W'))|$.
\item Each term is $\le\Lsc(\lambda)|\sigma_k(W)-\sigma_k(W')|$ by Proposition~\ref{prop:scalar}(i).
\item Summing and applying \eqref{eq:mirsky}: the sum is $\le\Lsc(\lambda)\|W-W'\|_*$. This is the first inequality of \eqref{eq:C5}.
\item For $B=W-W'$: $\|B\|_*=\sum_{k=1}^{\min(d,n)}\sigma_k(B)\le\sqrt{\min(d,n)}\big(\sum_k\sigma_k(B)^2\big)^{1/2}=\sqrt{\min(d,n)}\,\|B\|_F$ (Tool~\ref{tool:cs} in $\R^{\min(d,n)}$ against the all-ones vector; Tool~\ref{tool:sv}(iv)). This is the second inequality.
\item Sharpness, in both inequalities: Proposition~\ref{prop:scalar}, Step~11 (for $d=n=1$, $\sqrt{\min(d,n)}=1$ and $\|\cdot\|_*=\|\cdot\|_F=\|\cdot\|_{\op}$ on $1\times1$ matrices).
\item By Lemma~\ref{lem:P3}, $r-\deff=\sum_{k=1}^r\big(1-\frac{\sigma_k^2}{\sigma_k^2+\lambda}\big)=\sum_{k=1}^r\frac{\lambda}{\sigma_k^2+\lambda}\ge0$.
\item For $k\le r$, $\sigma_k\ge\sigma_r$ (ordering), so $\frac{\lambda}{\sigma_k^2+\lambda}\le\frac{\lambda}{\sigma_r^2+\lambda}$; summing $r$ terms gives the upper bound in \eqref{eq:rdeff}.
\item If $r=0$, the sum in Lemma~\ref{lem:P3} over $k\le r$ is empty, so $\deff=0$.
\end{steps}
\end{proof}

\noindent If the tool \eqref{eq:mirsky} is to be avoided, a proof by differentiating $\lambda\tr(S_\lambda^{-1})$ along the segment would be needed; we do not give it here.

\subsection{The regularised dimensionality and its defect identity}\label{ssec:C6}

$\Dt_i$ is $D_i$ with $\lambda\|W_i\|^2$ added to the denominator. It is continuous everywhere, lies below $D_i$, and its total satisfies an exact defect identity in which a continuous ridge term replaces the rank deficit. The identity (parts (b)--(d) below) is a corollary of Theorem~8 of Ivanov et al.\ \cite{ivanov2026}, applied to the augmented matrix $[W,\sqrt\lambda I_d]$; because a proof of ours may not depend on a preprint, we also prove it directly, and we claim no novelty for (b)--(d).

\begin{proposition}[Ivanov et al., Theorem~8, in our notation]\label{prop:I8}\MkRep
(\cite{ivanov2026}, Preprint (arXiv:2602.02224), App.~D, p.~29.) Let $B\in\R^{d\times p}$ have columns $B_1,\dots,B_p$, $r_B=\rank(BB^\top)$, unregularised leverage $\ell^0_j(B)=B_j^\top(BB^\top)^+B_j$, and $D_j(B)=\|B_j\|^4/(B_j^\top BB^\top B_j)$ for $B_j\ne0$. Then: (1) $\sum_j\ell^0_j(B)=r_B$; (2) $D_j(B)\le\ell^0_j(B)$ for all $j$, with equality if and only if $B_j$ is an eigenvector of $BB^\top$; (3) $r_B-\sum_jD_j(B)=\sum_j(\ell^0_j(B)-D_j(B))$.
\end{proposition}

\begin{theorem}[C6(a)]\label{thm:C6a}\MkOursChk
For all $d,n\ge1$, $\lambda>0$, $i\in[n]$ and $W\in\R^{d\times n}$: $0\le\Dt_i(W;\lambda)\le D_i(W)\le\Dt_i(W;\lambda)+\lambda/(\|W_i\|^2+\lambda)$; $\lambda\mapsto\Dt_i(W;\lambda)$ is non-increasing with $\lim_{\lambda\downarrow0}\Dt_i(W;\lambda)=D_i(W)$; $\Dt_i(\cdot;\lambda)$ is continuous on $\R^{d\times n}$; and $\Dt_i(QW;\lambda)=\Dt_i(W;\lambda)$ for $Q\in O(d)$.
\end{theorem}
\begin{proof}
Here $\mathsf q_i=W_i^\top SW_i$ and $\mathsf a=\|W_i\|$ (D1). Tools used: Tools~\ref{tool:calc}, \ref{tool:cont}; Lemmas~\ref{lem:P1}, \ref{lem:P2}, \ref{lem:P4}.
\begin{steps}
\item If $W_i=0$: $\Dt_i=D_i=0$ (D2, D5), and $\lambda/(0+\lambda)=1\ge0$, so the chain holds; monotonicity and the limit are trivial (constant $0$). Assume $W_i\ne0$ in Steps~2--5.
\item By Lemma~\ref{lem:P4}, $\mathsf q_i\ge\mathsf a^4>0$. Then $\Dt_i=\mathsf a^4/(\mathsf q_i+\lambda\mathsf a^2)>0$ and, since $\mathsf q_i+\lambda\mathsf a^2>\mathsf q_i$, $\Dt_i<\mathsf a^4/\mathsf q_i=D_i$ (D2, D5).
\item $D_i-\Dt_i=\mathsf a^4\Big(\frac1{\mathsf q_i}-\frac1{\mathsf q_i+\lambda\mathsf a^2}\Big)=\frac{\lambda\mathsf a^6}{\mathsf q_i(\mathsf q_i+\lambda\mathsf a^2)}$. This is non-increasing in $\mathsf q_i$, so with $\mathsf q_i\ge\mathsf a^4$ (Step~2) it is $\le\frac{\lambda\mathsf a^6}{\mathsf a^4(\mathsf a^4+\lambda\mathsf a^2)}=\frac{\lambda}{\mathsf a^2+\lambda}$.
\item The denominator $\mathsf q_i+\lambda\mathsf a^2$ is increasing in $\lambda$, so $\Dt_i$ is (strictly) decreasing in $\lambda$; as a rational function of $\lambda$ with denominator $\mathsf q_i>0$ at $\lambda=0$, it tends to $\mathsf a^4/\mathsf q_i=D_i$ as $\lambda\downarrow0$ (Tool~\ref{tool:calc}(e)).
\item \emph{Continuity at $W$ with $W_i\ne0$.} On the open set $\{W':W'_i\ne0\}$ (open by P1.1) $\Dt_i$ is a quotient of polynomials in the entries with denominator $\ge\lambda\|W'_i\|^2>0$, hence continuous (P1.2).
\item \emph{Continuity at $W$ with $W_i=0$.} Here $\Dt_i(W)=0$. For any $W'$ with $W'_i\ne0$, $\Dt_i(W')\le\|W'_i\|^4/(\lambda\|W'_i\|^2)=\|W'_i\|^2/\lambda$ (drop $\mathsf q_i'\ge0$), and for $W'_i=0$ it is $0$. By P1.1, $\|W'_i\|\le\|W'-W\|_{\op}$, so $|\Dt_i(W')-\Dt_i(W)|\le\|W'-W\|_{\op}^2/\lambda\to0$.
\item \emph{Invariance.} For $Q\in O(d)$: $(QW)_i=QW_i$, $\|QW_i\|=\|W_i\|$, $S(QW)=QSQ^\top$, so $(QW_i)^\top QSQ^\top(QW_i)=W_i^\top SW_i$ ($Q^\top Q=I_d$). Both numerator and denominator of D5 are unchanged, and $QW_i=0$ iff $W_i=0$.
\end{steps}
\end{proof}

\begin{theorem}[C6(b)--(d): regularised defect identity]\label{thm:C6bd}\MkRepOursChk
For all $d,n\ge1$, $\lambda>0$, $i\in[n]$ and $W\in\R^{d\times n}$:
\begin{enumerate}[label=(\alph*),start=2]
\item $\Dt_i(W;\lambda)\le\ell_i(W;\lambda)$, with equality iff $W_i=0$ or $SW_i=\nu W_i$ for some $\nu\in\R$.
\item \begin{equation}\label{eq:C6c}
d-\Dttot(W;\lambda)=\underbrace{(d-r)}_{\text{rank term}}+\underbrace{\sum_{k=1}^r\frac{\lambda}{\sigma_k^2+\lambda}}_{\text{ridge term}}+\underbrace{\sum_{i=1}^n\big(\ell_i(W;\lambda)-\Dt_i(W;\lambda)\big)}_{\text{eigenvector term}},
\end{equation}
each term is $\ge0$, and $\Dttot\le\deff\le r\le\min(d,n)$.
\item $\Dttot(W;\lambda)=\deff(W;\lambda)$ iff every nonzero column of $W$ is an eigenvector of $S$.
\end{enumerate}
The statement is a corollary of Proposition~\ref{prop:I8} (reduction below); the direct proof is ours.
\end{theorem}

\begin{proof}[Direct proof of (b)]
\begin{steps}
\item If $W_i=0$, both sides are $0$ (D4, D5), and the equality condition holds. Assume $W_i\ne0$.
\item $\langle z,z'\rangle_{S_\lambda}:=z^\top S_\lambda z'$ is an inner product on $\R^d$ (bilinear, symmetric, and positive definite by Lemma~\ref{lem:P2}).
\item Let $z:=S_\lambda^{-1}W_i\ne0$. Then $\langle W_i,z\rangle_{S_\lambda}=W_i^\top W_i=\|W_i\|^2$, $\langle W_i,W_i\rangle_{S_\lambda}=\mathsf q_i+\lambda\|W_i\|^2$ and $\langle z,z\rangle_{S_\lambda}=W_i^\top S_\lambda^{-1}W_i=\ell_i$ (symmetry of $S_\lambda^{-1}$).
\item By Tool~\ref{tool:cs}: $\|W_i\|^4\le(\mathsf q_i+\lambda\|W_i\|^2)\,\ell_i$, i.e.\ $\Dt_i\le\ell_i$ (D5).
\item Equality holds iff $z=\mathsf c'W_i$ for some $\mathsf c'$ (Tool~\ref{tool:cs}; both are nonzero), and then $\mathsf c'\ne0$. This is $W_i=\mathsf c'S_\lambda W_i$, i.e.\ $SW_i=(1/\mathsf c'-\lambda)W_i$.
\item Conversely, if $SW_i=\nu W_i$, then $S_\lambda W_i=(\nu+\lambda)W_i$ and $\nu+\lambda\ne0$ ($S_\lambda$ is injective and $W_i\ne0$), so $z=W_i/(\nu+\lambda)$ is a multiple of $W_i$ and equality holds by Step~5.
\end{steps}
\end{proof}

\begin{proof}[Direct proof of (c)]
\begin{steps}
\item $d-\Dttot=(d-\deff)+\sum_i(\ell_i-\Dt_i)$, since $\deff=\sum_i\ell_i$ (D4, D5).
\item By Lemmas~\ref{lem:P3} and \ref{lem:P2}, $d-\deff=\sum_{k=1}^d\frac{\lambda}{\nu_k(S)+\lambda}$; the $d-r$ indices with $\nu_k(S)=0$ contribute $1$ each, the others give $\sum_{k=1}^r\frac{\lambda}{\sigma_k^2+\lambda}$.
\item Nonnegativity: $d-r\ge0$ since $r\le\min(d,n)$ (Tool~\ref{tool:sv}(iii)); the ridge term is a sum of positive numbers; the eigenvector term is $\ge0$ by (b).
\item By Steps~1--2 and \eqref{eq:rdeff}: $\Dttot=\deff-\sum_i(\ell_i-\Dt_i)\le\deff=r-\sum_{k\le r}\frac{\lambda}{\sigma_k^2+\lambda}\le r\le\min(d,n)$. (If $r=0$, then $\deff=0=r$ by \eqref{eq:rdeff} and the ridge sum is empty.)
\end{steps}
\end{proof}

\begin{proof}[Direct proof of (d)]
\begin{steps}
\item By D4 and D5, $\deff-\Dttot=\sum_i(\ell_i-\Dt_i)$, so $\Dttot=\deff$ iff $\sum_i(\ell_i-\Dt_i)=0$.
\item Each term $\ell_i-\Dt_i$ is $\ge0$ by (b), so the sum is $0$ iff every term is $0$.
\item By the equality case of (b), $\ell_i=\Dt_i$ iff $W_i=0$ or $SW_i=\nu W_i$ for some $\nu$. So $\Dttot=\deff$ iff every nonzero column of $W$ is an eigenvector of $S$.
\end{steps}
\end{proof}

\begin{proof}[Proof by reduction to Proposition~\ref{prop:I8}]
This proof covers (b) and the identity in (c) by reduction to Proposition~\ref{prop:I8}; the inequalities in (c), and (d), follow from Lemmas~\ref{lem:P2}, \ref{lem:P3}, \eqref{eq:rdeff} and (b).
\begin{steps}
\item Let $\Waug:=[W,\ \sqrt\lambda I_d]\in\R^{d\times(n+d)}$, with columns $W_1,\dots,W_n,\sqrt\lambda\mathsf e_1,\dots,\sqrt\lambda\mathsf e_d$.
\item $S(\Waug)=\Waug\Waug^\top=\sum_jW_jW_j^\top+\lambda\sum_k\mathsf e_k\mathsf e_k^\top=S+\lambda I_d=S_\lambda$. It is invertible (Lemma~\ref{lem:P2}), so $\rank S(\Waug)=d$ and $S(\Waug)^+=S_\lambda^{-1}$ (Tool~\ref{tool:pinv}).
\item For $i\le n$, the leverage of column $i$ of $\Waug$ in Proposition~\ref{prop:I8} is $\ell^0_i(\Waug)=W_i^\top S_\lambda^{-1}W_i=\ell_i(W;\lambda)$, and for $W_i\ne0$, $D_i(\Waug)=\|W_i\|^4/(W_i^\top SW_i+\lambda\|W_i\|^2)=\Dt_i(W;\lambda)$ (D4, D5).
\item For $W_i\ne0$, Proposition~\ref{prop:I8}(2) for column $i$ of $\Waug$ gives $\Dt_i\le\ell_i$, with equality iff $W_i$ is an eigenvector of $S(\Waug)=S_\lambda$, i.e.\ of $S$ (as $S_\lambda W_i=(\nu+\lambda)W_i$ iff $SW_i=\nu W_i$). With the trivial case $W_i=0$ (both sides $0$ by D4, D5) this is (b).
\item Proposition~\ref{prop:I8}(1) applied to $\Waug$ gives $\sum_{i\le n}\ell_i+\sum_{k=1}^d\lambda\mathsf e_k^\top S_\lambda^{-1}\mathsf e_k=\rank S(\Waug)=d$ (Steps~2--3), i.e.\ $d-\deff=\lambda\tr(S_\lambda^{-1})$ (D4). This re-derives the second identity of \eqref{eq:deff}.
\item \emph{Split of $d-\deff$} (not from Proposition~\ref{prop:I8}). By \eqref{eq:deff} and Lemma~\ref{lem:P2}, $\lambda\tr(S_\lambda^{-1})=\sum_{k=1}^d\frac{\lambda}{\nu_k(S)+\lambda}$; the $d-r$ indices with $\nu_k(S)=0$ contribute $1$ each, and the others contribute $\sum_{k=1}^r\frac{\lambda}{\sigma_k^2+\lambda}$. So $d-\deff=(d-r)+\sum_{k=1}^r\frac{\lambda}{\sigma_k^2+\lambda}$.
\item \emph{Identity in (c).} $d-\Dttot=(d-\deff)+\sum_{i\le n}(\ell_i-\Dt_i)$ (D4, D5). With Step~6 this is \eqref{eq:C6c}.
\item \emph{Each term is $\ge0$.} By Lemma~\ref{lem:P2}, exactly $r$ of the $d$ eigenvalues of $S$ are nonzero, so $r\le d$; and $r\le\min(d,n)$ by Tool~\ref{tool:sv}(iii). So the rank term $d-r$ is $\ge0$. The ridge term is a sum of the positive numbers $\lambda/(\sigma_k^2+\lambda)$ ($\lambda>0$), hence $\ge0$ (empty if $r=0$). The eigenvector term is $\ge0$ by Step~4.
\item \emph{The chain.} By Step~7, $\Dttot=\deff-\sum_i(\ell_i-\Dt_i)\le\deff$. By \eqref{eq:rdeff} (Theorem~\ref{thm:C5}), $\deff=r-\sum_{k\le r}\frac{\lambda}{\sigma_k^2+\lambda}\le r$ (and $\deff=0=r$ if $r=0$). With Step~8, $r\le\min(d,n)$.
\item \emph{(d)} follows from Step~4 and Step~7 exactly as in the direct proof of (d) (Steps~1--3 there use only (b) and D4, D5).
\item Proposition~\ref{prop:I8}(3) applied to $\Waug$ is a different identity from (c): it also contains the defects $\ell^0-D$ of the columns $\sqrt\lambda\mathsf e_k$. The reduction uses Proposition~\ref{prop:I8}(1)--(2) only.
\end{steps}
\end{proof}

\noindent The case $\lambda=0$ of this structure (with $\sum D_i\le\rank$) is also in Scherlis et al.\ \cite{scherlis2022}.

\subsection{The Lipschitz profile of $\Dt_i$: definition and lower bounds}\label{ssec:profile}

Unlike $\ell_i$, the regularised dimensionality has no Lipschitz constant that depends on $\lambda$ alone. To measure how its constant grows with the size of $W$, we use the best constant among all pairs of a given scale-free size $\rho=R/\sqrt\lambda$.

\begin{definition}[Lipschitz profile]\label{def:profile}\MkDef
For $\rho>0$,
\begin{equation}\label{eq:Phi}
\begin{aligned}
\Phi(\rho):=\sup\Big\{&\frac{\sqrt\lambda\,|\Dt_i(W;\lambda)-\Dt_i(W';\lambda)|}{\|W-W'\|_{\op}}:\\
&d,n\ge1,\ i\in[n],\ \lambda>0,\ W\ne W',\ \max(\|W\|_{\op},\|W'\|_{\op})\le\rho\sqrt\lambda\Big\}\in[0,\infty],
\end{aligned}
\end{equation}
$\chi(\rho):=\Phi(\rho)-\rho/4$ and $\chi^\star:=\sup_{\rho>0}\chi(\rho)$. A tuple $(d,n,i,\lambda,W,W')$ satisfying the conditions in the braces is called \emph{admissible at $\rho$}. Let $\Phi_{\lambda=1}(\rho)$ be the same supremum with $\lambda=1$ fixed.
\end{definition}

\begin{proposition}[profile: scale invariance and lower bounds]\label{prop:profile}\MkOursChk
\begin{enumerate}[label=(\roman*)]
\item (Scale invariance) $\Dt_i(cW;c^2\lambda)=\Dt_i(W;\lambda)$ for $c>0$; hence $\Phi=\Phi_{\lambda=1}$. $\Phi$ is non-decreasing.
\item (\textsf{F1}) $\Phi(\rho)\ge\frac14\sqrt{\rho^2-1}$ for $\rho>1$, and $\liminf_{\rho\to\infty}\Phi(\rho)/\rho\ge1/4$. Moreover, let $\lambda>0$ and $L_a,L_c\in\R$ (which may depend on $\lambda$, but not on $W,W'$). If $|\Dt_1(W;\lambda)-\Dt_1(W';\lambda)|\le(L_a/\sqrt\lambda+L_cR/\lambda)\|W-W'\|_{\op}$ for all $W,W'\in\R^{2\times2}$, then $L_c\ge1/4$. In particular, no such bound required for all $d,n,i,\lambda,W,W'$ has $L_c<1/4$, whatever $L_a$.
\item (\textsf{F2}) $\Phi(\rho)\ge\frac14\sqrt{\rho^2-1}+\frac1{4\sqrt2}$ for $\rho>1$; so $\chi(\rho)\ge\frac1{4\sqrt2}-\frac1{4\rho}$ and $\liminf_{\rho\to\infty}\chi(\rho)\ge1/(4\sqrt2)$.
\item (Scalar case) $\Phi(\rho)\ge\Fsc(\rho)=2\rho/(1+\rho^2)^2$ for every $\rho>0$; in particular $\chi^\star\ge\chi(1/2)\ge103/200$.
\end{enumerate}
\end{proposition}
\begin{proof}[Proof of (i)]
\begin{steps}
\item If $W_i=0$, both sides are $0$. Otherwise, with $S(cW)=c^2S$: $\Dt_i(cW;c^2\lambda)=\frac{c^4\|W_i\|^4}{c^4W_i^\top SW_i+c^2\lambda\cdot c^2\|W_i\|^2}=\Dt_i(W;\lambda)$ (D5).
\item $\Phi_{\lambda=1}(\rho)\le\Phi(\rho)$, since the set for $\Phi_{\lambda=1}$ is a subset.
\item Given an admissible $(d,n,i,\lambda,W,W')$, put $c:=\lambda^{-1/2}$. Then $(cW,cW')$ with ridge parameter $c^2\lambda=1$ is admissible for $\Phi_{\lambda=1}(\rho)$: $\max(\|cW\|_{\op},\|cW'\|_{\op})\le c\rho\sqrt\lambda=\rho$; and its quotient is $\frac{1\cdot|\Dt_i(cW;1)-\Dt_i(cW';1)|}{c\|W-W'\|_{\op}}=\frac{\sqrt\lambda|\Dt_i(W;\lambda)-\Dt_i(W';\lambda)|}{\|W-W'\|_{\op}}$ (Step~1). So $\Phi(\rho)\le\Phi_{\lambda=1}(\rho)$.
\item For $\rho<\rho'$ the admissible set at $\rho$ is contained in that at $\rho'$, so $\Phi(\rho)\le\Phi(\rho')$.
\end{steps}
\end{proof}

\begin{proof}[Proof of (ii) and (iii): from a slope to a lower bound on $\Phi$]
Take $\lambda=1$ and the family of Definition~\ref{def:family} with $\alpha_0=1/\sqrt2$ and $\iota\in\{0,1\}$.
\begin{steps}
\item Let $\rho>1$ and $R_0\in\big(1/\sqrt2,\sqrt{\rho^2-1/2}\big)$ (a nonempty interval since $\rho^2-1/2>1/2$). Then $\|W^\star\|_{\op}\le(R_0^2+1/2)^{1/2}<\rho$ (Lemma~\ref{lem:family}(ii)).
\item For $0<|\eta|<\rho-(R_0^2+1/2)^{1/2}$: $\|W^\star+\eta E^{(\iota)}\|_{\op}\le\|W^\star\|_{\op}+|\eta|<\rho$ and $\|\eta E^{(\iota)}\|_{\op}=|\eta|$ (Lemma~\ref{lem:family}(ii)). So the pair $(W^\star,W^\star+\eta E^{(\iota)})$, $i=1$, $\lambda=1$, is admissible, and by Lemma~\ref{lem:family}(i), $\Phi(\rho)\ge|\mathcal C_1(\eta)-\mathcal C_1(0)|/|\eta|$.
\item Letting $\eta\to0$ (Lemma~\ref{lem:family}(iii)): $\Phi(\rho)\ge|\mathcal C_1'(0)|$. By Corollary~\ref{cor:special}(3)--(4) with $\lambda=1$ and $R_0\sqrt{1-1/(2R_0^2)}=\sqrt{R_0^2-1/2}$: $\Phi(\rho)\ge\frac14\sqrt{R_0^2-1/2}+\frac{\iota}{4\sqrt2}$.
\item Let $R_0\uparrow\sqrt{\rho^2-1/2}$; then $R_0^2-1/2\uparrow\rho^2-1$, so $\Phi(\rho)\ge\frac14\sqrt{\rho^2-1}+\frac{\iota}{4\sqrt2}$. With $\iota=0$ this is the bound of (ii) and with $\iota=1$ it is that of (iii).
\item $\Phi(\rho)/\rho\ge\frac14\sqrt{1-1/\rho^2}\to\frac14$, so $\liminf\Phi(\rho)/\rho\ge1/4$.
\item \emph{The bound with $L_a,L_c$} (any fixed $\lambda>0$; this step does not use Steps~1--5). Suppose the bound holds for all $W,W'\in\R^{2\times2}$, and put $L_c^+:=\max(L_c,0)$; since $R\ge0$, the bound also holds with $L_c^+$. Take the family of Definition~\ref{def:family} with this $\lambda$, $\alpha_0=\sqrt{\lambda/2}$, $\iota=0$ (\textsf{F1}) and $R_0>\sqrt{\lambda/2}$. For $\eta\ne0$, the pair $W=W^\star$, $W'=W^\star(\eta)$ has $\|W'-W\|_{\op}=|\eta|$ and $R\le(R_0^2+\lambda/2)^{1/2}+|\eta|$ (Lemma~\ref{lem:family}(ii), Tool~\ref{tool:op}). By Lemma~\ref{lem:family}(i), dividing by $|\eta|$, letting $\eta\to0$ (Lemma~\ref{lem:family}(iii)) and using Corollary~\ref{cor:special}(3),
\[
\frac{R_0}{4\lambda}\sqrt{1-\frac{\lambda}{2R_0^2}}\le\frac{L_a}{\sqrt\lambda}+\frac{L_c^+(R_0^2+\lambda/2)^{1/2}}{\lambda}.
\]
Multiply by $\lambda/R_0>0$: $\frac14\sqrt{1-\lambda/(2R_0^2)}\le L_a\sqrt\lambda/R_0+L_c^+\sqrt{1+\lambda/(2R_0^2)}$. Letting $R_0\to\infty$ ($L_a,L_c^+$ do not depend on $R_0$) gives $L_c^+\ge1/4>0$, so $L_c=L_c^+\ge1/4$. A bound required for all $d,n,i,\lambda,W,W'$ holds in particular for $d=n=2$, $i=1$ and each fixed $\lambda$.
\item $\chi(\rho)\ge\frac1{4\sqrt2}-\frac14\big(\rho-\sqrt{\rho^2-1}\big)$ and $\rho-\sqrt{\rho^2-1}=\frac{1}{\rho+\sqrt{\rho^2-1}}\le\frac1\rho$; so $\chi(\rho)\ge\frac1{4\sqrt2}-\frac1{4\rho}$ and $\liminf\chi(\rho)\ge\frac1{4\sqrt2}$.
\end{steps}
\end{proof}

\begin{proof}[Proof of (iv)]
\begin{steps}
\item Let $d=n=1$, $\lambda=1$. For $w\ne0$, $\Dt_1((w);1)=w^4/(w^4+w^2)=w^2/(w^2+1)=\ell^{\mathrm{sc}}_1(w)$ (D5, $S=w^2$), and for $w=0$ both are $0$.
\item For $\rho>0$ and $0<\eta<\rho$, the pair $w=\rho$, $w'=\rho-\eta$ is admissible ($\max(|w|,|w'|)=\rho$) with $\|W-W'\|_{\op}=\eta$, so $\Phi(\rho)\ge|\ell^{\mathrm{sc}}_1(\rho)-\ell^{\mathrm{sc}}_1(\rho-\eta)|/\eta\to(\ell^{\mathrm{sc}}_1)'(\rho)=2\rho/(1+\rho^2)^2$ (Proposition~\ref{prop:scalar}, Step~1, with $\lambda=1$).
\item At $\rho=1/2$: $2\cdot\frac12/(\frac54)^2=\frac{16}{25}$, so $\chi(\frac12)\ge\frac{16}{25}-\frac18=\frac{128-25}{200}=\frac{103}{200}$, and $\chi^\star\ge\chi(\frac12)$.
\end{steps}
\end{proof}

\begin{proposition}[C7(b): the dependence on $R$ cannot be removed]\label{prop:C7b}\MkOurs
Let $d=n=2$, $\lambda>0$, $R_0>\sqrt\lambda$, and $\phi\in(0,\pi/2)$ with $\cos\phi=\sqrt\lambda/R_0$. Let $W^\star$ have columns $W^\star_1=\sqrt\lambda(\cos\phi,\sin\phi)^\top$ and $W^\star_2=(R_0,0)^\top$, and let $E^{(0)}$ have columns $E^{(0)}_1=(-\sin\phi,\cos\phi)^\top$ and $E^{(0)}_2=0$, so $\|E^{(0)}\|_{\op}=1$. Then $\|W^\star\|_{\op}\le(R_0^2+\lambda)^{1/2}$ and
\[
\Big|\tfrac{\mathrm d}{\mathrm d\eta}\Dt_1(W^\star+\eta E^{(0)};\lambda)\big|_{\eta=0}\Big|=\frac{2R_0}{9\lambda}\sqrt{1-\lambda/R_0^2}.
\]
So no inequality $|\Dt_i(W;\lambda)-\Dt_i(W';\lambda)|\le L(\lambda)\|W-W'\|_{\op}$ with $L(\lambda)$ depending only on $\lambda$ holds for all $W,W'$.
\end{proposition}
\begin{proof}
\begin{steps}
\item The matrices are the family of Definition~\ref{def:family} with $\alpha_0=\sqrt\lambda$, $\iota=0$; the norm bound and the slope are Corollary~\ref{cor:special}(2) and Lemma~\ref{lem:family}(i).
\item Suppose $|\Dt_1(W;\lambda)-\Dt_1(W';\lambda)|\le L(\lambda)\|W-W'\|_{\op}$ for all $W,W'\in\R^{2\times2}$. Take $W=W^\star$, $W'=W^\star(\eta)$, $\eta\ne0$: by Lemma~\ref{lem:family}(i)--(ii), $|\mathcal C_\lambda(\eta)-\mathcal C_\lambda(0)|\le L(\lambda)|\eta|$.
\item Divide by $|\eta|$ and let $\eta\to0$ (Lemma~\ref{lem:family}(iii)): $|\mathcal C_\lambda'(0)|\le L(\lambda)$, for every $R_0>\sqrt\lambda$.
\item By Step~1, $\mathcal C_\lambda'(0)=2\sqrt{R_0^2-\lambda}/(9\lambda)$, which exceeds $L(\lambda)$ once $R_0^2>\lambda+(9\lambda L(\lambda)/2)^2$. This contradicts Step~3.
\end{steps}
\end{proof}

\subsection{Upper bounds: the coefficient $1/4$ and the asymptotic constant}\label{ssec:C7p}

The main theorem of this section is that $\Dt_i$ is Lipschitz with constant $\frac{3\sqrt3}{8\sqrt\lambda}+\frac{R}{4\lambda}$, so that, by Proposition~\ref{prop:profile}, the growth rate $1/4$ is exact. The proof differentiates $\Dt_i$ along the segment from $W$ to $W'$, splits the perturbation of column $i$ into a radial part, a tangential part, and a part that moves the other columns along $W_i$ (the coupling part), and maximises each coefficient over the two numbers $\xi=\|W_i\|$ and $\zeta$, the size of the other columns seen from $W_i$. A sharper treatment for large $R$ gives the asymptotic additive constant $1/(4\sqrt2)$, which matches \textsf{F2}. Appendix~\ref{app:B7} proves the coefficient $1/4$ by an independent reduction to a function of two variables.

\begin{definition}[notation for the segment]\label{def:C7pnot}\MkDef
Fix $W,W'$ and use $E$, $W(\eta)$ (defined for $\eta\in\R$ by D1; the segment from $W$ to $W'$ is $\eta\in[0,1]$, and derivatives at $\eta\in\{0,1\}$ are two-sided), $S(\eta)$, $S_{-i}(\eta)$, $R$ and $\Rb$ of D1, and $W_{-i}$; $E_{-i}$ is the matrix of the columns $E_j$, $j\ne i$. With $\lambda=1$ we write $\Dt_i(\eta):=\Dt_i(W(\eta);1)$. At a point with $W_i(\eta)\ne0$ (all quantities at $\eta$):
\begin{itemize}[itemsep=1pt]
\item $\xi=\|W_i\|$ and $\zeta=(\hat W_i^\top S_{-i}\hat W_i)^{1/2}\ge0$;
\item the \emph{radial part} $\qrad:=\langle E_i,\hat W_i\rangle$; the \emph{tangential part} $E_i^\perp:=(I_d-\hat W_i\hat W_i^\top)E_i$ and $\qtan:=\|E_i^\perp\|$; the \emph{coupling part} $\qcpl:=\|E_{-i}^\top\hat W_i\|$; and the coupling product $\Pcpl:=\langle E_{-i}^\top\hat W_i,\,W_{-i}^\top\hat W_i\rangle$.
\end{itemize}
For $\xi>0$, $\zeta\ge0$, the coefficients
\[
\arad:=\frac{2\xi(1+\zeta^2)}{(\xi^2+1+\zeta^2)^2},\qquad
\atan:=\frac{2\xi\zeta}{(\xi^2+1+\zeta^2)^2},\qquad
\acpl:=\frac{2\xi^2\zeta}{(\xi^2+1+\zeta^2)^2}=\xi\,\atan,
\]
and, for $0\le\zeta\le\Rb$, the pointwise bound $\Psi_{\Rb}(\xi,\zeta):=\big(\arad^2+(\acpl+\atan\sqrt{\Rb^2-\zeta^2})^2\big)^{1/2}$. Finally $\alpha_\star(\xi):=\frac{3\sqrt3}{8}\,\xi(1+\xi^2)^{-3/2}$, $\upsilon:=\xi(1+\xi^2)^{-1/2}\in(0,1)$, $\upsilon_\star:=1/\sqrt3$, $\Omega(\upsilon):=\upsilon^2\sqrt{1-\upsilon^2}$, and $\varpi:=\xi^2+\zeta^2$.
\end{definition}

\begin{lemma}[reduction to $\lambda=1$]\label{lem:A71}\MkOursChk
Let $\mathfrak a,\mathfrak b\ge0$ and $\mathfrak h:(0,\infty)\to[0,\infty)$, and fix $d,n\ge1$. Suppose that for all $i\in[n]$, all $W,W'\in\R^{d\times n}$ and every $\Rb>0$ with $\Rb\ge R$, $|\Dt_i(W;1)-\Dt_i(W';1)|\le\big(\mathfrak a+\mathfrak b\Rb+\mathfrak h(\Rb)\big)\|W-W'\|_{\op}$. Then for every $\lambda>0$, all $i$, $W,W'\in\R^{d\times n}$ and every $\Rb>0$ with $\Rb\ge R$,
\[
|\Dt_i(W;\lambda)-\Dt_i(W';\lambda)|\le\Big(\frac{\mathfrak a}{\sqrt\lambda}+\frac{\mathfrak b\Rb}{\lambda}+\frac{\mathfrak h(\Rb/\sqrt\lambda)}{\sqrt\lambda}\Big)\|W-W'\|_{\op} .
\]
\end{lemma}
\begin{proof}
\begin{steps}
\item For $c>0$ and $W_i\ne0$, D5 gives $\Dt_i(cW;c^2\lambda)=c^4\|W_i\|^4/(c^4W_i^\top SW_i+c^4\lambda\|W_i\|^2)=\Dt_i(W;\lambda)$; for $W_i=0$ both sides are $0$.
\item Put $c=\lambda^{-1/2}$, $W^{\mathrm s}=cW$, $W'^{\mathrm s}=cW'$. By Tool~\ref{tool:op}, $\max(\|W^{\mathrm s}\|_{\op},\|W'^{\mathrm s}\|_{\op})=cR\le c\Rb$, and $\|W^{\mathrm s}-W'^{\mathrm s}\|_{\op}=c\|W-W'\|_{\op}$.
\item $W^{\mathrm s},W'^{\mathrm s}\in\R^{d\times n}$ (same $d,n$) and $c\Rb>0$ is $\ge$ their $R$ (Step~2). By Step~1 and the hypothesis applied to $(W^{\mathrm s},W'^{\mathrm s},c\Rb)$: $|\Dt_i(W;\lambda)-\Dt_i(W';\lambda)|=|\Dt_i(W^{\mathrm s};1)-\Dt_i(W'^{\mathrm s};1)|\le(\mathfrak a+\mathfrak bc\Rb+\mathfrak h(c\Rb))\,c\,\|W-W'\|_{\op}$, which is the claim.
\end{steps}
\end{proof}

\begin{lemma}[first-order bound at a point]\label{lem:A72}\MkOursChk
Let $\lambda=1$, $\Rb\ge R$, $\Rb>0$, and let $\eta\in[0,1]$ with $W_i(\eta)\ne0$. Then $\Dt_i$ is differentiable at $\eta$,
\begin{equation}\label{eq:A7deriv}
\Dt_i'(\eta)=\frac{2\xi\big[\qrad(1+\zeta^2)-\langle E_i^\perp,S_{-i}\hat W_i\rangle-\xi\,\Pcpl\big]}{(\xi^2+1+\zeta^2)^2},
\end{equation}
(all quantities at $\eta$), and
\begin{equation}\label{eq:A7pointwise}
|\Dt_i'(\eta)|\le\Psi_{\Rb}(\xi,\zeta)\,\|E\|_{\op},\qquad\text{with }\ \xi^2+\zeta^2\le\Rb^2 .
\end{equation}
If $d=1$ or $n=1$, then in fact $|\Dt_i'(\eta)|\le\big(\arad^2+\acpl^2\big)^{1/2}\|E\|_{\op}$.
\end{lemma}
\begin{proof}
Tools used: Tools~\ref{tool:spec}, \ref{tool:op}, \ref{tool:cs}, \ref{tool:calc}.
\begin{steps}
\item \emph{Size along the segment.} By Tool~\ref{tool:op}, $\|W(\eta)\|_{\op}\le(1-\eta)\|W\|_{\op}+\eta\|W'\|_{\op}\le R\le\Rb$, and $\nu_{\max}(S(\eta))=\|W(\eta)\|_{\op}^2\le\Rb^2$.
\item \emph{Spectral bounds.} For $x\in\R^d$, $x^\top S_{-i}x=x^\top Sx-\langle W_i,x\rangle^2\le x^\top Sx\le\Rb^2\|x\|^2$ (Tool~\ref{tool:spec}, Step~1), and $x^\top S_{-i}x=\sum_{j\ne i}\langle W_j,x\rangle^2\ge0$. So $S_{-i}$ is positive semidefinite with $\nu_{\max}(S_{-i})\le\Rb^2$ (take $x$ a unit eigenvector). Moreover $\xi^2+\zeta^2=\|W_i\|^2+\hat W_i^\top S_{-i}\hat W_i=\hat W_i^\top S\hat W_i\le\Rb^2$ (Tool~\ref{tool:spec}), since $\hat W_i^\top W_iW_i^\top\hat W_i=\|W_i\|^2$.
\item \emph{Polynomial form.} With $\lambda=1$ and $S=W_iW_i^\top+S_{-i}$, at every $\eta$ with $W_i(\eta)\ne0$ we have $\Dt_i(\eta)=\Numseg(\eta)/\Denseg(\eta)$, where $\Numseg:=\|W_i\|^4$ and $\Denseg:=\|W_i\|^4+\|W_i\|^2+W_i^\top S_{-i}W_i$ are polynomials in $\eta$, and $\Denseg\ge\|W_i\|^2>0$. Since $W_i(\cdot)$ is continuous, $W_i\ne0$ on a neighbourhood of $\eta$ in $\R$, so $\Dt_i=\Numseg/\Denseg$ there and $\Dt_i$ is differentiable at $\eta$ (Tool~\ref{tool:calc}(e), a rational function with nonvanishing denominator; the derivative by the quotient rule, Tool~\ref{tool:calc}(a)).
\item \emph{Derivatives of the pieces.} Using $W_j'=E_j$: $(\|W_i\|^2)'=2\langle W_i,E_i\rangle$, and
\begin{align*}
(W_i^\top S_{-i}W_i)'&=\Big(\sum_{j\ne i}\langle W_j,W_i\rangle^2\Big)'=\sum_{j\neq i}2\langle W_j,W_i\rangle\big(\langle E_j,W_i\rangle+\langle W_j,E_i\rangle\big)\\
&=2\langle E_i,S_{-i}W_i\rangle+2\sum_{j\ne i}\langle W_i,E_j\rangle\langle W_i,W_j\rangle .
\end{align*}
\item \emph{In terms of $\xi,\zeta$.} Write $W_i=\xi\hat W_i$ and $E_i=\qrad\hat W_i+E_i^\perp$ (orthogonal decomposition). Then $\langle W_i,E_i\rangle=\xi\qrad$; $W_i^\top S_{-i}W_i=\xi^2\zeta^2$; $\langle E_i,S_{-i}W_i\rangle=\xi\big(\qrad\zeta^2+\langle E_i^\perp,S_{-i}\hat W_i\rangle\big)$; and $\sum_{j\neq i}\langle W_i,E_j\rangle\langle W_i,W_j\rangle=\xi^2\Pcpl$. Hence
\begin{align*}
\Numseg&=\xi^4,\\
\Denseg&=\xi^4+\xi^2+\xi^2\zeta^2,\\
\Numseg'&=2\xi^2\cdot2\xi\qrad=4\xi^3\qrad,\\
\Denseg'&=4\xi^3\qrad+2\xi\qrad+2\xi\qrad\zeta^2+2\xi\langle E_i^\perp,S_{-i}\hat W_i\rangle+2\xi^2\Pcpl .
\end{align*}
\item \emph{Quotient rule.} Expanding,
\begin{multline*}
\Numseg'\Denseg-\Numseg\Denseg'=4\xi^7\qrad+4\xi^5\qrad+4\xi^5\zeta^2\qrad\\
-4\xi^7\qrad-2\xi^5\qrad-2\xi^5\zeta^2\qrad-2\xi^5\langle E_i^\perp,S_{-i}\hat W_i\rangle-2\xi^6\Pcpl,
\end{multline*}
which equals $2\xi^5\big[\qrad(1+\zeta^2)-\langle E_i^\perp,S_{-i}\hat W_i\rangle-\xi\Pcpl\big]$. Since $\Denseg^2=\xi^4(\xi^2+1+\zeta^2)^2$, $\Dt_i'=(\Numseg'\Denseg-\Numseg\Denseg')/\Denseg^2$ is \eqref{eq:A7deriv}.
\item \emph{Tangential term.} Since $E_i^\perp\perp\hat W_i$, $\langle E_i^\perp,S_{-i}\hat W_i\rangle=\langle E_i^\perp,(I_d-\hat W_i\hat W_i^\top)S_{-i}\hat W_i\rangle$. For the unit vector $\hat W_i$ and any $y\in\R^d$, expanding gives $\|y-\langle\hat W_i,y\rangle\hat W_i\|^2=\|y\|^2-2\langle\hat W_i,y\rangle^2+\langle\hat W_i,y\rangle^2=\|y\|^2-\langle\hat W_i,y\rangle^2$. With $y=S_{-i}\hat W_i$: $\|(I_d-\hat W_i\hat W_i^\top)S_{-i}\hat W_i\|^2=\|S_{-i}\hat W_i\|^2-(\hat W_i^\top S_{-i}\hat W_i)^2\le\Rb^2\zeta^2-\zeta^4$, by Tool~\ref{tool:spec} (positive semidefinite case) and Step~2. Note $\zeta^2\le\Rb^2$ by Step~2. With Tool~\ref{tool:cs}: $|\langle E_i^\perp,S_{-i}\hat W_i\rangle|\le\qtan\,\zeta\sqrt{\Rb^2-\zeta^2}$.
\item \emph{Coupling term.} By Tool~\ref{tool:cs}, $|\Pcpl|\le\qcpl\|W_{-i}^\top\hat W_i\|$, and $\|W_{-i}^\top\hat W_i\|^2=\sum_{j\ne i}\langle W_j,\hat W_i\rangle^2=\zeta^2$. So $|\Pcpl|\le\qcpl\zeta$.
\item \emph{Budget.} $E^\top\hat W_i\in\R^n$ has $i$-th entry $\qrad$ and its other entries form $E_{-i}^\top\hat W_i$, so $\qrad^2+\qcpl^2=\|E^\top\hat W_i\|^2\le\|E\|_{\op}^2$ (Tool~\ref{tool:op}). Also $\qrad^2+\qtan^2=\|E_i\|^2=\|E\mathsf e_i\|^2\le\|E\|_{\op}^2$, with $\mathsf e_i\in\R^n$ (Tool~\ref{tool:op}). Hence $|\qrad|\le\|E\|_{\op}$ and $\qtan,\qcpl\le(\|E\|_{\op}^2-\qrad^2)^{1/2}$.
\item \emph{Combination.} From \eqref{eq:A7deriv}, Steps~7--9 and the definitions of the coefficients,
\begin{multline*}
|\Dt_i'|\le\arad|\qrad|+\atan\sqrt{\Rb^2-\zeta^2}\,\qtan+\acpl\qcpl\\
\le\arad|\qrad|+\big(\acpl+\atan\sqrt{\Rb^2-\zeta^2}\big)\sqrt{\|E\|_{\op}^2-\qrad^2}.
\end{multline*}
Cauchy--Schwarz in $\R^2$ (Tool~\ref{tool:cs}) with the vector $(|\qrad|,(\|E\|^2_{\op}-\qrad^2)^{1/2})$, whose norm is $\|E\|_{\op}$, gives \eqref{eq:A7pointwise}; the constraint is Step~2.
\item \emph{$d=1$ or $n=1$.} If $d=1$, every vector orthogonal to $\hat W_i$ is $0$, so $E_i^\perp=0$; if $n=1$, $S_{-i}=0$. In both cases the tangential term in \eqref{eq:A7deriv} vanishes, and Step~10 without it gives $|\Dt_i'|\le\arad|\qrad|+\acpl\qcpl\le(\arad^2+\acpl^2)^{1/2}\|E\|_{\op}$ (Tool~\ref{tool:cs} and Step~9).
\end{steps}
\end{proof}

\begin{lemma}[from the pointwise bound to the segment]\label{lem:A73}\MkOursChk
Let $\lambda=1$ and $L\ge0$. If $|\Dt_i'(\eta)|\le L\|E\|_{\op}$ at every $\eta\in(0,1)$ with $W_i(\eta)\ne0$, then $|\Dt_i(W;1)-\Dt_i(W';1)|\le L\|W-W'\|_{\op}$.
\end{lemma}
\begin{proof}
\begin{steps}
\item \emph{The zero set.} Let $\mathcal Z:=\{\eta\in[0,1]:W_i(\eta)=0\}$, with $W_i(\eta)=W_i+\eta E_i$. If $E_i\ne0$, the map $\eta\mapsto W_i(\eta)$ is injective, so $\mathcal Z$ has at most one point. If $E_i=0$ and $W_i\ne0$, $\mathcal Z=\emptyset$. If $E_i=0$ and $W_i=0$, then $W_i=W'_i=0$, both values are $0$ by D5, and there is nothing to prove.
\item \emph{Continuity.} At $\eta\notin\mathcal Z$, $\Dt_i$ is continuous (Lemma~\ref{lem:A72}, Step~3). At $\eta_\circ\in\mathcal Z$: $\Dt_i(\eta_\circ)=0$, and for $\eta\notin\mathcal Z$, $0\le\Dt_i(\eta)\le\|W_i(\eta)\|^4/\|W_i(\eta)\|^2=\|W_i(\eta)\|^2$, because $W_i^\top SW_i\ge0$ in the denominator of D5. Since $\|W_i(\eta)\|^2\to0$ as $\eta\to\eta_\circ$, $\Dt_i$ is continuous at $\eta_\circ$. So $\Dt_i$ is continuous on $[0,1]$.
\item \emph{Mean value theorem on the pieces.} If $\mathcal Z\cap(0,1)=\emptyset$, $\Dt_i$ is continuous on $[0,1]$ and differentiable on $(0,1)$ (Lemma~\ref{lem:A72}), and Tool~\ref{tool:calc}(b) gives $|\Dt_i(1)-\Dt_i(0)|\le L\|E\|_{\op}$. If $\mathcal Z\cap(0,1)=\{\eta_\circ\}$, apply Tool~\ref{tool:calc}(b) on $[0,\eta_\circ]$ and on $[\eta_\circ,1]$: $|\Dt_i(1)-\Dt_i(0)|\le|\Dt_i(1)-\Dt_i(\eta_\circ)|+|\Dt_i(\eta_\circ)-\Dt_i(0)|\le L\|E\|_{\op}(1-\eta_\circ)+L\|E\|_{\op}\eta_\circ$.
\item \emph{Conclusion.} In both cases $|\Dt_i(1)-\Dt_i(0)|\le L\|E\|_{\op}$. Since $\Dt_i(0)=\Dt_i(W;1)$, $\Dt_i(1)=\Dt_i(W';1)$ and $E=W'-W$, this is $|\Dt_i(W;1)-\Dt_i(W';1)|\le L\|W-W'\|_{\op}$.
\end{steps}
\end{proof}

\begin{lemma}[the coefficients]\label{lem:A74}\MkOursChk
For all $\xi>0$ and $\zeta\ge0$:
(a) $\arad\le3\sqrt3/8$; (b) $\atan\le1/4$, with equality only at $\xi^2=\zeta^2=1/2$; (c) $\acpl\le1/4$; (d) $\arad^2+\acpl^2\le27/64$; (e) $\atan\le\alpha_\star(\xi)\le1/4$.
\end{lemma}
\begin{proof}
Tool used: Tool~\ref{tool:calc}(a),(d).
\begin{steps}
\item \emph{A scalar maximum.} For $u\ge0$, by the quotient rule, $\Fsc'(u)=2\big[(1+u^2)^2-4u^2(1+u^2)\big]/(1+u^2)^4=2(1-3u^2)/(1+u^2)^3$. This is $>0$ on $(0,1/\sqrt3)$ and $<0$ on $(1/\sqrt3,\infty)$, so by Tool~\ref{tool:calc}(d) $\Fsc$ is strictly increasing on $[0,1/\sqrt3]$ and strictly decreasing on $[1/\sqrt3,\infty)$, and $\max\Fsc=\Fsc(1/\sqrt3)=(2/\sqrt3)/(16/9)=9/(8\sqrt3)=3\sqrt3/8$.
\item \emph{(a).} Put $\mathsf w=1+\zeta^2\ge1$ and $u=\xi/\sqrt{\mathsf w}$. Then $\xi^2+\mathsf w=\mathsf w(1+u^2)$, so $\arad=2\sqrt{\mathsf w}\,u\,\mathsf w/(\mathsf w^2(1+u^2)^2)=\mathsf w^{-1/2}\Fsc(u)\le\Fsc(u)\le3\sqrt3/8$ (Step~1).
\item \emph{(b).} Since $\xi^2+\zeta^2\ge2\xi\zeta$, $(\xi^2+\zeta^2+1)^2\ge(2\xi\zeta+1)^2=(2\xi\zeta-1)^2+8\xi\zeta\ge8\xi\zeta$, which is $\atan\le1/4$. Equality needs $\xi=\zeta$ and $2\xi\zeta=1$.
\item \emph{(c).} Since $1+\zeta^2\ge2\zeta$, $(\xi^2+1+\zeta^2)^2\ge(\xi^2+2\zeta)^2\ge4\cdot\xi^2\cdot2\zeta$ (because $(y+y')^2-4yy'=(y-y')^2\ge0$), which is $\acpl\le1/4$.
\item \emph{(d), reduction.} With $\varpi=\xi^2+\zeta^2\ (\ge\xi^2>0)$: $(1+\zeta^2)^2+\xi^2\zeta^2=(1+\varpi-\xi^2)^2+\xi^2(\varpi-\xi^2)=(1+\varpi)^2-(2+\varpi)\xi^2$, so
\[
\arad^2+\acpl^2=\frac{4\xi^2\big[(1+\zeta^2)^2+\xi^2\zeta^2\big]}{(1+\varpi)^4}=\frac{4\mathcal Q(\xi^2)}{(1+\varpi)^4},\qquad \mathcal Q(y):=y\big[(1+\varpi)^2-(2+\varpi)y\big].
\]
\item \emph{(d), maximisation in $y\in(0,\varpi]$ for fixed $\varpi$.} $\mathcal Q$ is a concave quadratic with vertex $y_{\mathrm v}=(1+\varpi)^2/(2(2+\varpi))$ and $\mathcal Q(y_{\mathrm v})=(1+\varpi)^4/(4(2+\varpi))$. Now $y_{\mathrm v}\le\varpi\iff(1+\varpi)^2\le2\varpi(2+\varpi)\iff\varpi^2+2\varpi-1\ge0\iff\varpi\ge\sqrt2-1$.
 \begin{itemize}
 \item If $\varpi\ge\sqrt2-1$: $\arad^2+\acpl^2\le1/(2+\varpi)\le1/(1+\sqrt2)=\sqrt2-1$.
 \item If $\varpi<\sqrt2-1$: $y\le\varpi<y_{\mathrm v}$; $\mathcal Q'(y)=(1+\varpi)^2-2(2+\varpi)y>0$ for $y<y_{\mathrm v}$, so $\mathcal Q$ is increasing on $[0,\varpi]$ (Tool~\ref{tool:calc}(d)) and $\mathcal Q\le\mathcal Q(\varpi)=\varpi\big[(1+\varpi)^2-(2+\varpi)\varpi\big]=\varpi$. Hence $\arad^2+\acpl^2\le4\varpi/(1+\varpi)^4=\Fsc(\sqrt\varpi)^2\le27/64$ (Step~1).
 \end{itemize}
 Finally $\sqrt2-1<27/64$, since $\sqrt2<91/64$ is $2<8281/4096$. So (d) holds in both cases.
\item \emph{(e), maximisation over $\zeta$ for fixed $\xi$.} Let $\mathsf c_0=1+\xi^2$ and $\mathcal K(\zeta)=\zeta/(\mathsf c_0+\zeta^2)^2$. Then $\mathcal K'(\zeta)=\big[(\mathsf c_0+\zeta^2)-4\zeta^2\big]/(\mathsf c_0+\zeta^2)^3=(\mathsf c_0-3\zeta^2)/(\mathsf c_0+\zeta^2)^3$, which is $>0$ for $\zeta<\sqrt{\mathsf c_0/3}$ and $<0$ for $\zeta>\sqrt{\mathsf c_0/3}$; so $\mathcal K$ increases on $[0,\sqrt{\mathsf c_0/3}]$ and decreases on $[\sqrt{\mathsf c_0/3},\infty)$ (Tool~\ref{tool:calc}(d)); $\max\mathcal K=\sqrt{\mathsf c_0/3}\big/(4\mathsf c_0/3)^2=\frac{9}{16\sqrt3}\mathsf c_0^{-3/2}=\frac{3\sqrt3}{16}\mathsf c_0^{-3/2}$. Hence $\atan=2\xi\mathcal K(\zeta)\le\frac{3\sqrt3}8\xi(1+\xi^2)^{-3/2}=\alpha_\star(\xi)$. The bound $\alpha_\star\le1/4$ is Step~8 below.
\item \emph{$\alpha_\star$ in the variable $\upsilon$.} $\xi\mapsto\upsilon=\xi(1+\xi^2)^{-1/2}$ is an increasing bijection $(0,\infty)\to(0,1)$, with $1+\xi^2=1/(1-\upsilon^2)$ and $\xi=\upsilon/\sqrt{1-\upsilon^2}$. Therefore
\[
\alpha_\star=\tfrac{3\sqrt3}{8}\,\upsilon(1-\upsilon^2),\qquad \xi\,\alpha_\star=\tfrac{3\sqrt3}8\,\Omega(\upsilon).
\]
Using $\frac{3\sqrt3}8\cdot\frac2{3\sqrt3}=\frac14$ and the factorisation $\upsilon^3-\upsilon+\frac2{3\sqrt3}=(\upsilon-\upsilon_\star)^2(\upsilon+2\upsilon_\star)$ (expand: $(\upsilon-y)^2(\upsilon+2y)=\upsilon^3-3y^2\upsilon+2y^3$ with $y=1/\sqrt3$),
\begin{equation}\label{eq:A7astar}
\tfrac14-\alpha_\star=\tfrac{3\sqrt3}8(\upsilon-\upsilon_\star)^2(\upsilon+2\upsilon_\star)\ \ge\ \tfrac{3\sqrt3}8\cdot\tfrac2{\sqrt3}(\upsilon-\upsilon_\star)^2=\tfrac34(\upsilon-\upsilon_\star)^2\ \ge0 .
\end{equation}
\end{steps}
\end{proof}

\begin{lemma}[one-sided Lipschitz bound for $\Omega$]\label{lem:A75}\MkOursChk
For $\upsilon\in(0,1)$, $\Omega(\upsilon)-\Omega(\upsilon_\star)\le\sqrt2\,(\upsilon-\upsilon_\star)_+$, and $\frac{3\sqrt3}8\Omega(\upsilon_\star)=\frac1{4\sqrt2}$.
\end{lemma}
\begin{proof}
\begin{steps}
\item $\Omega(\upsilon)=\upsilon^2\sqrt{1-\upsilon^2}$ is continuous on $[0,1)$ and differentiable on $(0,1)$ (product rule, Tool~\ref{tool:calc}(a)), with $\Omega'(\upsilon)=2\upsilon\sqrt{1-\upsilon^2}-\upsilon^3/\sqrt{1-\upsilon^2}=\upsilon(2-3\upsilon^2)/\sqrt{1-\upsilon^2}$. This is $>0$ on $(0,\sqrt{2/3})$ and $<0$ on $(\sqrt{2/3},1)$, so by Tool~\ref{tool:calc}(d) $\Omega$ is increasing on $(0,\sqrt{2/3}]$ and decreasing on $[\sqrt{2/3},1)$; note $\upsilon_\star=1/\sqrt3<\sqrt{2/3}$.
\item If $\upsilon\le\upsilon_\star$: $\Omega(\upsilon)\le\Omega(\upsilon_\star)$ by Step~1.
\item If $\upsilon_\star<\upsilon\le\sqrt{2/3}$: $\Omega$ is continuous on $[\upsilon_\star,\upsilon]$ and differentiable on $(\upsilon_\star,\upsilon)$ (Step~1), so by Tool~\ref{tool:calc}(b), $\Omega(\upsilon)-\Omega(\upsilon_\star)=\Omega'(\bar\upsilon)(\upsilon-\upsilon_\star)$ with $\bar\upsilon\in(\upsilon_\star,\upsilon)$; there $\bar\upsilon\le\sqrt{2/3}$, $0\le2-3\bar\upsilon^2<1$ and $\sqrt{1-\bar\upsilon^2}\ge1/\sqrt3$, so $\Omega'(\bar\upsilon)\le\sqrt{2/3}\cdot\sqrt3=\sqrt2$.
\item If $\upsilon>\sqrt{2/3}$: by Steps~1 and 3, $\Omega(\upsilon)\le\Omega(\sqrt{2/3})\le\Omega(\upsilon_\star)+\sqrt2(\sqrt{2/3}-\upsilon_\star)\le\Omega(\upsilon_\star)+\sqrt2(\upsilon-\upsilon_\star)$.
\item $\Omega(\upsilon_\star)=\frac13\sqrt{2/3}$, and $\frac{3\sqrt3}8\cdot\frac{\sqrt2}{3\sqrt3}=\frac{\sqrt2}8=\frac1{4\sqrt2}$.
\end{steps}
\end{proof}

\begin{theorem}[C7$'$(ii), with the constant $3\sqrt3/8$]\label{thm:A76}\MkOursChk
For all $d,n\ge1$, $\lambda>0$, $i\in[n]$ and $W,W'\in\R^{d\times n}$, with $R=\max(\|W\|_{\op},\|W'\|_{\op})$:
\begin{align}
|\Dt_i(W;\lambda)-\Dt_i(W';\lambda)|&\le\Big(\frac{\sqrt{31}}{8\sqrt\lambda}+\frac R{4\lambda}\Big)\|W-W'\|_{\op},\label{eq:A7ii}\\
|\Dt_i(W;\lambda)-\Dt_i(W';\lambda)|&\le\Big(\frac{3\sqrt3}{8\sqrt\lambda}+\frac R{4\lambda}\Big)\|W-W'\|_{\op}.\label{eq:A7iisharp}
\end{align}
\eqref{eq:A7iisharp} implies \eqref{eq:A7ii}, since $3\sqrt3/8=\sqrt{27}/8<\sqrt{31}/8$. Both hold with $R$ replaced by any $\Rb>0$ with $\Rb\ge R$.
\end{theorem}
\begin{proof}
\begin{steps}
\item \emph{Set-up at $\lambda=1$.} Fix $d,n\ge1$, $i\in[n]$, $W,W'\in\R^{d\times n}$, $\lambda=1$, and an arbitrary $\Rb>0$ with $\Rb\ge R$. Steps~2--4 prove, for this $\Rb$, the two inequalities with $\Rb$ in place of $R$ at $\lambda=1$.
\item \emph{Separate maximisation, giving \eqref{eq:A7ii}.} By Tool~\ref{tool:mink} with $y_1=\arad$, $y_2=\acpl$, $y_3=\atan\sqrt{\Rb^2-\zeta^2}$, then Lemma~\ref{lem:A74}(a)--(c) and $\sqrt{\Rb^2-\zeta^2}\le\Rb$: $\Psi_{\Rb}\le(\arad^2+\acpl^2)^{1/2}+\atan\Rb\le\big((3\sqrt3/8)^2+(1/4)^2\big)^{1/2}+\Rb/4$. Now $(3\sqrt3/8)^2+(1/4)^2=27/64+4/64=31/64$, so $\Psi_{\Rb}\le\sqrt{31}/8+\Rb/4$.
\item \emph{Joint maximisation, giving \eqref{eq:A7iisharp}.} The same Minkowski step with Lemma~\ref{lem:A74}(d) instead of (a) and (c): $\Psi_{\Rb}\le\sqrt{27/64}+\Rb/4=3\sqrt3/8+\Rb/4$.
\item \emph{Conclusion.} By Lemma~\ref{lem:A72}, $|\Dt_i'(\eta)|\le\Psi_{\Rb}(\xi(\eta),\zeta(\eta))\|E\|_{\op}$ wherever $W_i(\eta)\ne0$; by Steps~2--3 this is at most $(\sqrt{31}/8+\Rb/4)\|E\|_{\op}$, resp.\ $(3\sqrt3/8+\Rb/4)\|E\|_{\op}$. Lemma~\ref{lem:A73} gives $|\Dt_i(W;1)-\Dt_i(W';1)|\le(\mathfrak a+\Rb/4)\|W-W'\|_{\op}$ with $\mathfrak a=\sqrt{31}/8$, resp.\ $\mathfrak a=3\sqrt3/8$, for every $\Rb>0$ with $\Rb\ge R$.
\item \emph{All $\lambda$.} Step~4 is the hypothesis of Lemma~\ref{lem:A71} (for these $d,n$, with $\mathfrak b=1/4$, $\mathfrak h=0$). The lemma gives, for every $\lambda>0$ and every $\Rb>0$ with $\Rb\ge R$, $|\Dt_i(W;\lambda)-\Dt_i(W';\lambda)|\le(\mathfrak a/\sqrt\lambda+\Rb/(4\lambda))\|W-W'\|_{\op}$; this is the last sentence of the theorem.
\item \emph{The stated forms.} If $R>0$, take $\Rb=R$ in Step~5: this is \eqref{eq:A7ii} and \eqref{eq:A7iisharp}. If $R=0$, then $W=W'=0$ and both sides of \eqref{eq:A7ii} and \eqref{eq:A7iisharp} are $0$.
\end{steps}
\end{proof}

\begin{corollary}[consequences of Theorem~\ref{thm:A76}]\label{cor:A77}\MkOursChk
\begin{enumerate}[label=(\alph*)]
\item $\Phi(\rho)\le3\sqrt3/8+\rho/4\le\sqrt{31}/8+\rho/4$ for every $\rho>0$. \emph{Derivation:} for a pair in the supremum defining $\Phi(\rho)$, $R\le\rho\sqrt\lambda$; divide \eqref{eq:A7iisharp} by $\|W-W'\|_{\op}$ and multiply by $\sqrt\lambda$.
\item (C7$'$(i).) $\limsup_{\rho\to\infty}\Phi(\rho)/\rho\le1/4$. \emph{Derivation:} by (a), $\Phi(\rho)/\rho\le3\sqrt3/(8\rho)+1/4\to1/4$. With Proposition~\ref{prop:profile}(ii), $\lim_{\rho\to\infty}\Phi(\rho)/\rho=1/4$.
\item (C7(a).) $|\Dt_i(W;\lambda)-\Dt_i(W';\lambda)|\le(2/\sqrt\lambda+R/\lambda)\|W-W'\|_{\op}$, and $|\Dttot(W;\lambda)-\Dttot(W';\lambda)|\le n(2/\sqrt\lambda+R/\lambda)\|W-W'\|_{\op}$. \emph{Derivation:} $\sqrt{31}/8<2$ and $1/4<1$ in \eqref{eq:A7ii}; for the total, sum over $i$ and use the triangle inequality.
\item (C7$'$(ii-w).) The form of \eqref{eq:A7ii} with $2/\sqrt\lambda$ in place of $\sqrt{31}/(8\sqrt\lambda)$. \emph{Derivation:} as in (c), since $\sqrt{31}/8<2$ ($31<256$).
\end{enumerate}
\end{corollary}

\begin{theorem}[C7$'$(iii): the asymptotic additive constant]\label{thm:A78}\MkOursChk
For all $d,n\ge1$, $\lambda>0$, $i\in[n]$, $W,W'$, and every $\Rb\ge R$ with $\Rb>0$,
\begin{equation}\label{eq:A7iii}
|\Dt_i(W;\lambda)-\Dt_i(W';\lambda)|\le\Big(\frac{\Rb}{4\lambda}+\frac1{4\sqrt2\,\sqrt\lambda}+\frac{9}{8\Rb}\Big)\|W-W'\|_{\op} .
\end{equation}
Consequently $\chi(\rho)\le\frac1{4\sqrt2}+\frac9{8\rho}$ for every $\rho>0$, $\limsup_{\rho\to\infty}\chi(\rho)\le1/(4\sqrt2)$, and, with Proposition~\ref{prop:profile}(iii), $\lim_{\rho\to\infty}\chi(\rho)=1/(4\sqrt2)$.
\end{theorem}
\begin{proof}
\begin{steps}
\item By Lemma~\ref{lem:A71}, applied for each fixed $d,n$ (with $\mathfrak a=1/(4\sqrt2)$, $\mathfrak b=1/4$, $\mathfrak h(\Rb)=9/(8\Rb)$, so that $\mathfrak h(\Rb/\sqrt\lambda)/\sqrt\lambda=9/(8\Rb)$), it suffices to prove the bound at $\lambda=1$ for every $\Rb>0$ with $\Rb\ge R$. By Lemmas~\ref{lem:A72} and \ref{lem:A73} it suffices to show $\Psi_{\Rb}(\xi,\zeta)\le\Rb/4+1/(4\sqrt2)+9/(8\Rb)$ for all $\xi>0$, $0\le\zeta\le\Rb$.
\item \emph{The non-radial coefficient.} $\acpl+\atan\sqrt{\Rb^2-\zeta^2}\le\acpl+\atan\Rb=\atan(\Rb+\xi)\le\alpha_\star(\xi)(\Rb+\xi)$, by $\acpl=\xi\atan$ and Lemma~\ref{lem:A74}(e).
\item \emph{Split.} $\alpha_\star(\Rb+\xi)=\Rb/4-\Rb(1/4-\alpha_\star)+\xi\alpha_\star$. By \eqref{eq:A7astar}, $\Rb(1/4-\alpha_\star)\ge\frac34\Rb\,y^2$ with $y:=|\upsilon-\upsilon_\star|$. By Lemma~\ref{lem:A74}, Step~8, and Lemma~\ref{lem:A75}, $\xi\alpha_\star=\frac{3\sqrt3}8\Omega(\upsilon)\le\frac1{4\sqrt2}+\frac{3\sqrt3}8\sqrt2\,y=\frac1{4\sqrt2}+\frac{3\sqrt6}8y$.
\item \emph{Maximisation in $y\ge0$.} $-\frac34\Rb y^2+\frac{3\sqrt6}8y$ is a concave quadratic with maximum at $y=\frac{3\sqrt6/8}{3\Rb/2}=\frac{\sqrt6}{4\Rb}$ and maximal value $\frac{(3\sqrt6/8)^2}{4\cdot3\Rb/4}=\frac{54/64}{3\Rb}=\frac9{32\Rb}$. With Steps~2--3,
\[
0\le\acpl+\atan\sqrt{\Rb^2-\zeta^2}\le\Xi_{\Rb}:=\frac{\Rb}4+\frac1{4\sqrt2}+\frac9{32\Rb}.
\]
\item \emph{Adding the radial coefficient.} By Lemma~\ref{lem:A74}(a), $\arad^2\le27/64$. For $\mathsf r_1\ge0$ and $\mathsf r_2>0$, $(\mathsf r_2+\mathsf r_1/(2\mathsf r_2))^2=\mathsf r_2^2+\mathsf r_1+\mathsf r_1^2/(4\mathsf r_2^2)\ge\mathsf r_2^2+\mathsf r_1$, so $\sqrt{\mathsf r_1+\mathsf r_2^2}\le\mathsf r_2+\mathsf r_1/(2\mathsf r_2)$. With Step~4 and $\Xi_{\Rb}\ge\Rb/4$:
\[
\Psi_{\Rb}\le\sqrt{\tfrac{27}{64}+\Xi_{\Rb}^2}\le\Xi_{\Rb}+\frac{27/64}{2\Xi_{\Rb}}\le\Xi_{\Rb}+\frac{27}{32\Rb}=\frac{\Rb}4+\frac1{4\sqrt2}+\frac{9+27}{32\Rb}=\frac{\Rb}4+\frac1{4\sqrt2}+\frac9{8\Rb}.
\]
\item \emph{Consequences.} For a pair in the supremum defining $\Phi(\rho)$, take $\Rb=\rho\sqrt\lambda\ge R$ in \eqref{eq:A7iii}; multiplying by $\sqrt\lambda/\|W-W'\|_{\op}$ gives $\sqrt\lambda\,|\Dt_i(W;\lambda)-\Dt_i(W';\lambda)|/\|W-W'\|_{\op}\le\rho/4+1/(4\sqrt2)+9/(8\rho)$. So $\chi(\rho)\le1/(4\sqrt2)+9/(8\rho)$, and the $\limsup$ statement follows. With Proposition~\ref{prop:profile}(iii), $\lim_{\rho\to\infty}\chi(\rho)=1/(4\sqrt2)$.
\end{steps}
\end{proof}

\begin{remark}[combined bound, and the structure of \textsf{F2}]\label{rem:A7loss}\MkOursChk
(1) For every $\rho>0$, $\Phi(\rho)\le\rho/4+\min\big(3\sqrt3/8,\ 1/(4\sqrt2)+9/(8\rho)\big)$.
(2) Let $\lambda>0$ and $R_0^2>\lambda/2$, and take the base point $W^\star$ of \textsf{F2} ($\alpha_0=\sqrt{\lambda/2}$, $\iota=1$), column $i=1$ and $E=E^{(1)}$. Then, with the quantities of Definition~\ref{def:C7pnot} formed at $W^\star$ (they do not involve $\lambda$): $\xi^2=\zeta^2=\lambda/2$, $\qrad=0$, $\qtan=\qcpl=1=\|E^{(1)}\|_{\op}$ (a purely tangential $E_1$ and a full coupling part). For $\lambda=1$ this is $\xi^2=\zeta^2=1/2$, the unique maximiser of $\atan$ (Lemma~\ref{lem:A74}(b)).
(3) Let $\lambda=1$ and $\Rb_0:=(R_0^2+1/2)^{1/2}$, an upper bound for $\|W^\star\|_{\op}$. The constant of \eqref{eq:A7iii} at $\Rb=\Rb_0$ exceeds the \textsf{F2} slope $|\mathcal C_1'(0)|$ by $\frac14\big(\Rb_0-\sqrt{R_0^2-1/2}\big)+\frac9{8\Rb_0}$, a number in $\big(0,\frac{11}{8\Rb_0}\big]$. So for large $R_0$ the bound of Theorem~\ref{thm:A78} and the \textsf{F2} slope agree up to $O(1/\Rb_0)$.
\end{remark}
\begin{proof}
\begin{steps}
\item (1): by Corollary~\ref{cor:A77}(a), $\Phi(\rho)\le3\sqrt3/8+\rho/4$; by Theorem~\ref{thm:A78}, $\Phi(\rho)=\chi(\rho)+\rho/4\le\rho/4+1/(4\sqrt2)+9/(8\rho)$. Take the smaller of the two bounds.
\item (2), the column: $W^\star_1=\alpha_0\nphi$, so $\xi=\|W^\star_1\|=\alpha_0$ and $\hat W^\star_1=\nphi$; $\xi^2=\lambda/2$.
\item (2), $\zeta$: here $n=2$ and $S_{-1}=W^\star_2W_2^{\star\top}$ with $W^\star_2=R_0\mathsf e_1$, so $\zeta^2=\langle\nphi,R_0\mathsf e_1\rangle^2=R_0^2\cos^2\phi=\alpha_0^2=\lambda/2$ (Definition~\ref{def:family}: $\cos\phi=\alpha_0/R_0$).
\item (2), the perturbation: $E_1=\nphip$ and $E_{-1}=E_2=-\nphi$ ($\iota=1$). By Lemma~\ref{lem:family}, Step~1, $\qrad=\langle\nphip,\nphi\rangle=0$, $E_1^\perp=\nphip$, $\qtan=\|\nphip\|=1$ and $\qcpl=|\langle-\nphi,\nphi\rangle|=1$; and $\|E^{(1)}\|_{\op}=1$ by Lemma~\ref{lem:family}(ii). The last sentence of (2) is Lemma~\ref{lem:A74}(b).
\item (3): $\Rb_0\ge\|W^\star\|_{\op}$ by Lemma~\ref{lem:family}(ii). By Corollary~\ref{cor:special}(4) with $\lambda=1$, $|\mathcal C_1'(0)|=\frac14\sqrt{R_0^2-1/2}+\frac1{4\sqrt2}$, and the constant of \eqref{eq:A7iii} at $\lambda=1$, $\Rb=\Rb_0$ is $\frac{\Rb_0}4+\frac1{4\sqrt2}+\frac9{8\Rb_0}$. Their difference is the stated number.
\item $\Rb_0-\sqrt{R_0^2-1/2}=\frac{\Rb_0^2-(R_0^2-1/2)}{\Rb_0+\sqrt{R_0^2-1/2}}=\frac1{\Rb_0+\sqrt{R_0^2-1/2}}\in\big(0,\frac1{\Rb_0}\big]$. Hence the difference lies in $\big(0,\frac1{4\Rb_0}+\frac9{8\Rb_0}\big]=\big(0,\frac{11}{8\Rb_0}\big]$.
\end{steps}
\end{proof}

\begin{heuristic}[where the upper bounds may lose]\label{heu:A7loss}\MkSpec
Among the inequalities used in Lemma~\ref{lem:A72} and Theorems~\ref{thm:A76} and \ref{thm:A78} that are strict in general are: the eigenvalue-variance bound (Lemma~\ref{lem:A72}, Step~7); $\sqrt{\Rb^2-\zeta^2}\le\Rb$ (Theorems~\ref{thm:A76} and \ref{thm:A78}, Step~2); the separate use of the two budget constraints (Lemma~\ref{lem:A72}, Step~9); the Minkowski step (Theorem~\ref{thm:A76}, Steps~2--3); $\atan\le\alpha_\star(\xi)$ (Lemma~\ref{lem:A74}(e), used in Theorem~\ref{thm:A78}, Step~2); the linearisation of $\Omega$ (Lemma~\ref{lem:A75}); and $\sqrt{\mathsf r_1+\mathsf r_2^2}\le\mathsf r_2+\mathsf r_1/(2\mathsf r_2)$ (Theorem~\ref{thm:A78}, Step~5). The list is not claimed to be complete. Remark~\ref{rem:A7loss}(2)--(3) suggests that for large $\Rb$ these losses are of lower order at the structure of \textsf{F2}; we do not prove which of them must be removed to settle C7$''$ on the open range.
\end{heuristic}

\begin{remark}[\textsf{F1} and \textsf{F2} are asymptotically extremal]\label{rem:F1F2}\MkOursChk
(1) No bound $\Phi(\rho)\le a+c\rho$ for all $\rho>0$, with real $a$ and $c<1/4$, holds; and $\lim_{\rho\to\infty}\Phi(\rho)/\rho=1/4$, approached by \textsf{F1} pairs: the lower bound $\frac14\sqrt{\rho^2-1}$ of Proposition~\ref{prop:profile}(ii) is obtained from \textsf{F1} pairs alone. (2) No bound $\chi(\rho)\le c'+o(1)$ as $\rho\to\infty$ with $c'<1/(4\sqrt2)$ holds; and for $\rho>1$, $\frac1{4\sqrt2}-\frac1{4\rho}\le\chi(\rho)\le\frac1{4\sqrt2}+\frac9{8\rho}$, where the lower bound is obtained from \textsf{F2} pairs alone. The gap between the two bounds is $\frac{11}{8\rho}$.
\end{remark}
\begin{proof}
\begin{steps}
\item (1): if $\Phi(\rho)\le a+c\rho$ for all $\rho$, then by Proposition~\ref{prop:profile}(ii), $\frac14\sqrt{1-1/\rho^2}\le a/\rho+c$ for $\rho>1$; letting $\rho\to\infty$ gives $c\ge1/4$. The limit is Corollary~\ref{cor:A77}(b).
\item (2): if $\limsup_{\rho\to\infty}\chi(\rho)\le c'$, then Proposition~\ref{prop:profile}(iii) ($\liminf\chi\ge1/(4\sqrt2)$) gives $c'\ge1/(4\sqrt2)$. The two-sided bound is Proposition~\ref{prop:profile}(iii) and Theorem~\ref{thm:A78}.
\item The gap is $\frac9{8\rho}+\frac1{4\rho}=\frac{11}{8\rho}$.
\item In the proof of Proposition~\ref{prop:profile}(ii)--(iii), Steps~1--4 use only the pairs $(W^\star,W^\star(\eta))$ of \textsf{F1} ($\iota=0$) and \textsf{F2} ($\iota=1$) at $\lambda=1$, and the lower bounds arise as limits $\eta\to0$, $R_0\uparrow\sqrt{\rho^2-1/2}$; this is the meaning of ``approached''.
\end{steps}
\end{proof}

\begin{remark}[comparison with the second proof]\label{rem:A7vsB7}\MkOurs
Compare \eqref{eq:A7iisharp} with \eqref{eq:B7main} of Theorem~\ref{thm:B7} (Appendix~\ref{app:B7}), whose right side is $\big(\frac{1/\sqrt2+1/4}{\sqrt\lambda}+\frac{R_{\min}}{4\lambda}\big)\|W-W'\|_{\op}$. Neither right side is at most the other for all pairs: when $R_{\min}=R$ (for instance $W'=-W\ne0$), the right side of \eqref{eq:A7iisharp} is strictly smaller; when $R_{\min}=0<R$ (for instance $W=0$) and $R/\sqrt\lambda>4(1/\sqrt2+1/4-3\sqrt3/8)$, that of \eqref{eq:B7main} is strictly smaller. Both have coefficient $1/4$, of $R/\lambda$ and of $R_{\min}/\lambda$ respectively.
\end{remark}
\begin{proof}
\begin{steps}
\item $3\sqrt3/8<1/\sqrt2+1/4$: $(3\sqrt3)^2=27<27.04=5.2^2$ gives $3\sqrt3/8<0.65$, and $2\cdot0.7^2=0.98<1$ gives $1/\sqrt2>0.7$, so $1/\sqrt2+1/4>0.95$ (Tool~\ref{tool:mono}, square root).
\item If $R_{\min}=R$, the two right sides differ only in the additive constants, and Step~1 gives the claim. For $W'=-W\ne0$, $\|W'\|_{\op}=\|W\|_{\op}$ (Tool~\ref{tool:op}), so $R_{\min}=R$.
\item If $R_{\min}=0<R$, the right side of \eqref{eq:B7main} is $\frac{1/\sqrt2+1/4}{\sqrt\lambda}\|W-W'\|_{\op}$, and it is strictly smaller than that of \eqref{eq:A7iisharp} iff $\frac{1/\sqrt2+1/4}{\sqrt\lambda}<\frac{3\sqrt3}{8\sqrt\lambda}+\frac R{4\lambda}$, i.e.\ iff $R/\sqrt\lambda>4(1/\sqrt2+1/4-3\sqrt3/8)$. For $W=0$ and $\|W'\|_{\op}$ large enough, both conditions hold.
\end{steps}
\end{proof}

\subsection{The additive constant: partial results on C7$''$}\label{ssec:C7pp}

Searches (\S\ref{sec:T5}) suggest that the worst additive constant is the scalar one: the largest value of $\Phi(\rho)-\rho/4$ comes from $d=n=1$. We prove this for small and large $\rho$ and for $d=1$ or $n=1$; the middle range is open.

\begin{definition}[scalar excess]\label{def:chis}\MkDef
For $\rho>0$, $\chi_s(\rho):=\Fsc(\rho)-\rho/4=\frac{2\rho}{(1+\rho^2)^2}-\frac\rho4$, and $\chi^\star_s:=\max_{0<\rho\le1/\sqrt3}\chi_s(\rho)$ (the maximum exists by Derivation~\ref{der:chis}). Also $\rho_1:=\big((\sqrt5-1)/4\big)^{1/2}$ ($\approx0.5559$), and
\[
\rho_2:=\frac{9}{8\big(\chi^\star_s-1/(4\sqrt2)\big)},\qquad \rho_c:=\frac{9}{8\big(103/200-1/(4\sqrt2)\big)}
\]
($\rho_2\approx3.326$; both are positive and $\rho_2\le\rho_c<10/3$ by Theorem~\ref{thm:A84}; the decimals are for orientation only).
\end{definition}

\begin{conjecture}[C7$''$: the additive constant is set by the scalar case]\label{conj:C7pp}\MkConj
$\Phi(\rho)\le\chi^\star_s+\rho/4$ for every $\rho>0$; equivalently, for all $d,n\ge1$, $\lambda>0$, $i\in[n]$ and $W,W'$, $|\Dt_i(W;\lambda)-\Dt_i(W';\lambda)|\le(\chi^\star_s/\sqrt\lambda+R/(4\lambda))\|W-W'\|_{\op}$. Together with $\chi^\star\ge\chi^\star_s$ (Proposition~\ref{prop:profile}(iv)), C7$''$ says $\chi^\star=\chi^\star_s$.
\end{conjecture}

\begin{derivation}[the value of $\chi^\star_s$]\label{der:chis}\MkOursChk
The cubic $y^3+3y^2+27y-7$ has exactly one real root $y_0$, and $y_0\in(0,1/3)$. The function $\chi_s$ is strictly increasing on $(0,\sqrt{y_0}]$ and strictly decreasing on $[\sqrt{y_0},1/\sqrt3]$, so the maximum defining $\chi^\star_s$ exists and $\chi^\star_s=\chi_s(\sqrt{y_0})$. Numerically $y_0\approx0.2516$, $\sqrt{y_0}\approx0.5016$ and $\chi^\star_s\approx0.515005$; these decimals are not used in any proof. The proved bound is $\chi^\star_s\ge103/200$ (Lemma~\ref{lem:A80}(a)).
\end{derivation}
\begin{proof}
\begin{steps}
\item \emph{Sign of $\chi_s'$.} By Lemma~\ref{lem:A74}, Step~1, $\chi_s'(\rho)=2(1-3\rho^2)/(1+\rho^2)^3-1/4$. Multiplying by $4(1+\rho^2)^3>0$ and putting $y=\rho^2$: $\chi_s'(\rho)$ has the sign of $8(1-3y)-(1+y)^3=7-27y-3y^2-y^3=-(y^3+3y^2+27y-7)$.
\item \emph{The cubic.} Its derivative $3y^2+6y+27=3(y+1)^2+24$ is $>0$, so it is strictly increasing on $\R$ (Tool~\ref{tool:calc}(d)) and has at most one real root. Its value is $-7<0$ at $y=0$ and $1/27+1/3+9-7>0$ at $y=1/3$; by Tool~\ref{tool:calc}(h) it has a root $y_0\in(0,1/3)$, which is the only one.
\item \emph{Monotonicity of $\chi_s$.} By Steps~1--2, $\chi_s'>0$ for $0<\rho<\sqrt{y_0}$ and $\chi_s'<0$ for $\sqrt{y_0}<\rho\le1/\sqrt3$. By Tool~\ref{tool:calc}(d), $\chi_s$ is strictly increasing on $(0,\sqrt{y_0}]$ and strictly decreasing on $[\sqrt{y_0},1/\sqrt3]$; so its maximum over $(0,1/\sqrt3]$ is attained at $\sqrt{y_0}\in(0,1/\sqrt3)$.
\end{steps}
\end{proof}

\begin{lemma}[numerical inequalities]\label{lem:A80}\MkOursChk
(a) $\chi^\star_s\ge103/200$. (b) $7/(8\sqrt3)\le103/200$. (c) $(\sqrt2-1)^{1/2}\le\frac{103}{200}+\frac14(\sqrt2-1)^{1/2}$. (d) $\frac1{4\sqrt2}+\frac{27}{80}<\frac{103}{200}$. (e) $\rho_1<1/\sqrt3$.
\end{lemma}
\begin{proof}
\begin{steps}
\item (a) $1/2\le1/\sqrt3$ (as $1/4\le1/3$), and $\chi_s(1/2)=1/(5/4)^2-1/8=16/25-1/8=103/200$.
\item (b) Both sides are positive; squaring and clearing denominators, it is $1400\le824\sqrt3$, i.e.\ $1400^2=1\,960\,000\le824^2\cdot3=678\,976\cdot3=2\,036\,928$.
\item (c) It is $\frac34(\sqrt2-1)^{1/2}\le\frac{103}{200}$, i.e.\ $(\sqrt2-1)^{1/2}\le\frac{103}{150}$, i.e.\ $\sqrt2-1\le\frac{10609}{22500}$, i.e.\ $\sqrt2\le\frac{33109}{22500}$. This holds because $\sqrt2<1.415$ (as $1.415^2=2.002225>2$) and $33109/22500>1.47$.
\item (d) $\frac{103}{200}-\frac{27}{80}=\frac{412-270}{800}=\frac{71}{400}$, and $\frac1{4\sqrt2}<\frac{71}{400}$ is $400<284\sqrt2$, i.e.\ (both sides positive, Tool~\ref{tool:mono}) $160\,000<284^2\cdot2=80\,656\cdot2=161\,312$.
\item (e) $\rho_1^2=(\sqrt5-1)/4<1/3\iff3\sqrt5<7\iff45<49$.
\end{steps}
\end{proof}

\begin{lemma}[scalar comparison]\label{lem:A81}\MkOursChk
For $0\le u\le\Rb$: $\Fsc(u)\le\chi^\star_s+\Rb/4$.
\end{lemma}
\begin{proof}
\begin{steps}
\item If $\Rb\le1/\sqrt3$: $\Fsc$ is increasing on $[0,1/\sqrt3]$ (Lemma~\ref{lem:A74}, Step~1), so $\Fsc(u)\le\Fsc(\Rb)=\chi_s(\Rb)+\Rb/4\le\chi^\star_s+\Rb/4$ by the definition of $\chi^\star_s$.
\item If $\Rb>1/\sqrt3$: $\Fsc(u)\le3\sqrt3/8=\frac{9}{8\sqrt3}=\frac{7}{8\sqrt3}+\frac1{4\sqrt3}\le\frac{103}{200}+\frac{\Rb}4\le\chi^\star_s+\frac{\Rb}{4}$, by Lemma~\ref{lem:A80}(b) and (a).
\end{steps}
\end{proof}

\begin{theorem}[C7$''$ when $d=1$ or $n=1$]\label{thm:A82}\MkOursChk
If $d=1$ or $n=1$, then for all $\lambda>0$, $i\in[n]$ and $W,W'\in\R^{d\times n}$,
\[
|\Dt_i(W;\lambda)-\Dt_i(W';\lambda)|\le\Big(\frac{\chi^\star_s}{\sqrt\lambda}+\frac R{4\lambda}\Big)\|W-W'\|_{\op} .
\]
\end{theorem}
\begin{proof}
\begin{steps}
\item Fix $d,n$ with $d=1$ or $n=1$. By Lemma~\ref{lem:A71} (stated for fixed $d,n$, with $\mathfrak a=\chi^\star_s$, $\mathfrak b=1/4$, $\mathfrak h=0$) it suffices to prove the bound at $\lambda=1$ with $R$ replaced by an arbitrary $\Rb>0$ with $\Rb\ge R$; Steps~2--5 do this. At the end, take $\Rb=R$ if $R>0$; if $R=0$, then $W=W'=0$ and both sides are $0$.
\item By Lemma~\ref{lem:A72} (last clause), at each $\eta$ with $W_i(\eta)\ne0$, $|\Dt_i'(\eta)|\le(\arad^2+\acpl^2)^{1/2}\|E\|_{\op}$, with $\varpi=\xi^2+\zeta^2\le\Rb^2$.
\item By Lemma~\ref{lem:A74}, Step~6: if $\varpi<\sqrt2-1$, then $(\arad^2+\acpl^2)^{1/2}\le\Fsc(\sqrt\varpi)$ with $\sqrt\varpi\le\Rb$, which is $\le\chi^\star_s+\Rb/4$ by Lemma~\ref{lem:A81}.
\item If $\varpi\ge\sqrt2-1$, then $(\arad^2+\acpl^2)^{1/2}\le(\sqrt2-1)^{1/2}$ (Lemma~\ref{lem:A74}, Step~6), and $\Rb\ge\sqrt\varpi\ge(\sqrt2-1)^{1/2}$. By Lemma~\ref{lem:A80}(c) and (a), $(\sqrt2-1)^{1/2}\le\frac{103}{200}+\frac14(\sqrt2-1)^{1/2}\le\chi^\star_s+\Rb/4$.
\item Lemma~\ref{lem:A73} with $L=\chi^\star_s+\Rb/4$ gives the claim.
\end{steps}
\end{proof}

\begin{theorem}[small radius: the profile equals the scalar slope]\label{thm:A83}\MkOursChk
For all $d,n\ge1$, $\lambda>0$, $i$, and $W,W'$ with $R\le\rho_1\sqrt\lambda$,
\begin{equation}\label{eq:A8small}
|\Dt_i(W;\lambda)-\Dt_i(W';\lambda)|\le\frac{2R/\sqrt\lambda}{\big(1+R^2/\lambda\big)^2}\cdot\frac{\|W-W'\|_{\op}}{\sqrt\lambda}\le\Big(\frac{\chi^\star_s}{\sqrt\lambda}+\frac{R}{4\lambda}\Big)\|W-W'\|_{\op} ,
\end{equation}
and $\Phi(\rho)=2\rho/(1+\rho^2)^2$ for $0<\rho\le\rho_1$. In particular C7$''$ holds for every $\rho\in(0,\rho_1]$, and the pairwise inequality of C7$''$ holds for every pair with $R\le\rho_1\sqrt\lambda$.
\end{theorem}
\begin{proof}
\begin{steps}
\item By the scaling of Lemma~\ref{lem:A71}, Steps~1--2 (applied with $c=\lambda^{-1/2}$), take $\lambda=1$; let $0<\Rb\le\rho_1$ with $\Rb\ge R$. It suffices to show $\Psi_{\Rb}(\xi,\zeta)\le\Fsc(\Rb)$ whenever $\xi>0$, $\zeta\ge0$, $\xi^2+\zeta^2\le\Rb^2$ (Lemmas~\ref{lem:A72} and \ref{lem:A73}).
\item \emph{Expression.} Since $\acpl=\xi\atan$, with $\mathsf p:=\xi^2+1+\zeta^2$,
\[
\Psi_{\Rb}^2=\frac{4\xi^2}{\mathsf p^4}\Big[(1+\zeta^2)^2+\zeta^2\big(\xi+\sqrt{\Rb^2-\zeta^2}\big)^2\Big].
\]
\item \emph{Removing $\xi$ from the bracket.} $\xi^2\le\Rb^2-\zeta^2$, so $(\xi+\sqrt{\Rb^2-\zeta^2})^2\le4(\Rb^2-\zeta^2)$ and the bracket is at most $\mathcal B(\zeta):=(1+\zeta^2)^2+4\zeta^2(\Rb^2-\zeta^2)=1+(2+4\Rb^2)\zeta^2-3\zeta^4\ (\ge0)$.
\item \emph{Maximisation in $y=\xi^2\in(0,\Rb^2-\zeta^2]$ for fixed $\zeta$.} $\frac{\mathrm d}{\mathrm dy}\frac{y}{(1+\zeta^2+y)^4}=\frac{1+\zeta^2-3y}{(1+\zeta^2+y)^5}$, which is $\ge0$ for $y\le(1+\zeta^2)/3$. Since $y\le\Rb^2\le\rho_1^2<1/3$ (Lemma~\ref{lem:A80}(e)), the function is increasing on the range, and its maximum is at $y=\Rb^2-\zeta^2$, where $1+\zeta^2+y=1+\Rb^2$. (The range is non-empty, since $\xi>0$ forces $\zeta^2<\Rb^2$.) Hence $\Psi_{\Rb}^2\le4(\Rb^2-\zeta^2)\mathcal B(\zeta)/(1+\Rb^2)^4$.
\item \emph{The one-variable inequality.} Put $\mathsf v_\zeta:=\zeta^2\in[0,\Rb^2)$ and $\mathsf c_1:=2+4\Rb^2$. Expanding, $(\Rb^2-\mathsf v_\zeta)(1+\mathsf c_1\mathsf v_\zeta-3\mathsf v_\zeta^2)-\Rb^2=\mathsf v_\zeta\,\mathcal Y(\mathsf v_\zeta)$ with $\mathcal Y(\mathsf v_\zeta):=(\mathsf c_1\Rb^2-1)-(3\Rb^2+\mathsf c_1)\mathsf v_\zeta+3\mathsf v_\zeta^2$. $\mathcal Y$ is a convex quadratic, so on $[0,\Rb^2]$ it is at most $\max(\mathcal Y(0),\mathcal Y(\Rb^2))$. Here $\mathcal Y(\Rb^2)=\mathsf c_1\Rb^2-1-3\Rb^4-\mathsf c_1\Rb^2+3\Rb^4=-1$, and $\mathcal Y(0)=4\Rb^4+2\Rb^2-1\le0$ because $\Rb^2\le\rho_1^2=(\sqrt5-1)/4$, the positive root of $4y^2+2y-1$. So $(\Rb^2-\mathsf v_\zeta)\mathcal B\le\Rb^2$.
\item \emph{Conclusion of the bound.} By Steps~4--5, $\Psi_{\Rb}^2\le4\Rb^2/(1+\Rb^2)^4=\Fsc(\Rb)^2$. Lemmas~\ref{lem:A72}--\ref{lem:A73} then give $|\Dt_i(W;1)-\Dt_i(W';1)|\le\Fsc(\Rb)\|W-W'\|_{\op}$. If $R>0$, take $\Rb=R$: this is the first inequality of \eqref{eq:A8small} at $\lambda=1$; the second is $\Fsc(R)=\chi_s(R)+R/4\le\chi^\star_s+R/4$, valid since $0<R\le\rho_1<1/\sqrt3$ (Definition~\ref{def:chis}, Lemma~\ref{lem:A80}(e)). If $R=0$, then $W=W'=0$ and all three members of \eqref{eq:A8small} are $0$.
\item \emph{The profile.} For a pair in the supremum defining $\Phi(\rho)$ with $\rho\le\rho_1$, take $\Rb=\rho$ (after scaling to $\lambda=1$): $\Phi(\rho)\le\Fsc(\rho)$. With Proposition~\ref{prop:profile}(iv), $\Phi(\rho)=\Fsc(\rho)$. Since $\rho\le\rho_1<1/\sqrt3$, $\Fsc(\rho)=\chi_s(\rho)+\rho/4\le\chi^\star_s+\rho/4$ (Definition~\ref{def:chis}), which is C7$''$ at $\rho$.
\end{steps}
\end{proof}

\begin{theorem}[large radius]\label{thm:A84}\MkOursChk
The numbers $\rho_2,\rho_c$ of Definition~\ref{def:chis} are well defined and $0<\rho_2\le\rho_c<10/3$. For all $d,n\ge1$, $\lambda>0$, $i$, and $W,W'$ with $R\ge\rho_2\sqrt\lambda$, $|\Dt_i(W;\lambda)-\Dt_i(W';\lambda)|\le\big(\chi^\star_s/\sqrt\lambda+R/(4\lambda)\big)\|W-W'\|_{\op}$. Moreover $\Phi(\rho)\le\chi^\star_s+\rho/4$ for every $\rho\ge\rho_2$. In particular both hold when $R\ge\rho_c\sqrt\lambda$, respectively $\rho\ge\rho_c$, a threshold computed from explicit numbers.
\end{theorem}
\begin{proof}
\begin{steps}
\item \emph{The thresholds.} By Lemma~\ref{lem:A80}(d), $\frac{103}{200}-\frac1{4\sqrt2}>\frac{27}{80}>0$, and by Lemma~\ref{lem:A80}(a), $\chi^\star_s-\frac1{4\sqrt2}\ge\frac{103}{200}-\frac1{4\sqrt2}$. So both denominators in Definition~\ref{def:chis} are positive, $\rho_2\le\rho_c$ (larger denominator), and $\rho_c<\frac{9}{8\cdot27/80}=\frac{720}{216}=\frac{10}3$.
\item \emph{Pairwise form.} Let $R\ge\rho_2\sqrt\lambda$; then $R>0$. Theorem~\ref{thm:A78} with $\Rb=R$ gives the constant $\frac R{4\lambda}+\frac1{4\sqrt2\sqrt\lambda}+\frac9{8R}$. By the choice of $R$ and the definition of $\rho_2$, $\frac9{8R}\le\frac{9}{8\rho_2\sqrt\lambda}=\frac{\chi^\star_s-1/(4\sqrt2)}{\sqrt\lambda}$. So the constant is at most $\frac{\chi^\star_s}{\sqrt\lambda}+\frac R{4\lambda}$.
\item \emph{Profile form.} For $\rho\ge\rho_2$, Theorem~\ref{thm:A78} gives $\chi(\rho)\le\frac1{4\sqrt2}+\frac9{8\rho}\le\frac1{4\sqrt2}+\frac9{8\rho_2}=\chi^\star_s$, i.e.\ $\Phi(\rho)\le\chi^\star_s+\rho/4$.
\item \emph{Explicit threshold.} $R\ge\rho_c\sqrt\lambda$ implies $R\ge\rho_2\sqrt\lambda$, and $\rho\ge\rho_c$ implies $\rho\ge\rho_2$, by Step~1.
\end{steps}
\end{proof}

\begin{corollary}[C7$''$: what is proved]\label{cor:C7pp}\MkOurs
(a) $\Phi(\rho)\le\chi^\star_s+\rho/4$ holds for $\rho\in(0,\rho_1]\cup[\rho_2,\infty)$. (b) The pairwise inequality of C7$''$ holds for all pairs with $d=1$ or $n=1$, and for all pairs with $R\le\rho_1\sqrt\lambda$ or $R\ge\rho_2\sqrt\lambda$. (c) The upper bounds on $\Phi$ proved in this document are $\Phi(\rho)\le3\sqrt3/8+\rho/4$ (Corollary~\ref{cor:A77}(a)), $\Phi(\rho)\le\rho/4+1/(4\sqrt2)+9/(8\rho)$ (Theorem~\ref{thm:A78}) and $\Phi(\rho)\le1/\sqrt2+1/4+\rho/4$ (Theorem~\ref{thm:B7}). For every $\rho\in(\rho_1,\rho_2)$, each of the three right sides is strictly larger than $\chi^\star_s+\rho/4$; and $\chi^\star_s<3\sqrt3/8$.
\end{corollary}
\begin{proof}
\begin{steps}
\item (a): Theorem~\ref{thm:A83} for $\rho\in(0,\rho_1]$ and Theorem~\ref{thm:A84} for $\rho\ge\rho_2$.
\item (b): Theorem~\ref{thm:A82} for $d=1$ or $n=1$; Theorem~\ref{thm:A83} for $R\le\rho_1\sqrt\lambda$; Theorem~\ref{thm:A84} for $R\ge\rho_2\sqrt\lambda$.
\item $\chi^\star_s<3\sqrt3/8$: by Derivation~\ref{der:chis}, $\chi^\star_s=\chi_s(\sqrt{y_0})=\Fsc(\sqrt{y_0})-\sqrt{y_0}/4<\Fsc(\sqrt{y_0})\le3\sqrt3/8$ (Lemma~\ref{lem:A74}, Step~1), since $\sqrt{y_0}>0$.
\item First bound: by Step~3, $3\sqrt3/8+\rho/4>\chi^\star_s+\rho/4$ for every $\rho>0$.
\item Third bound: $1/\sqrt2+1/4>3\sqrt3/8$ (Remark~\ref{rem:A7vsB7}, Step~1), so with Step~3, $1/\sqrt2+1/4+\rho/4>\chi^\star_s+\rho/4$ for every $\rho>0$.
\item Second bound: for $0<\rho<\rho_2$, $\frac9{8\rho}>\frac9{8\rho_2}=\chi^\star_s-\frac1{4\sqrt2}$ (Definition~\ref{def:chis}; the denominator is positive by Theorem~\ref{thm:A84}, Step~1). Adding $\rho/4+1/(4\sqrt2)$ gives $\rho/4+1/(4\sqrt2)+9/(8\rho)>\chi^\star_s+\rho/4$. In particular this holds on $(\rho_1,\rho_2)$.
\end{steps}
\end{proof}

\noindent So on $(\rho_1,\rho_2)$ none of the upper bounds proved here gives C7$''$; for $d,n\ge2$ it is open there (Conjecture~\ref{conj:C7ppopen}).

\begin{conjecture}[C7$''$ on the open range]\label{conj:C7ppopen}\MkConj
For all $d,n\ge2$, $\lambda>0$, $i\in[n]$ and $W,W'$ with $\rho_1\sqrt\lambda<R<\rho_2\sqrt\lambda$, $|\Dt_i(W;\lambda)-\Dt_i(W';\lambda)|\le(\chi^\star_s/\sqrt\lambda+R/(4\lambda))\|W-W'\|_{\op}$; and $\Phi(\rho)\le\chi^\star_s+\rho/4$ for every $\rho\in(\rho_1,\rho_2)$. With Corollary~\ref{cor:C7pp}, this is equivalent to Conjecture~\ref{conj:C7pp}.
\end{conjecture}

\begin{conjecture}[a sufficient two-variable inequality]\label{conj:twovar}\MkConj
For every $\Rb\in(\rho_1,\rho_2)$,
\[
\sup\big\{\Psi_{\Rb}(\xi,\zeta):\ \xi>0,\ \zeta\ge0,\ \xi^2+\zeta^2\le\Rb^2\big\}\le\chi^\star_s+\Rb/4 .
\]
By Lemmas~\ref{lem:A71}--\ref{lem:A73} (with $L=\chi^\star_s+\Rb/4$, as in the proof of Theorem~\ref{thm:A83}), this inequality implies Conjecture~\ref{conj:C7ppopen}; it is sufficient, not necessary. It is a candidate for a computer-assisted check (interval arithmetic), or for Lean later. If it fails somewhere, the method of \S\ref{ssec:C7p} cannot give C7$''$ there, and some of the losses listed in Heuristic~\ref{heu:A7loss} would have to be removed.
\end{conjecture}

\begin{heuristic}[spot evaluations]\label{heu:spot}\MkSpec
Spot evaluations by hand (not a proof) of the supremum in Conjecture~\ref{conj:twovar} at $\Rb\in\{0.6,1,2\}$ gave margins from about $0.01$ to about $0.2$, with the smallest margin near $\Rb\approx0.56$--$0.6$ and $\xi^2\approx1/3$. Dropping the constraint $\xi^2+\zeta^2\le\Rb^2$ makes the margin there nearly vanish, so the constraint is needed.
\end{heuristic}

\section{T4: certificates for the feature recovery rate}\label{sec:T4}

$\FRR_t$ counts features whose best atom has absolute cosine at least $t$ with them; it jumps whenever a score crosses $t$. This section gives checkable conditions under which the verdict ``feature $i$ is recovered'' or ``is not recovered'' is the same for two dictionaries $A$ and $A'$ (for instance two training seeds), and bounds the change of $\FRR_t$ by the number of features without such a certificate. The tools are angles between lines: the scores are Lipschitz in a line-Hausdorff distance (C8), and above the threshold $\cos(\theta_V/2)$ an atom can serve at most one feature (C9(a)), which turns per-atom counts into per-feature counts.

\subsection{Line angles and chords}\label{ssec:angle}

\begin{definition}[line angle, local distances]\label{def:T4not}\MkDef
For nonzero $x,y\in\R^d$, the \emph{line angle} is
\begin{equation}\label{eq:lineangle}
\angle(x,y):=\arccos|\cos(x,y)|\in[0,\pi/2].
\end{equation}
The \emph{chord function} is $\ch(\vartheta):=2\sin(\vartheta/2)$ for $\vartheta\in[0,\pi/2]$. For $A\in\cA_M$, $A'\in\cA_{M'}$, $i\in[n]$ and $\kappa\in(0,1]$:
\begin{itemize}[itemsep=1pt]
\item $e^\angle_m(A\to A'):=\min_{m'}\angle(a_m,a'_{m'})$ and $\angle_i(A):=\arccos c_i(A)$;
\item the \emph{near set} $P_i(A;\kappa):=\{m\in[M]:|\cos(a_m,v_i)|\ge\kappa\}$;
\item $\delta_i(A\to A';\kappa):=\max_{m\in P_i(A;\kappa)}e_m(A\to A')$ and $\delta^\angle_i(A\to A';\kappa):=\max_{m\in P_i(A;\kappa)}e^\angle_m(A\to A')$, each $0$ if $P_i(A;\kappa)=\emptyset$; $\delta_i(A,A';\kappa):=\max\big(\delta_i(A\to A';\kappa),\delta_i(A'\to A;\kappa)\big)$, and likewise $\delta^\angle_i(A,A';\kappa)$;
\item the \emph{recovered set} $\Rec_t(A):=\{i\in[n]:c_i(A)\ge t\}$, so $n\,\FRR_t(A)=\#\Rec_t(A)$ by D7;
\item a \emph{maximiser of $c_i(A)$} is an $m\in[M]$ with $|\cos(a_m,v_i)|=c_i(A)$. One exists because $[M]$ is finite and nonempty.
\end{itemize}
\end{definition}

\begin{lemma}[angle reformulation]\label{lem:T41}\MkOursChk
For nonzero $x,y$ and nonzero scalars $\mathsf k,\mathsf k'$:
(i) $|\cos(x,y)|=\cos\angle(x,y)$, $\angle(x,y)=\angle(y,x)$, $|\cos(\mathsf kx,\mathsf k'y)|=|\cos(x,y)|$ and $\angle(\mathsf kx,\mathsf k'y)=\angle(x,y)$;
(ii) for $u\in[0,1]$ and $\vartheta\in[0,\pi/2]$: $\cos\vartheta\ge u\iff\vartheta\le\arccos u$, and $\cos\vartheta>u\iff\vartheta<\arccos u$;
(iii) for every $A\in\cA_M$ and $i\in[n]$: $\angle_i(A)=\min_m\angle(a_m,v_i)$, and $\angle_i(A)=\angle(a_m,v_i)$ for every maximiser $m$ of $c_i(A)$.
\end{lemma}
\begin{proof}
\begin{steps}
\item $|\cos(x,y)|\in[0,1]$ by Tool~\ref{tool:cs}, and $\cos(\arccos y')=y'$ for $y'\in[-1,1]$ (Tool~\ref{tool:trig}); this gives the first identity in (i). Symmetry holds because $\cos(x,y)=\cos(y,x)$ (D0).
\item $\cos(\mathsf kx,\mathsf k'y)=\mathsf k\mathsf k'\langle x,y\rangle/(|\mathsf k||\mathsf k'|\|x\|\|y\|)=\pm\cos(x,y)$ (D0), so $|\cos(\mathsf kx,\mathsf k'y)|=|\cos(x,y)|$, and hence $\angle$ is unchanged.
\item For (ii): $u\in[0,1]$ gives $\arccos u\in[0,\pi/2]$. Both $\vartheta$ and $\arccos u$ lie in $[0,\pi]$, where $\cos$ is strictly decreasing (Tool~\ref{tool:trig}); and $u=\cos(\arccos u)$. A strictly decreasing function reverses both $\le$ and $<$.
\item For (iii): $\arccos$ is decreasing (Tool~\ref{tool:trig}), so by Tool~\ref{tool:mono}
\[
\arccos\max_m|\cos(a_m,v_i)|=\min_m\arccos|\cos(a_m,v_i)|,
\]
i.e.\ $\angle_i(A)=\min_m\angle(a_m,v_i)$ by D7 and \eqref{eq:lineangle}. If $m$ is a maximiser, $\angle(a_m,v_i)=\arccos c_i(A)=\angle_i(A)$.
\end{steps}
\end{proof}

The certificates below use the triangle inequality for the line angle. We prove it from the spherical triangle inequality, which we prove from the addition formula for the cosine.

\begin{lemma}[triangle inequality for the line angle]\label{lem:triangle}\MkOursChk
(i) For unit vectors $x,y,z\in\R^d$: $\arccos\langle x,z\rangle\le\arccos\langle x,y\rangle+\arccos\langle y,z\rangle$.
(ii) For nonzero $x,y\in\R^d$: $\angle(x,y)=\min_{s\in\{\pm1\}}\arccos\langle\hat x,s\hat y\rangle$.
(iii) For nonzero $x,y,z\in\R^d$: $\angle(x,y)=\angle(y,x)$ and
\begin{equation}\label{eq:triangle}
\angle(x,z)\le\angle(x,y)+\angle(y,z).
\end{equation}
\end{lemma}
\begin{proof}
Tools used: Tools~\ref{tool:cs}, \ref{tool:trig}; Lemma~\ref{lem:T41}.
\begin{steps}
\item \emph{(i), angles.} By Tool~\ref{tool:cs}, $\langle x,y\rangle,\langle y,z\rangle,\langle x,z\rangle\in[-1,1]$. Put $\vartheta_1:=\arccos\langle x,y\rangle$ and $\vartheta_2:=\arccos\langle y,z\rangle$, both in $[0,\pi]$, so $\cos\vartheta_1=\langle x,y\rangle$ and $\cos\vartheta_2=\langle y,z\rangle$ (Tool~\ref{tool:trig}).
\item \emph{(i), components orthogonal to $y$.} Let $x^\perp:=x-\langle x,y\rangle y$ and $z^\perp:=z-\langle z,y\rangle y$. Since $\|y\|=1$, $\langle x^\perp,y\rangle=0$ and $\langle z^\perp,y\rangle=0$, and $\|x^\perp\|^2=\|x\|^2-\langle x,y\rangle^2=1-\cos^2\vartheta_1=\sin^2\vartheta_1$; as $\sin\vartheta_1\ge0$ on $[0,\pi]$ (Tool~\ref{tool:trig}), $\|x^\perp\|=\sin\vartheta_1$. Likewise $\|z^\perp\|=\sin\vartheta_2$.
\item \emph{(i), lower bound on $\langle x,z\rangle$.} $x=x^\perp+\cos\vartheta_1\,y$ and $z=z^\perp+\cos\vartheta_2\,y$. Expanding, the cross terms $\cos\vartheta_2\langle x^\perp,y\rangle$ and $\cos\vartheta_1\langle y,z^\perp\rangle$ vanish by Step~2, and $\langle y,y\rangle=1$; so $\langle x,z\rangle=\langle x^\perp,z^\perp\rangle+\cos\vartheta_1\cos\vartheta_2$. By Tool~\ref{tool:cs}, $\langle x^\perp,z^\perp\rangle\ge-\|x^\perp\|\,\|z^\perp\|=-\sin\vartheta_1\sin\vartheta_2$. Hence, by the addition formula (Tool~\ref{tool:trig}), $\langle x,z\rangle\ge\cos\vartheta_1\cos\vartheta_2-\sin\vartheta_1\sin\vartheta_2=\cos(\vartheta_1+\vartheta_2)$.
\item \emph{(i), conclusion.} If $\vartheta_1+\vartheta_2\le\pi$: $\arccos$ is decreasing on $[-1,1]$ and $\arccos(\cos(\vartheta_1+\vartheta_2))=\vartheta_1+\vartheta_2$ (as $\vartheta_1+\vartheta_2\in[0,\pi]$; Tool~\ref{tool:trig}), so Step~3 gives $\arccos\langle x,z\rangle\le\vartheta_1+\vartheta_2$. If $\vartheta_1+\vartheta_2>\pi$: $\arccos\langle x,z\rangle\le\pi<\vartheta_1+\vartheta_2$.
\item \emph{(ii).} Put $u:=\langle\hat x,\hat y\rangle\in[-1,1]$, so $|\cos(x,y)|=|u|$ and $\angle(x,y)=\arccos|u|$. By Tool~\ref{tool:trig}, $\arccos(-u)=\pi-\arccos u$, $\arccos0=\pi/2$ and $\arccos$ is decreasing. If $u\ge0$: $\arccos u\le\arccos0=\pi/2\le\pi-\arccos u$, so $\min_s\arccos(su)=\arccos u=\arccos|u|$. If $u<0$: $\arccos(-u)=\arccos|u|\le\pi/2<\arccos u$, so $\min_s\arccos(su)=\arccos|u|$. In both cases $\min_s\arccos\langle\hat x,s\hat y\rangle=\angle(x,y)$.
\item \emph{(iii), symmetry.} Lemma~\ref{lem:T41}(i).
\item \emph{(iii), triangle inequality.} By (ii), choose $s_1\in\{\pm1\}$ with $\arccos\langle\hat x,s_1\hat y\rangle=\angle(x,y)$. Since $\angle(s_1y,z)=\angle(y,z)$ (Lemma~\ref{lem:T41}(i)) and $\widehat{s_1y}=s_1\hat y$, (ii) applied to $(s_1y,z)$ gives $s_2\in\{\pm1\}$ with $\arccos\langle s_1\hat y,s_2\hat z\rangle=\angle(y,z)$. By (i) for the unit vectors $\hat x$, $s_1\hat y$, $s_2\hat z$: $\arccos\langle\hat x,s_2\hat z\rangle\le\angle(x,y)+\angle(y,z)$. By (ii), $\angle(x,z)\le\arccos\langle\hat x,s_2\hat z\rangle$. This is \eqref{eq:triangle}.
\end{steps}
\end{proof}

\begin{lemma}[chord--angle identity]\label{lem:T42}\MkOursChk
For nonzero $x,y\in\R^d$,
\begin{equation}\label{eq:chordangle}
\min_{s\in\{\pm1\}}\|\hat x-s\hat y\|=\ch\big(\angle(x,y)\big).
\end{equation}
$\ch$ is strictly increasing on $[0,\pi/2]$ with $\ch(0)=0$. Consequently, for every $m$, $i$ and $\kappa\in(0,1]$:
\begin{equation}\label{eq:chordangle2}
e_m(A\to A')=\ch\big(e^\angle_m(A\to A')\big),\qquad
\delta_i(A\to A';\kappa)=\ch\big(\delta^\angle_i(A\to A';\kappa)\big),
\end{equation}
and the same with $A,A'$ exchanged.
\end{lemma}
\begin{proof}
\begin{steps}
\item $\|\hat x\|=\|\hat y\|=1$ and $s^2=1$, so $\|\hat x-s\hat y\|^2=2-2s\langle\hat x,\hat y\rangle$.
\item Over $s\in\{\pm1\}$ the right side of Step~1 is smallest when $s\langle\hat x,\hat y\rangle=|\langle\hat x,\hat y\rangle|$, with value $2-2|\cos(x,y)|=2-2\cos\angle(x,y)$ (Lemma~\ref{lem:T41}(i)).
\item By Tool~\ref{tool:trig}, $2-2\cos\angle(x,y)=4\sin^2(\angle(x,y)/2)$.
\item $\sqrt{\cdot}$ is increasing on $[0,\infty)$ (Tool~\ref{tool:mono}), so the minimum of the norms is the square root of the minimum of the squares (Tool~\ref{tool:mono}): $\min_s\|\hat x-s\hat y\|=2|\sin(\angle(x,y)/2)|=2\sin(\angle(x,y)/2)$, since $\angle(x,y)/2\in[0,\pi/4]$ and $\sin\ge0$ there (Tool~\ref{tool:trig}). This is \eqref{eq:chordangle}.
\item $\ch$ is strictly increasing on $[0,\pi/2]$ because $\vartheta/2\in[0,\pi/4]$ and $\sin$ is strictly increasing there (Tool~\ref{tool:trig}); $\ch(0)=0$.
\item By D7 and Step~4, $e_m(A\to A')=\min_{m'}\ch(\angle(a_m,a'_{m'}))$; by Step~5 and Tool~\ref{tool:mono} this is $\ch(\min_{m'}\angle(a_m,a'_{m'}))=\ch(e^\angle_m(A\to A'))$.
\item Both $\delta_i(A\to A';\kappa)$ and $\delta^\angle_i(A\to A';\kappa)$ are maxima over the same set $P_i(A;\kappa)$ (Definition~\ref{def:T4not}). If it is empty, both local distances are $0=\ch(0)$. Otherwise, by Step~6 and Tool~\ref{tool:mono}, $\max_{m\in P_i(A;\kappa)}e_m=\ch(\max_{m\in P_i(A;\kappa)}e^\angle_m)$. This is \eqref{eq:chordangle2}; the definitions are symmetric in $(A,M)$ and $(A',M')$.
\end{steps}
\end{proof}

\begin{lemma}[one-atom transfer]\label{lem:T43}\MkOursChk
(i) For nonzero $x,y,z\in\R^d$, $\big||\cos(x,z)|-|\cos(y,z)|\big|\le\min_{s\in\{\pm1\}}\|\hat x-s\hat y\|$.
(ii) For every $m\in[M]$ and $i\in[n]$,
\begin{equation}\label{eq:transfer}
|\cos(a_m,v_i)|\le c_i(A')+e_m(A\to A'),
\end{equation}
and, with the roles exchanged, $|\cos(a'_{m'},v_i)|\le c_i(A)+e_{m'}(A'\to A)$ for every $m'\in[M']$.
\end{lemma}
\begin{proof}
\begin{steps}
\item Fix $s\in\{\pm1\}$. By D0, $|\cos(x,z)|=|\langle\hat x,\hat z\rangle|$ and $|\cos(y,z)|=|\langle\hat y,\hat z\rangle|=|\langle s\hat y,\hat z\rangle|$.
\item By Tool~\ref{tool:op} (real numbers), $\big||\langle\hat x,\hat z\rangle|-|\langle s\hat y,\hat z\rangle|\big|\le|\langle\hat x-s\hat y,\hat z\rangle|$.
\item By Tool~\ref{tool:cs} and $\|\hat z\|=1$, the right side of Step~2 is at most $\|\hat x-s\hat y\|$.
\item Steps~1--3 hold for both values of $s$; taking the smaller bound gives (i).
\item For (ii), choose $m'$ and $s$ attaining the minimum in the definition of $e_m(A\to A')$ (D7; finite sets). By (i), $|\cos(a_m,v_i)|\le|\cos(a'_{m'},v_i)|+\|\hat a_m-s\hat a'_{m'}\|$.
\item $|\cos(a'_{m'},v_i)|\le c_i(A')$ (definition of $c_i$, D7), and $\|\hat a_m-s\hat a'_{m'}\|=e_m(A\to A')$ by the choice in Step~5. This gives \eqref{eq:transfer}. The definitions are symmetric in $A$ and $A'$, which gives the exchanged form.
\end{steps}
\end{proof}

\begin{lemma}[cosine-difference bound]\label{lem:T44}\MkOursChk
For $\vartheta_1,\vartheta\in[0,\pi/2]$: $\vartheta_1+\vartheta\in[0,\pi]$ and
\begin{equation}\label{eq:cosdiff}
\cos\vartheta_1-\cos(\vartheta_1+\vartheta)\le\ch(\vartheta).
\end{equation}
\end{lemma}
\begin{proof}
\begin{steps}
\item $0\le\vartheta_1+\vartheta\le\pi/2+\pi/2=\pi$.
\item By Tool~\ref{tool:trig} with $\vartheta_1$ and $\vartheta_2=\vartheta_1+\vartheta$: $\cos\vartheta_1-\cos(\vartheta_1+\vartheta)=2\sin(\vartheta_1+\vartheta/2)\sin(\vartheta/2)$.
\item $\sin(\vartheta/2)\ge0$ since $\vartheta/2\in[0,\pi/4]$, and $\sin(\vartheta_1+\vartheta/2)\le1$ (Tool~\ref{tool:trig}). So the product in Step~2 is at most $2\sin(\vartheta/2)=\ch(\vartheta)$.
\end{steps}
\end{proof}

\subsection{C8: invariances and Lipschitz bounds for the feature scores}\label{ssec:C8}

A feature score $c_i(A)$ is a maximum of absolute cosines, so it does not depend on the order, the signs or the scales of the atoms, and it moves by at most the line-Hausdorff distance $H(A,A')$ when the dictionary changes. The smoothed rate $\FRR_{t,\beta}$ therefore changes by at most $H(A,A')/\beta$. Theorem~\ref{thm:C8} gives no comparable bound for the recovery rate $\FRR_t$, which can jump when a score crosses $t$; \eqref{eq:C8c2} bounds the gap between the two rates. The area $\bar c$ and the smoothed rate are the quantities that Theorem~\ref{thm:C8} shows to be Lipschitz in $H$.

\begin{theorem}[C8]\label{thm:C8}\MkOursChk
Let $V$ be as in D7, $A\in\cA_M$, $A'\in\cA_{M'}$, $t\in(0,1]$ and $\beta\in(0,1)$.
\begin{enumerate}[label=(\alph*)]
\item $c_i$ ($i\in[n]$), $\bar c$, $\FRR_t$ and $\FRR_{t,\beta}$ are unchanged when the atoms of $A$ are permuted, multiplied by signs, or multiplied by positive scalars (more generally, by any nonzero scalars).
\item For every $i$, $|c_i(A)-c_i(A')|\le H(A,A')$, and $|\bar c(A)-\bar c(A')|\le H(A,A')$.
\item
\begin{align}
|\FRR_{t,\beta}(A)-\FRR_{t,\beta}(A')|&\le H(A,A')/\beta,\label{eq:C8c1}\\
|\FRR_{t,\beta}(A)-\FRR_t(A)|&\le\tfrac1n\#\{i:|c_i(A)-t|<\beta/2\}.\label{eq:C8c2}
\end{align}
\end{enumerate}
Credit: on normalised atoms, the one-sided distance $\max_me_m(A\to A')$ is the one-sided dictionary distance of Pali and Schnass \cite{pali2023} (Eq.~(3) of the arXiv text; full text not seen); we did not find the symmetrised maximum $H$ there. We claim no novelty.
\end{theorem}

\begin{proof}[Proof of (a)]
\begin{steps}
\item Let $\varsigma$ be a permutation of $[M]$ and $\mathsf k_1,\dots,\mathsf k_M$ nonzero reals (signs and positive scalars are special cases), and let $\tilde A$ have atoms $\tilde a_m=\mathsf k_ma_{\varsigma(m)}$. Each $\tilde a_m\ne0$, so $\tilde A\in\cA_M$.
\item By Lemma~\ref{lem:T41}(i) (with $\mathsf k=\mathsf k_m$, $\mathsf k'=1$), $|\cos(\tilde a_m,v_i)|=|\cos(a_{\varsigma(m)},v_i)|$.
\item Hence $c_i(\tilde A)=\max_m|\cos(a_{\varsigma(m)},v_i)|=\max_m|\cos(a_m,v_i)|=c_i(A)$, since $\varsigma$ is a bijection of $[M]$ and the maximum is over the same set of numbers.
\item By D7, $\bar c$, $\FRR_t$ and $\FRR_{t,\beta}$ are functions of the vector $(c_1,\dots,c_n)$ alone, so by Step~3 they are unchanged. Compositions of such operations are again of the form in Step~1.
\end{steps}
\end{proof}

\begin{proof}[Proof of (b)]
\begin{steps}
\item Let $m$ be a maximiser of $c_i(A)$. By \eqref{eq:transfer}, $c_i(A)=|\cos(a_m,v_i)|\le c_i(A')+e_m(A\to A')$.
\item $e_m(A\to A')\le\max_me_m(A\to A')\le H(A,A')$ (D7). So $c_i(A)-c_i(A')\le H(A,A')$.
\item Exchanging $A$ and $A'$ (the exchanged form of \eqref{eq:transfer} and the symmetry of $H$ in D7) gives $c_i(A')-c_i(A)\le H(A,A')$. Steps~2--3 give the first claim.
\item $|\bar c(A)-\bar c(A')|=\big|\tfrac1n\sum_i(c_i(A)-c_i(A'))\big|\le\tfrac1n\sum_i|c_i(A)-c_i(A')|\le H(A,A')$, by the triangle inequality and Step~3.
\end{steps}
\end{proof}

\begin{proof}[Proof of (c)]
\begin{steps}
\item Closed form: for all real $u$, $\psi_{t,\beta}(u)=\min\big(1,\max\big(0,(u-t+\beta/2)/\beta\big)\big)$. Check against D7: for $u\le t-\beta/2$ the inner value is $\le0$, so the result is $0$; for $u\ge t+\beta/2$ it is $\ge1$, so the result is $1$; in between it lies in $(0,1)$ and the result is the affine function $(u-t+\beta/2)/\beta$, which takes the values $0$ and $1$ at the two ends, so it is the affine interpolation of D7.
\item By Step~1 and Tool~\ref{tool:lip}, $\psi_{t,\beta}$ is $(1/\beta)$-Lipschitz: it is the composition of $u\mapsto\min(1,u)$, $u\mapsto\max(0,u)$ (both $1$-Lipschitz) and $u\mapsto(u-t+\beta/2)/\beta$ (Lipschitz with constant $1/\beta$).
\item By D7, Step~2, the triangle inequality and (b): $|\FRR_{t,\beta}(A)-\FRR_{t,\beta}(A')|\le\tfrac1n\sum_i|\psi_{t,\beta}(c_i(A))-\psi_{t,\beta}(c_i(A'))|\le\tfrac1n\sum_i|c_i(A)-c_i(A')|/\beta\le H(A,A')/\beta$. This is \eqref{eq:C8c1}.
\item Pointwise comparison: for every real $u$, $|\psi_{t,\beta}(u)-\one[u\ge t]|\le\one[|u-t|<\beta/2]$. Cases: if $u\ge t+\beta/2$, both terms are $1$ (Step~1 and $u\ge t$); if $u\le t-\beta/2$, both are $0$ (Step~1 and $u<t$, since $\beta>0$); otherwise $|u-t|<\beta/2$ and the difference of two numbers in $[0,1]$ is at most $1$.
\item By D7, $\FRR_{t,\beta}(A)-\FRR_t(A)=\tfrac1n\sum_i\big(\psi_{t,\beta}(c_i(A))-\one[c_i(A)\ge t]\big)$. The triangle inequality and Step~4 give \eqref{eq:C8c2}.
\end{steps}
\end{proof}

\subsection{C9(a): uniqueness of the aligned feature}\label{ssec:C9a}

Two features are at least $\theta_V$ apart as lines. A line at angle less than $\theta_V/2$ from one feature is therefore at angle more than $\theta_V/2$ from every other, by the triangle inequality for the line angle. So above the threshold $\cos(\theta_V/2)$ an atom is aligned with at most one feature: distinct aligned features use distinct atoms, and the number of features aligned above the threshold is at most the number of atoms aligned above it. The threshold cannot be lowered (Remark~\ref{rem:T45}).

\begin{theorem}[C9(a)]\label{thm:C9a}\MkOursChk
Let $V$ be as in D7 and $t\in(\cos(\theta_V/2),1]$. For every nonzero $x\in\R^d$, the inequality $|\cos(x,v_i)|\ge t$ holds for at most one $i\in[n]$. In particular, for every $A\in\cA_M$ and every $m$, $|\cos(a_m,v_i)|\ge t$ holds for at most one $i$.

Credit: Chen et al.\ \cite{chen2026} (Appendix~C.4, p.~28, text after Eq.~(C.5) and Figure~11) state without proof that every neuron whose cosine with a feature exceeds $\talign$ is uniquely aligned with a single feature, by a non-overlapping-cones argument with half-angle $\frac13\arccos\mu(V)$. Our statement is the line version with the threshold $\cos(\theta_V/2)$. We claim no novelty for the mechanism.
\end{theorem}
\begin{proof}
Here $\theta_V=\min_{i\ne j}\angle(v_i,v_j)$ (D7 and \eqref{eq:lineangle}). Tools used: Tool~\ref{tool:trig}; Lemmas~\ref{lem:T41}, \ref{lem:triangle}.
\begin{steps}
\item The interval is nonempty: $\theta_V/2\in(0,\pi/4]$, so, as $\cos$ is strictly decreasing on $[0,\pi]$ with $\cos0=1$ (Tool~\ref{tool:trig}), $\cos(\theta_V/2)<1$.
\item Suppose, for a contradiction, that $i\ne j$ and $|\cos(x,v_i)|\ge t$, $|\cos(x,v_j)|\ge t$.
\item By Lemma~\ref{lem:T41}(i),(ii) with $u=t$: $\angle(x,v_i)\le\arccos t$. Since $t>\cos(\theta_V/2)$ and $\arccos$ is strictly decreasing (Tool~\ref{tool:trig}), $\arccos t<\arccos\cos(\theta_V/2)=\theta_V/2$ (as $\theta_V/2\in[0,\pi]$). So $\angle(x,v_i)<\theta_V/2$, and likewise $\angle(x,v_j)<\theta_V/2$.
\item By Lemma~\ref{lem:triangle}(iii), $\angle(v_i,v_j)\le\angle(v_i,x)+\angle(x,v_j)<\theta_V$.
\item By the definition of $\theta_V$ (D7), $\angle(v_i,v_j)\ge\theta_V$ for $i\ne j$. This contradicts Step~4.
\end{steps}
\end{proof}

\begin{remark}[the threshold is sharp]\label{rem:T45}\MkOursChk
For every $V$ as in D7 there are $i\ne j$ and a nonzero $x\in\R^d$ with $|\cos(x,v_i)|=|\cos(x,v_j)|=\cos(\theta_V/2)$. So at $t=\cos(\theta_V/2)$ the conclusion of Theorem~\ref{thm:C9a} fails for every $V$.
\end{remark}
\begin{proof}
Since $n\ge2$, the minimum defining $\theta_V$ is over a finite nonempty set; let $i\ne j$ attain it, so $|\mu_{ij}|=\cos\theta_V<1$ (Tool~\ref{tool:trig}: $\cos(\arccos u)=u$). Let $s:=1$ if $\mu_{ij}\ge0$ and $s:=-1$ otherwise, and $x:=\hat v_i+s\hat v_j$.
\begin{steps}
\item $s\mu_{ij}=|\mu_{ij}|$, so $\|x\|^2=2+2s\mu_{ij}=2(1+|\mu_{ij}|)>0$ (note $\langle\hat v_i,\hat v_j\rangle=\mu_{ij}$).
\item $\langle x,\hat v_i\rangle=1+s\mu_{ij}=1+|\mu_{ij}|$ and $\langle x,\hat v_j\rangle=\mu_{ij}+s=s(1+|\mu_{ij}|)$ (using $s^2=1$).
\item Hence $|\cos(x,v_i)|=|\cos(x,v_j)|=(1+|\mu_{ij}|)/\sqrt{2(1+|\mu_{ij}|)}=\sqrt{(1+\cos\theta_V)/2}=\cos(\theta_V/2)$, using $|\mu_{ij}|=\cos\theta_V$ (definition of $\theta_V$ and the choice of $i,j$), then Tool~\ref{tool:trig} and $\cos(\theta_V/2)\ge0$. Here $x\ne0$ by Step~1.
\end{steps}
\end{proof}

\begin{remark}[a gap in the cited uniqueness sentence]\label{rem:chen}\MkOurs
In Chen et al.\ \cite{chen2026} (p.~28) the FRR test uses the absolute cosine, while $\talign$ uses the signed coherence $\mu(V)=\max_{i\ne j}\cos(v_i,v_j)$ (Eq.~(C.5)). We read ``a neuron whose cosine with a feature exceeds $\talign$'' with the absolute cosine, as in the FRR test; the sentence is cited by paraphrase (Theorem~\ref{thm:C9a}). (a) With these two conventions the uniqueness sentence can fail: for $d=n=2$, $\vartheta_0\in(0,\pi/4)$, $v_1=(1,0)^\top$, $v_2=(-\cos\vartheta_0,\sin\vartheta_0)^\top$ and $x=(1,0)^\top$, we have $|\cos(x,v_1)|>\talign(V)$ and $|\cos(x,v_2)|>\talign(V)$, with strict inequalities. Under the signed reading the same $x$ has $\cos(x,v_2)<0<\talign(V)$, so the example concerns the absolute-cosine reading only. (b) If $\talign(V)>\cos(\theta_V/2)$, the sentence holds for every $V$ as in D7 (with either ``$>$'' or ``$\ge$''). If the threshold is formed from the absolute coherence $\max_{i\ne j}|\cos(v_i,v_j)|$ instead of $\mu(V)$, it equals $\cos(\theta_V/3)>\cos(\theta_V/2)$, so the sentence holds.
\end{remark}
\begin{proof}
\begin{steps}
\item \emph{(a), the threshold.} $\cos(v_1,v_2)=\langle v_1,v_2\rangle=-\cos\vartheta_0\in(-1,0)$ (both vectors are unit vectors), so $|\cos(v_1,v_2)|<1$ (D7 holds), $\mu(V)=-\cos\vartheta_0$, $\arccos\mu(V)=\pi-\vartheta_0$ (Tool~\ref{tool:trig}, $\arccos(-u)=\pi-\arccos u$) and $\talign(V)=\cos\big((\pi-\vartheta_0)/3\big)$.
\item \emph{(a), feature $1$.} $|\cos(x,v_1)|=1$. Since $(\pi-\vartheta_0)/3\in(\pi/4,\pi/3)\subset(0,\pi]$ and $\cos$ is strictly decreasing on $[0,\pi]$ with $\cos0=1$ (Tool~\ref{tool:trig}), $\talign(V)<1=|\cos(x,v_1)|$.
\item \emph{(a), feature $2$.} $|\cos(x,v_2)|=\cos\vartheta_0$. Now $\vartheta_0<\pi/4\iff3\vartheta_0<\pi-\vartheta_0\iff\vartheta_0<(\pi-\vartheta_0)/3$. Both angles lie in $[0,\pi]$, so by Tool~\ref{tool:trig} $\cos\vartheta_0>\cos((\pi-\vartheta_0)/3)=\talign(V)$.
\item \emph{(a), consistency with Theorem~\ref{thm:C9a}.} Here $n=2$, so $\theta_V=\arccos|\cos(v_1,v_2)|=\arccos(\cos\vartheta_0)=\vartheta_0$ (Tool~\ref{tool:trig}). Since $\vartheta_0<\pi/4$, $(\pi-\vartheta_0)/3>\pi/4>\pi/8>\vartheta_0/2$, all in $[0,\pi]$, so $\talign(V)<\cos(\vartheta_0/2)=\cos(\theta_V/2)$ (Tool~\ref{tool:trig}); Theorem~\ref{thm:C9a} does not apply at $t=\talign(V)$, consistently with Steps~2--3.
\item \emph{(a), signed reading.} $\cos(x,v_2)=-\cos\vartheta_0<0$, while $\talign(V)>0$ because $(\pi-\vartheta_0)/3<\pi/2$ and $\cos>0$ on $[0,\pi/2)$ (Tool~\ref{tool:trig}: strictly decreasing with $\cos(\pi/2)=0$).
\item \emph{(b), first claim.} Assume $\talign(V)>\cos(\theta_V/2)$; also $\talign(V)\le1$, being a cosine. Let $x\ne0$. If $|\cos(x,v_i)|>\talign(V)$ (or $\ge\talign(V)$) held for two indices $i$, then $|\cos(x,v_i)|\ge\talign(V)$ would hold for both, which Theorem~\ref{thm:C9a} with $t=\talign(V)\in(\cos(\theta_V/2),1]$ excludes.
\item \emph{(b), absolute coherence.} $\arccos$ is decreasing (Tool~\ref{tool:trig}), so by Tool~\ref{tool:mono}
\[
\arccos\max_{i\ne j}|\cos(v_i,v_j)|=\min_{i\ne j}\arccos|\cos(v_i,v_j)|=\theta_V
\]
(D7). The threshold is therefore $\cos(\theta_V/3)$. As $0<\theta_V/3<\theta_V/2\le\pi/4$ and $\cos$ is strictly decreasing on $[0,\pi]$, $\cos(\theta_V/3)>\cos(\theta_V/2)$; also $\cos(\theta_V/3)\le1$. Step~6, with this threshold in place of $\talign(V)$, gives the claim.
\end{steps}
\end{proof}

\subsection{C11: a global certificate for the feature recovery rate}\label{ssec:C11}

The global certificate counts the features whose score is within $\gamma$ of the threshold, plus the aligned atoms that moved by more than $\gamma$. Everything except the distances to the other dictionary is computed from $A$ alone.

\begin{definition}[counts for C11]\label{def:C11counts}\MkDef
For $t\in(0,1]$ and $\gamma>0$:
\begin{align}
X_\gamma&:=\#\{m\in[M]:g_m(A)\ge t+\gamma,\ e_m(A\to A')>\gamma\},\label{eq:X}\\
Y_\gamma&:=\#\{m'\in[M']:g_{m'}(A')\ge t,\ e_{m'}(A'\to A)\ge\gamma\},\label{eq:Y}
\end{align}
$N_\gamma(A;t)$ as in D7, and its two halves $N^+_\gamma:=\#\{i:t\le c_i(A)<t+\gamma\}$ and $N^-_\gamma:=\#\{i:t-\gamma<c_i(A)<t\}$. Lost and gained features: $\Lost:=\Rec_t(A)\setminus\Rec_t(A')$ and $\Gain:=\Rec_t(A')\setminus\Rec_t(A)$.
\end{definition}

\begin{theorem}[C11]\label{thm:C11}\MkOursChk
Let $V$ be as in D7, $t\in(\cos(\theta_V/2),1]$, $A\in\cA_M$, $A'\in\cA_{M'}$. For every $\gamma>0$,
\begin{equation}\label{eq:C11main}
n\,|\FRR_t(A)-\FRR_t(A')|\le N_\gamma(A;t)+\max(X_\gamma,Y_\gamma),
\end{equation}
\begin{equation}\label{eq:C11markov}
X_\gamma\le\gamma^{-2}\sum_{m:\,g_m(A)\ge t+\gamma}e_m(A\to A')^2,\qquad
Y_\gamma\le\gamma^{-2}\sum_{m':\,g_{m'}(A')\ge t}e_{m'}(A'\to A)^2.
\end{equation}
Moreover, for every $t\in(0,1]$ and every $i$:
\begin{itemize}
\item[(C11-R)] if $m_i$ is a maximiser of $c_i(A)$ and $c_i(A)-t\ge e_{m_i}(A\to A')$, then $c_i(A')\ge t$;
\item[(C11-NR)] if $t-c_i(A)>H(A,A')$, then $c_i(A')<t$.
\end{itemize}
The quantities $N_\gamma(A;t)$, $g_m(A)$, $c_i(A)$ and $m_i$ depend on $A$ alone; $e_m(A\to A')$, $e_{m'}(A'\to A)$ and $H(A,A')$ depend on both; $Y_\gamma$ and its bound also need the scores $g_{m'}(A')$.

Credit: the mechanism (a margin larger than a Lipschitz bound on the perturbation preserves a decision) is that of Tsuzuku, Sato and Sugiyama \cite{tsuzuku2018}, Proposition~1; the one-sided distances are those of Pali and Schnass \cite{pali2023} (Eq.~(3) of the arXiv text). We claim no novelty for the mechanism.
\end{theorem}

\begin{proof}[Proof of \eqref{eq:C11main}]
Tools used: Lemma~\ref{lem:T43}; Theorems~\ref{thm:C8}(b) and \ref{thm:C9a}; D7.
\begin{steps}
\item Counting: $n\,\FRR_t(A)-n\,\FRR_t(A')=\#\Rec_t(A)-\#\Rec_t(A')=\#\Lost-\#\Gain$, since both recovered sets contain $\Rec_t(A)\cap\Rec_t(A')$. As both counts are $\ge0$, $n|\FRR_t(A)-\FRR_t(A')|\le\max(\#\Lost,\#\Gain)$.
\item Split $\Lost$ by the value of $c_i(A)$ (which is $\ge t$ on $\Lost$): $\Lost_1:=\{i\in\Lost:c_i(A)<t+\gamma\}$ and $\Lost_2:=\{i\in\Lost:c_i(A)\ge t+\gamma\}$. Every $i\in\Lost_1$ has $0\le c_i(A)-t<\gamma$, so $\#\Lost_1\le N^+_\gamma$.
\item Let $i\in\Lost_2$ and let $m_i$ be a maximiser of $c_i(A)$. Then $g_{m_i}(A)\ge|\cos(a_{m_i},v_i)|=c_i(A)\ge t+\gamma$ (definition of $g_m$, D7).
\item For the same $i$: by \eqref{eq:transfer}, $e_{m_i}(A\to A')\ge c_i(A)-c_i(A')$. Here $c_i(A)\ge t+\gamma$ and $c_i(A')<t$ (as $i\notin\Rec_t(A')$), so $c_i(A)-c_i(A')>(t+\gamma)-t=\gamma$: the inequality is strict because $c_i(A')<t$ is strict. Hence $e_{m_i}(A\to A')>\gamma$, which is the strict condition in \eqref{eq:X}. With Step~3, $m_i$ belongs to the set counted by $X_\gamma$.
\item The map $i\mapsto m_i$ on $\Lost_2$ is injective: if $m_i=m_j=m$ with $i\ne j$, then $|\cos(a_m,v_i)|\ge t$ and $|\cos(a_m,v_j)|\ge t$, contradicting Theorem~\ref{thm:C9a} (here $t>\cos(\theta_V/2)$ is used). So $\#\Lost_2\le X_\gamma$, and $\#\Lost\le N^+_\gamma+X_\gamma$.
\item Split $\Gain$ (where $c_i(A)<t\le c_i(A')$): $\Gain_1:=\{i\in\Gain:c_i(A)>t-\gamma\}$ and $\Gain_2:=\{i\in\Gain:c_i(A)\le t-\gamma\}$. Every $i\in\Gain_1$ has $0<t-c_i(A)<\gamma$, so $\#\Gain_1\le N^-_\gamma$.
\item Let $i\in\Gain_2$ and let $m'_i$ be a maximiser of $c_i(A')$. Then $g_{m'_i}(A')\ge c_i(A')\ge t$, which is the non-strict condition in \eqref{eq:Y}.
\item For the same $i$: by the exchanged form of \eqref{eq:transfer}, $e_{m'_i}(A'\to A)\ge c_i(A')-c_i(A)\ge t-(t-\gamma)=\gamma$. Both inputs are non-strict ($c_i(A')\ge t$, $c_i(A)\le t-\gamma$), so only $\ge\gamma$ follows, which is exactly the non-strict condition in \eqref{eq:Y}. So $m'_i$ is counted by $Y_\gamma$.
\item As in Step~5, applied to $A'$ (Theorem~\ref{thm:C9a} holds for every nonzero vector), $i\mapsto m'_i$ is injective on $\Gain_2$. So $\#\Gain_2\le Y_\gamma$ and $\#\Gain\le N^-_\gamma+Y_\gamma$.
\item The two halves are disjoint ($c_i(A)\ge t$ in one, $<t$ in the other) and their union is $\{i:t-\gamma<c_i(A)<t+\gamma\}$, so $N^+_\gamma+N^-_\gamma=N_\gamma(A;t)$.
\item By Steps~1, 5, 9 and 10:
\begin{equation}\label{eq:C11sharp}
n|\FRR_t(A)-\FRR_t(A')|\le\max\big(N^+_\gamma+X_\gamma,\ N^-_\gamma+Y_\gamma\big)\le N_\gamma(A;t)+\max(X_\gamma,Y_\gamma).
\end{equation}
\end{steps}
\end{proof}

\begin{proof}[Proof of \eqref{eq:C11markov}]
\begin{steps}
\item Each $m$ counted by $X_\gamma$ has $e_m(A\to A')>\gamma>0$, hence $e_m(A\to A')^2/\gamma^2>1$.
\item So $X_\gamma=\sum_{m\text{ counted}}1\le\sum_{m\text{ counted}}e_m(A\to A')^2/\gamma^2\le\sum_{m:\,g_m(A)\ge t+\gamma}e_m(A\to A')^2/\gamma^2$: the first inequality by Step~1 (an empty sum is $0$), the second because the counted $m$ are among those with $g_m(A)\ge t+\gamma$ and the added terms are $\ge0$.
\item Each $m'$ counted by $Y_\gamma$ has $e_{m'}(A'\to A)\ge\gamma$, hence $e_{m'}(A'\to A)^2/\gamma^2\ge1$; the same argument gives the second bound.
\end{steps}
\end{proof}

\begin{proof}[{{Proof of the two per-feature certificates ($t\in(0,1]$)}}]
\begin{steps}
\item (C11-R): by \eqref{eq:transfer} with $m=m_i$, $c_i(A')\ge|\cos(a_{m_i},v_i)|-e_{m_i}(A\to A')=c_i(A)-e_{m_i}(A\to A')\ge t$.
\item (C11-NR): by Theorem~\ref{thm:C8}(b), $c_i(A')\le c_i(A)+H(A,A')<t$.
\item Neither step uses Theorem~\ref{thm:C9a}, so $t>\cos(\theta_V/2)$ is not needed.
\end{steps}
\end{proof}

\begin{remark}[strict and non-strict inequalities; the split form]\label{rem:C11}\MkOursChk
(1) The strict and non-strict inequalities in \eqref{eq:X}, \eqref{eq:Y} and in $N_\gamma$ fit exactly: a feature with $c_i(A)=t+\gamma$ is not counted in $N_\gamma$, and if it is in $\Lost$, it lies in $\Lost_2$ and Step~4 applies (strict $>\gamma$); a feature with $c_i(A)=t-\gamma$ is not counted in $N_\gamma$, and if it is in $\Gain$, it lies in $\Gain_2$ and Step~8 applies (non-strict $\ge\gamma$). (2) The first inequality of \eqref{eq:C11sharp} is a sharper form of \eqref{eq:C11main}, with the margin count split by side. (3) Theorem~\ref{thm:C9a} enters only in Steps~5 and 9 (injectivity). If $t+\gamma>1$, then $X_\gamma=0$ and $\Lost_2=\emptyset$.
\end{remark}
\begin{proof}
The step numbers refer to the proof of \eqref{eq:C11main}.
\begin{steps}
\item (1), first case: $|c_i(A)-t|=\gamma\not<\gamma$, so $i$ is not counted in $N_\gamma$ (D7). If $i\in\Lost$, then $c_i(A)=t+\gamma\ge t+\gamma$ puts $i$ in $\Lost_2$ (Step~2), and Step~4 gives $e_{m_i}(A\to A')>\gamma$.
\item (1), second case: again $|c_i(A)-t|=\gamma$, so $i$ is not counted in $N_\gamma$. If $i\in\Gain$, then $c_i(A)=t-\gamma\le t-\gamma$ puts $i$ in $\Gain_2$ (Step~6), and Step~8 gives $e_{m'_i}(A'\to A)\ge\gamma$.
\item (2): $\max(N^+_\gamma+X_\gamma,N^-_\gamma+Y_\gamma)\le N^+_\gamma+N^-_\gamma+\max(X_\gamma,Y_\gamma)$ since $N^\pm_\gamma\ge0$; this is the second inequality of \eqref{eq:C11sharp}.
\item (3): the only steps that cite Theorem~\ref{thm:C9a} are Steps~5 and 9. If $t+\gamma>1$: every $g_m(A)\le1<t+\gamma$ (D7, Tool~\ref{tool:cs}), so no $m$ is counted in \eqref{eq:X} and $X_\gamma=0$; and every $c_i(A)\le1<t+\gamma$, so $\Lost_2=\emptyset$.
\end{steps}
\end{proof}

\subsection{C11-loc: feature-local certificates}\label{ssec:C11loc}

The global distance $H$ is a maximum over all atoms, so one unaligned atom without a partner makes it large and the certificate (C11-NR) useless. The local certificates look only at the atoms near feature $i$. Throughout, $t\in(0,1]$ and $\kappa\in(0,t]$; a certificate \emph{fires} for $i$ if its hypothesis holds for $i$.

\begin{theorem}[C11-loc]\label{thm:C11loc}\MkOursChk
Let $V$ be as in D7, $A\in\cA_M$, $A'\in\cA_{M'}$.
\begin{enumerate}[label=(\alph*)]
\item \emph{Per-feature certificates} (every $t\in(0,1]$):
\begin{itemize}
\item[(R)] if $\max_m\big(|\cos(a_m,v_i)|-e_m(A\to A')\big)\ge t$, then $c_i(A')\ge t$;
\item[(NR)] if $t-c_i(A)>\delta_i(A'\to A;t)$, then $c_i(A')<t$;
\item[(R$\angle$)] if $\min_m\big(\angle(a_m,v_i)+e^\angle_m(A\to A')\big)\le\arccos t$, then $c_i(A')\ge t$;
\item[(NR$\angle$)] if $\angle_i(A)>\arccos t+\delta^\angle_i(A'\to A;t)$, then $c_i(A')<t$.
\end{itemize}
\item \emph{Inclusions} (per feature, every $t$): (C11-R) fires $\Rightarrow$ (R) fires $\Rightarrow$ (R$\angle$) fires; (C11-NR) fires $\Rightarrow$ (NR) fires $\Rightarrow$ (NR$\angle$) fires.
\item \emph{A priori form.} Let $\Delta\ge0$. If $\delta_i(A,A';\kappa)\le\Delta$, then $c_i(A)\ge t+\Delta\Rightarrow c_i(A')\ge t$ and $c_i(A)<t-\Delta\Rightarrow c_i(A')<t$. If $\delta^\angle_i(A,A';\kappa)\le\Delta$, then $\angle_i(A)\le\arccos t-\Delta\Rightarrow c_i(A')\ge t$ and $\angle_i(A)>\arccos t+\Delta\Rightarrow c_i(A')<t$.
\item \emph{Count chain} (every $t$). With $N^\angle_t(A\to A')$ the number of $i$ for which neither (R$\angle$) nor (NR$\angle$) fires, and $N^{\mathrm{loc}}_t(A\to A')$ the same for (R) and (NR),
\begin{equation}\label{eq:chain}
n|\FRR_t(A)-\FRR_t(A')|\le\#\{i:\one[c_i(A)\ge t]\ne\one[c_i(A')\ge t]\}\le N^\angle_t(A\to A')\le N^{\mathrm{loc}}_t(A\to A'),
\end{equation}
and $N^{\mathrm{loc}}_t(A\to A')$ is at most the number of $i$ for which neither (C11-R), for any maximiser $m_i$ of $c_i(A)$, nor (C11-NR) fires.
\item \emph{Aggregate form.} Let $t\in(\cos(\theta_V/2),1]$. For every $\gamma\ge0$,
\begin{multline}\label{eq:aggr1}
\#\{i:\delta_i(A,A';t)>\gamma\}\le\#\{m:g_m(A)\ge t,\ e_m(A\to A')>\gamma\}\\
+\#\{m':g_{m'}(A')\ge t,\ e_{m'}(A'\to A)>\gamma\},
\end{multline}
and hence
\begin{equation}\label{eq:aggr2}
n|\FRR_t(A)-\FRR_t(A')|\le\#\{i:t-\gamma\le c_i(A)<t+\gamma\}+(\text{the two atom counts in \eqref{eq:aggr1}}).
\end{equation}
\end{enumerate}
(a)--(d) need only $t\in(0,1]$, $\kappa\in(0,t]$ and nonzero atoms; (e) needs $t>\cos(\theta_V/2)$ and $\gamma\ge0$. Credit: the mechanism is that of Tsuzuku, Sato and Sugiyama \cite{tsuzuku2018}, Proposition~1; we claim no novelty for the mechanism.
\end{theorem}

\begin{proof}[Proof of (a)]
Used: Lemmas~\ref{lem:T41}, \ref{lem:triangle} and \ref{lem:T43}.
\begin{steps}
\item (R). Let $m$ attain the maximum (finite set). By \eqref{eq:transfer}, $c_i(A')\ge|\cos(a_m,v_i)|-e_m(A\to A')\ge t$.
\item (NR). Suppose, for a contradiction, that $c_i(A')\ge t$. Let $m'$ be a maximiser of $c_i(A')$; then $|\cos(a'_{m'},v_i)|=c_i(A')\ge t$, so $m'\in P_i(A';t)$, this set is nonempty, and $e_{m'}(A'\to A)\le\delta_i(A'\to A;t)$ (definition of $\delta_i$).
\item By the exchanged form of \eqref{eq:transfer}, $c_i(A)\ge c_i(A')-e_{m'}(A'\to A)\ge t-\delta_i(A'\to A;t)$, i.e.\ $t-c_i(A)\le\delta_i(A'\to A;t)$. This contradicts the hypothesis of (NR). So $c_i(A')<t$.
\item (R$\angle$). Let $m$ attain the minimum, and let $m'$ attain $e^\angle_m(A\to A')=\angle(a_m,a'_{m'})$. By Lemma~\ref{lem:triangle}(iii), $\angle(a'_{m'},v_i)\le\angle(a'_{m'},a_m)+\angle(a_m,v_i)\le\arccos t$.
\item By Lemma~\ref{lem:T41}(i),(ii) with $u=t$ (and $\angle(a'_{m'},v_i)\in[0,\pi/2]$), $|\cos(a'_{m'},v_i)|=\cos\angle(a'_{m'},v_i)\ge t$, so $c_i(A')\ge t$.
\item (NR$\angle$). Suppose, for a contradiction, that $c_i(A')\ge t$. As in Step~2, a maximiser $m'$ of $c_i(A')$ lies in $P_i(A';t)$, so $e^\angle_{m'}(A'\to A)\le\delta^\angle_i(A'\to A;t)$. Also $\angle(a'_{m'},v_i)=\angle_i(A')\le\arccos t$ (Lemma~\ref{lem:T41}(ii),(iii)).
\item Let $m$ attain $e^\angle_{m'}(A'\to A)=\angle(a_m,a'_{m'})$. By Lemma~\ref{lem:T41}(iii) and Lemma~\ref{lem:triangle}(iii), $\angle_i(A)\le\angle(a_m,v_i)\le\angle(a_m,a'_{m'})+\angle(a'_{m'},v_i)\le\delta^\angle_i(A'\to A;t)+\arccos t$. This contradicts the hypothesis of (NR$\angle$).
\end{steps}
\end{proof}

\begin{remark}[empty near sets]\label{rem:T47}\MkOursChk
$P_i(A';t)=\emptyset$ holds if and only if $c_i(A')<t$. In that case $\delta_i(A'\to A;t)=\delta^\angle_i(A'\to A;t)=0$, (NR) fires if and only if $c_i(A)<t$, and (NR$\angle$) fires if and only if $c_i(A)<t$. Such a firing adds nothing to the known fact $c_i(A')<t$; certificates with empty near sets must be counted separately.
\end{remark}
\begin{proof}
\begin{steps}
\item $P_i(A';t)\ne\emptyset$ iff some $m'$ has $|\cos(a'_{m'},v_i)|\ge t$ iff $c_i(A')=\max_{m'}|\cos(a'_{m'},v_i)|\ge t$ (D7, Definition~\ref{def:T4not}). Negating gives the first claim.
\item If $P_i(A';t)=\emptyset$, both local distances are $0$ by Definition~\ref{def:T4not}. Then (NR) reads $t-c_i(A)>0$, i.e.\ $c_i(A)<t$.
\item (NR$\angle$) reads $\angle_i(A)>\arccos t$. By Lemma~\ref{lem:T41}(ii), negated in its non-strict form, with $\vartheta=\angle_i(A)\in[0,\pi/2]$ and $\cos\angle_i(A)=c_i(A)$, this holds iff $c_i(A)<t$.
\end{steps}
\end{proof}

\begin{proof}[Proof of (b)]
\begin{steps}
\item (C11-R) $\Rightarrow$ (R). With $m_i$ as in (C11-R), $\max_m(|\cos(a_m,v_i)|-e_m(A\to A'))\ge|\cos(a_{m_i},v_i)|-e_{m_i}(A\to A')=c_i(A)-e_{m_i}(A\to A')\ge t$.
\item (R) $\Rightarrow$ (R$\angle$). Let $m$ attain the maximum in (R), so $|\cos(a_m,v_i)|-e_m(A\to A')\ge t$. Put $\vartheta_1:=\angle(a_m,v_i)$ and $\vartheta:=e^\angle_m(A\to A')$; both lie in $[0,\pi/2]$. By Lemma~\ref{lem:T41}(i), $\cos\vartheta_1=|\cos(a_m,v_i)|$; by \eqref{eq:chordangle2}, $e_m(A\to A')=\ch(\vartheta)$.
\item By Lemma~\ref{lem:T44}, $\cos(\vartheta_1+\vartheta)\ge\cos\vartheta_1-\ch(\vartheta)\ge t$.
\item Angle step: $\vartheta_1+\vartheta\in[0,\pi]$ (Lemma~\ref{lem:T44}, since $\vartheta_1,\vartheta\le\pi/2$) and $\arccos t\in[0,\pi/2)\subset[0,\pi]$. $\cos$ is strictly decreasing on $[0,\pi]$ (Tool~\ref{tool:trig}) and $t=\cos(\arccos t)$, so Step~3 gives $\vartheta_1+\vartheta\le\arccos t$. Hence $\min_m(\angle(a_m,v_i)+e^\angle_m(A\to A'))\le\vartheta_1+\vartheta\le\arccos t$: (R$\angle$) fires.
\item (C11-NR) $\Rightarrow$ (NR). $\delta_i(A'\to A;t)$ is either $0$ or a value $e_{m'}(A'\to A)$, and $0\le e_{m'}(A'\to A)\le H(A,A')$ for all $m'$ (D7); so $\delta_i(A'\to A;t)\le H(A,A')<t-c_i(A)$.
\item (NR) $\Rightarrow$ (NR$\angle$). Note $\arccos t\in[0,\pi/2)$ and write $\vartheta':=\delta^\angle_i(A'\to A;t)\in[0,\pi/2]$, so $\delta_i(A'\to A;t)=\ch(\vartheta')$ by \eqref{eq:chordangle2}. Suppose, for a contradiction, that $\angle_i(A)\le\arccos t+\vartheta'$.
\item Both $\angle_i(A)\in[0,\pi/2]$ and $\arccos t+\vartheta'\in[0,\pi)$ lie in $[0,\pi]$, where $\cos$ is decreasing (Tool~\ref{tool:trig}); so $c_i(A)=\cos\angle_i(A)\ge\cos(\arccos t+\vartheta')$.
\item By Lemma~\ref{lem:T44} with $\vartheta_1=\arccos t$ and $\vartheta=\vartheta'$, $\cos(\arccos t+\vartheta')\ge\cos(\arccos t)-\ch(\vartheta')=t-\delta_i(A'\to A;t)$. With Step~7, $t-c_i(A)\le\delta_i(A'\to A;t)$, which contradicts the hypothesis of (NR). So $\angle_i(A)>\arccos t+\vartheta'$: (NR$\angle$) fires.
\end{steps}
\end{proof}

\begin{proof}[Proof of (c), chord version]
\begin{steps}
\item Assume $\delta_i(A,A';\kappa)\le\Delta$ and $c_i(A)\ge t+\Delta$. Let $m$ be a maximiser of $c_i(A)$. Then $|\cos(a_m,v_i)|\ge t+\Delta\ge t\ge\kappa$, so $m\in P_i(A;\kappa)$ (here $\kappa\le t$ is used), and $e_m(A\to A')\le\delta_i(A\to A';\kappa)\le\delta_i(A,A';\kappa)\le\Delta$.
\item By \eqref{eq:transfer}, $c_i(A')\ge c_i(A)-e_m(A\to A')\ge t+\Delta-\Delta=t$.
\item Assume $\delta_i(A,A';\kappa)\le\Delta$ and $c_i(A)<t-\Delta$, and suppose, for a contradiction, that $c_i(A')\ge t$. A maximiser $m'$ of $c_i(A')$ has $|\cos(a'_{m'},v_i)|\ge t\ge\kappa$, so $m'\in P_i(A';\kappa)$ and $e_{m'}(A'\to A)\le\delta_i(A'\to A;\kappa)\le\delta_i(A,A';\kappa)\le\Delta$.
\item By the exchanged form of \eqref{eq:transfer}, $c_i(A)\ge c_i(A')-e_{m'}(A'\to A)\ge t-\Delta$, contradicting $c_i(A)<t-\Delta$.
\end{steps}
\end{proof}

\begin{proof}[Proof of (c), angular version]
\begin{steps}
\item Assume $\delta^\angle_i(A,A';\kappa)\le\Delta$ and $\angle_i(A)\le\arccos t-\Delta$. Then $\angle_i(A)\le\arccos t$, so $c_i(A)=\cos\angle_i(A)\ge t\ge\kappa$ (Lemma~\ref{lem:T41}(ii)). A maximiser $m$ of $c_i(A)$ lies in $P_i(A;\kappa)$, has $\angle(a_m,v_i)=\angle_i(A)$ (Lemma~\ref{lem:T41}(iii)) and $e^\angle_m(A\to A')\le\delta^\angle_i(A\to A';\kappa)\le\delta^\angle_i(A,A';\kappa)\le\Delta$.
\item Let $m'$ attain $e^\angle_m(A\to A')$. By Lemma~\ref{lem:triangle}(iii), $\angle(a'_{m'},v_i)\le e^\angle_m(A\to A')+\angle_i(A)\le\Delta+\arccos t-\Delta=\arccos t$, so $c_i(A')\ge|\cos(a'_{m'},v_i)|\ge t$ (Lemma~\ref{lem:T41}(ii)).
\item Assume $\delta^\angle_i(A,A';\kappa)\le\Delta$ and $\angle_i(A)>\arccos t+\Delta$, and suppose $c_i(A')\ge t$. A maximiser $m'$ of $c_i(A')$ lies in $P_i(A';\kappa)$ (as $t\ge\kappa$), has $\angle(a'_{m'},v_i)\le\arccos t$ (Lemma~\ref{lem:T41}(ii),(iii)) and $e^\angle_{m'}(A'\to A)\le\delta^\angle_i(A'\to A;\kappa)\le\delta^\angle_i(A,A';\kappa)\le\Delta$.
\item Let $m$ attain $e^\angle_{m'}(A'\to A)$. By Lemma~\ref{lem:T41}(iii) and Lemma~\ref{lem:triangle}(iii), $\angle_i(A)\le\angle(a_m,v_i)\le\Delta+\arccos t$, a contradiction.
\end{steps}
\end{proof}

\begin{remark}[edge cases of (c)]\label{rem:T48}\MkOursChk
(i) If $t+\Delta>1$ or $t-\Delta\le0$ (respectively $\arccos t-\Delta<0$ or $\arccos t+\Delta\ge\pi/2$), the corresponding implication is vacuous. (ii) $\delta_i(\cdot;\kappa)$ and $\delta^\angle_i(\cdot;\kappa)$ are non-increasing in $\kappa$; so for fixed $t$ the hypothesis of (c) is weakest at $\kappa=t$. (iii) The condition $\kappa\le t$ cannot be dropped: for $\kappa>t$, both the recovery and the non-recovery implications of (c) can fail, in the chord and in the angular version.
\end{remark}
\begin{proof}
\begin{steps}
\item (i): $c_i\in[0,1]$ (D7) and $\angle_i=\arccos c_i\in[0,\pi/2]$. So $c_i(A)\ge t+\Delta>1$, $c_i(A)<t-\Delta\le0$, $\angle_i(A)\le\arccos t-\Delta<0$ and $\angle_i(A)>\arccos t+\Delta\ge\pi/2$ are impossible, and the implication with that hypothesis holds vacuously.
\item (ii): if $\kappa\le\kappa'$, then $P_i(A;\kappa')\subseteq P_i(A;\kappa)$ (Definition~\ref{def:T4not}), so a maximum of the nonnegative numbers $e_m$ (or $e^\angle_m$) over the smaller set, or $0$ if it is empty, is at most the maximum over the larger set (or $0$). The same holds for $A'$ and for the symmetric distances. So for $\kappa\in(0,t]$, $\delta_i(A,A';t)\le\delta_i(A,A';\kappa)$ (and likewise for $\delta^\angle_i$): the hypothesis of (c) at any such $\kappa$ implies the hypothesis at $\kappa=t$, which is therefore the weakest.
\item (iii), the example. Let $d=n=2$, $v_1=(1,0)^\top$, $v_2=(0,1)^\top$, $t\in(0,1)$, $\kappa\in(t,1]$ and $\Delta=0$, and consider the two one-atom dictionaries $[v_2]$ and $[(t,\sqrt{1-t^2})^\top]$. Their scores for feature $1$ are $0$ and $t$. As $0<\kappa$ and $t<\kappa$, the near sets of feature $1$ at $\kappa$ are empty for both, so $\delta_1=\delta^\angle_1=0\le\Delta$, whichever of the two is called $A$.
\item (iii), non-recovery half: take $A=[v_2]$, $A'=[(t,\sqrt{1-t^2})^\top]$. Then $c_1(A)=0<t=t-\Delta$ and $\angle_1(A)=\arccos0=\pi/2>\arccos t=\arccos t+\Delta$ (Tool~\ref{tool:trig}, $t>0$), yet $c_1(A')=t\ge t$.
\item (iii), recovery half: take $A=[(t,\sqrt{1-t^2})^\top]$, $A'=[v_2]$. Then $c_1(A)=t\ge t+\Delta$ and $\angle_1(A)=\arccos t\le\arccos t-\Delta$, yet $c_1(A')=0<t$.
\end{steps}
\end{proof}

\begin{proof}[Proof of (d)]
\begin{steps}
\item First inequality: by D7, $n\FRR_t(A)-n\FRR_t(A')=\sum_i(\one[c_i(A)\ge t]-\one[c_i(A')\ge t])$; each term lies in $\{-1,0,1\}$ and is nonzero exactly when the status of $i$ changes. The triangle inequality gives the bound.
\item If (R$\angle$) fires for $i$, then $c_i(A')\ge t$ by (a); and with $m$ attaining the minimum, $\angle(a_m,v_i)\le\arccos t$ (as $e^\angle_m\ge0$), so $c_i(A)\ge|\cos(a_m,v_i)|\ge t$ (Lemma~\ref{lem:T41}(ii)). The status of $i$ does not change.
\item If (NR$\angle$) fires for $i$, then $c_i(A')<t$ by (a); and $\angle_i(A)>\arccos t$ (as $\delta^\angle_i\ge0$), so $c_i(A)=\cos\angle_i(A)<t$ (Lemma~\ref{lem:T41}(ii), negation of the non-strict form: $\cos\vartheta<u\iff\vartheta>\arccos u$). The status of $i$ does not change.
\item By Steps~2--3, every feature whose status changes is one for which neither angular certificate fires. This gives the middle inequality.
\item By (b), if (R) fires then (R$\angle$) fires, and if (NR) fires then (NR$\angle$) fires. So every feature uncertified in the angular sense is uncertified in the chord sense, and $N^\angle_t\le N^{\mathrm{loc}}_t$.
\item By (b) again, if (C11-R) fires for some maximiser $m_i$ then (R) fires, and if (C11-NR) fires then (NR) fires. So every feature for which neither (R) nor (NR) fires is one for which neither (C11-R), for any $m_i$, nor (C11-NR) fires; this is the last claim of (d).
\end{steps}
\end{proof}

\begin{proof}[Proof of (e)]
\begin{steps}
\item Let $\mathcal I_1:=\{i:\delta_i(A\to A';t)>\gamma\}$ and $\mathcal I_2:=\{i:\delta_i(A'\to A;t)>\gamma\}$. By the definition of the symmetric distance, $\{i:\delta_i(A,A';t)>\gamma\}=\mathcal I_1\cup\mathcal I_2$, so its size is at most $\#\mathcal I_1+\#\mathcal I_2$.
\item Let $i\in\mathcal I_1$. Since $\gamma\ge0$ and $\delta_i(A\to A';t)>\gamma$, $\delta_i(A\to A';t)\ne0$, so $P_i(A;t)\ne\emptyset$ (an empty near set gives $0$). Choose a witness $m^\dagger(i)\in P_i(A;t)$ with $e_{m^\dagger(i)}(A\to A')=\delta_i(A\to A';t)>\gamma$.
\item $g_{m^\dagger(i)}(A)\ge|\cos(a_{m^\dagger(i)},v_i)|\ge t$ (D7 and $m^\dagger(i)\in P_i(A;t)$). So $m^\dagger(i)$ belongs to the first atom set in \eqref{eq:aggr1}.
\item $m^\dagger$ is injective on $\mathcal I_1$: if $m^\dagger(i)=m^\dagger(j)=m$ with $i\ne j$, then $|\cos(a_m,v_i)|\ge t$ and $|\cos(a_m,v_j)|\ge t$, contradicting Theorem~\ref{thm:C9a} (here $t>\cos(\theta_V/2)$ is used). So $\#\mathcal I_1$ is at most the first atom count.
\item The same argument with $A$ and $A'$ exchanged bounds $\#\mathcal I_2$ by the second atom count. Steps~1--5 give \eqref{eq:aggr1}.
\item For \eqref{eq:aggr2}: let $i$ change status. If $\delta_i(A,A';t)\le\gamma$, apply (c) with $\kappa=t$ and $\Delta=\gamma\ge0$: $c_i(A)\ge t+\gamma$ would give $c_i(A)\ge t$ and $c_i(A')\ge t$, and $c_i(A)<t-\gamma$ would give $c_i(A)<t$ and $c_i(A')<t$; both contradict the status change. So $t-\gamma\le c_i(A)<t+\gamma$.
\item So the set of status changes is contained in $\{i:t-\gamma\le c_i(A)<t+\gamma\}\cup\{i:\delta_i(A,A';t)>\gamma\}$. The first inequality of \eqref{eq:chain}, a union bound and \eqref{eq:aggr1} give \eqref{eq:aggr2}.
\end{steps}
\end{proof}

\begin{remark}[comparison of (e) with C11]\label{rem:eC11}\MkOursChk
(1) The margin set in \eqref{eq:aggr2} is half-open, $[t-\gamma,t+\gamma)$, against the open interval in $N_\gamma$; the atom counts use $g_m(A)\ge t$ (C11: $\ge t+\gamma$) and the strict $>\gamma$ on both sides (C11's $Y_\gamma$: $\ge\gamma$); and \eqref{eq:aggr2} adds the two atom counts where \eqref{eq:C11main} takes their maximum. Neither right side is at most the other in general: there are $V$, $t\in(\cos(\theta_V/2),1]$, $\gamma>0$ and dictionaries for which the right side of \eqref{eq:C11main} is strictly smaller than that of \eqref{eq:aggr2}, and others for which it is strictly larger. So we keep Theorem~\ref{thm:C11} as the aggregate statement; the gain of C11-loc is in (a)--(d). (2) For $t\in(\cos(\theta_V/2),1]$, $\gamma=0$ is allowed in \eqref{eq:aggr2}: then the margin set is empty and \eqref{eq:aggr2} bounds the change by the numbers of atoms with $g\ge t$ that have no exact partner line (distance $0$) in the other dictionary.
\end{remark}
\begin{proof}
In Steps~2--6, $d=n=2$, $v_1=\mathsf e_1$, $v_2=\mathsf e_2$ and $t=4/5$. Then $\theta_V=\arccos0=\pi/2$ and $\cos(\theta_V/2)=1/\sqrt2<4/5$, since $2\cdot16=32>25$ (Tool~\ref{tool:trig}, Tool~\ref{tool:mono}). So both \eqref{eq:C11main} (any $\gamma>0$) and \eqref{eq:aggr2} (any $\gamma\ge0$) apply.
\begin{steps}
\item (1), the structural differences: read off \eqref{eq:X}, \eqref{eq:Y}, D7 ($N_\gamma$) and \eqref{eq:aggr1}--\eqref{eq:aggr2}.
\item \emph{Example where \eqref{eq:C11main} is smaller.} $A=[v_1]$ ($M=1$), $A'=[v_2]$ ($M'=1$), $\gamma=1/10$. Scores: $c_1(A)=1$, $c_2(A)=0$, $g_1(A)=1$; $c_1(A')=0$, $c_2(A')=1$, $g_1(A')=1$. Distances: $\|\mathsf e_1\mp\mathsf e_2\|=\sqrt2$, so $e_1(A\to A')=e_1(A'\to A)=\sqrt2$.
\item In Step~2: neither $|1-4/5|=1/5$ nor $|0-4/5|$ is $<1/10$, so $N_\gamma(A;t)=0$; and neither $1$ nor $0$ lies in $[7/10,9/10)$, so the margin count of \eqref{eq:aggr2} is $0$. $X_\gamma=1$ ($g_1(A)=1\ge9/10$, $\sqrt2>1/10$) and $Y_\gamma=1$ ($g_1(A')=1\ge4/5$, $\sqrt2\ge1/10$); the two atom counts of \eqref{eq:aggr1} are $1$ and $1$. So the right side of \eqref{eq:C11main} is $0+\max(1,1)=1$ and that of \eqref{eq:aggr2} is $0+1+1=2$.
\item \emph{Example where \eqref{eq:aggr2} is smaller.} $A=[v_1,v_2]$ ($M=2$), $A'=[a'_1,v_2]$ ($M'=2$) with $a'_1=(99/101,20/101)^\top$, a unit vector ($99^2+20^2=101^2$), and $\gamma:=2/\sqrt{101}$. Note $\gamma<1/5$, since $10<\sqrt{101}$. Scores: $c_1(A)=c_2(A)=1$, $g_1(A)=g_2(A)=1$; $g_1(A')=99/101\ge4/5$ ($495\ge404$) and $g_2(A')=1$.
\item Distances (Step~1 of the proof of Lemma~\ref{lem:T42}: $\|\hat x-s\hat y\|^2=2-2s\langle\hat x,\hat y\rangle$): $\|\mathsf e_1-a'_1\|^2=2-198/101=4/101$, $\|\mathsf e_1+a'_1\|^2>2$, $\|\mathsf e_1\mp\mathsf e_2\|^2=2$ and $\|a'_1\mp\mathsf e_2\|^2=2\mp40/101\ge162/101>4/101$. Hence $e_1(A\to A')=e_1(A'\to A)=2/\sqrt{101}=\gamma$, and $e_2(A\to A')=e_2(A'\to A)=0$ (the atom $v_2$ is in both).
\item In Step~4: $|1-4/5|=1/5>\gamma$, so $N_\gamma(A;t)=0$; and $1\notin[4/5-\gamma,4/5+\gamma)$ because $4/5+\gamma<1$, so the margin count of \eqref{eq:aggr2} is $0$. $X_\gamma=0$: $e_1(A\to A')=\gamma\not>\gamma$ and $e_2(A\to A')=0$. $Y_\gamma=1$: $m'=1$ has $g_1(A')\ge4/5$ and $e_1(A'\to A)=\gamma\ge\gamma$; $m'=2$ has distance $0<\gamma$. The two atom counts of \eqref{eq:aggr1} are $0$, since no distance is $>\gamma$. So the right side of \eqref{eq:C11main} is $0+\max(0,1)=1$ and that of \eqref{eq:aggr2} is $0$.
\item (2): for $\gamma=0$ the margin set $\{i:t\le c_i(A)<t\}$ is empty, and the atom counts of \eqref{eq:aggr1} count atoms with $g\ge t$ and distance $>0$; Theorem~\ref{thm:C11loc}(e) allows $\gamma=0$.
\end{steps}
\end{proof}

\begin{remark}[an oracle in the hypothesis]\label{rem:oracle}\MkSpec
$\delta_i(A'\to A;t)$ and $\delta_i(A,A';\kappa)$ depend on $A'$. Like $H$ in Theorem~\ref{thm:C11}, they give certificates in the form ``a bound on a distance implies a conclusion''. On saved pairs where $A'$ is known, they test whether a local distance bound would be small enough. Bounding $\delta_i$ before $A'$ is trained (from a stability result for training, for example) is an open problem (\S\ref{sec:open}).
\end{remark}

\subsection{The area under the FRR curve}\label{ssec:area}

\begin{proposition}[area identity]\label{prop:area}\MkOursChk
For $A\in\cA_M$,
\begin{equation}\label{eq:area}
\int_0^1\FRR_t(A)\,\mathrm dt=\bar c(A).
\end{equation}
($\FRR_t$ is defined for $t\in(0,1]$; the value at $t=0$ does not affect the integral.) Consequently, for $A'\in\cA_{M'}$, the area satisfies $|\bar c(A)-\bar c(A')|\le H(A,A')$, a bound for which no threshold $t$ has to be chosen. The identity is an instance of the layer-cake representation; we claim no novelty.
\end{proposition}
\begin{proof}
\begin{steps}
\item By D7, $\FRR_t(A)=\frac1n\sum_{i=1}^n\one[c_i(A)\ge t]$ for $t\in(0,1]$; extend each indicator to $t=0$ by the same formula (Tool~\ref{tool:riemann}).
\item Each $c_i(A)\in[0,1]$ (D7), so $\int_0^1\one[c_i(A)\ge t]\,\mathrm dt=c_i(A)$ (Tool~\ref{tool:riemann}).
\item By linearity (Tool~\ref{tool:riemann}), $\int_0^1\FRR_t(A)\,\mathrm dt=\frac1n\sum_ic_i(A)=\bar c(A)$ (D7).
\item The bound on $|\bar c(A)-\bar c(A')|$ is Theorem~\ref{thm:C8}(b).
\end{steps}
\end{proof}

\section{T5: measurements on toy models and synthetic dictionaries}\label{sec:T5}

This section reports our own runs of our own code. Every number here is a measurement (marker ``measured''): it has not been reproduced by anyone else. For each finding we say what was measured, how, the value expected in advance and where that expectation comes from, and whether the check passed (PASS/FAIL) or is reported for information only (INFO). The point of the section is to show where the threshold metrics change their verdict and the stable ones do not, and how far the proved certificates reach in practice.

\subsection{The runs}\label{ssec:runs}

\begin{definition}[D9: runs]\label{def:D9}\MkDef
\begin{itemize}[itemsep=1pt]
\item \emph{Toy-model sweep $\mathcal E$.} The model is the ReLU output model $\mathbf h\mapsto\ReLU(W^\top W\mathbf h+\mathbf b)$ with $W\in\R^{30\times400}$ ($d=30$, $n=400$) and $\mathbf b\in\R^{400}$. The input $\mathbf h\in\R^{400}$ has independent coordinates, each equal to an independent $U[0,1]$ variable with probability $p_{\mathrm f}$ and to $0$ otherwise. The loss is the expected squared error between $\mathbf h$ and the output (all importances equal to $1$). The sweep has one trained model for each $1/p_{\mathrm f}\in\{1,1.25,1.5,1.75,2,2.5,3,3.5,4.5,6,8,10\}$, so 12 models. Training: AdamW with learning rate $10^{-3}$ and weight decay $0.01$, $10^4$ steps, batch 1024, Xavier-normal initialisation, $\mathbf b=0$ at the start, training seed $0$. The metrics are computed in float64 from the saved weights.
\item \emph{Dictionary runs $\mathcal G$.} All arithmetic is in float32, with PyTorch CPU pseudo-random generators.
 \begin{itemize}
 \item \emph{Ground truth.} $V\in\R^{100\times384}$ ($d=100$, $n=384$) is drawn once, from a generator seeded with $3000$: it draws a $384\times100$ array of i.i.d.\ standard normal entries, and $v_i$ is row $i$ divided by its norm, so the columns of $V$ are i.i.d.\ uniform on the unit sphere. $V$ is the same in every run.
 \item \emph{Data.} Each data point is $\mathbf x=\mathbf y/\|\mathbf y\|$ with $\mathbf y=3^{-1/2}(v_{j_1}+v_{j_2}+v_{j_3})$, where $j_1,j_2,j_3$ are i.i.d.\ uniform on $[384]$ (drawn with replacement). Every step draws a fresh batch of 512 i.i.d.\ data points; there is no fixed training set.
 \item \emph{Training seed.} One generator, seeded with the training seed $\mathfrak s\in\{0,1,2\}$, draws first the initial atoms and then all batches.
 \item \emph{Model.} The sparse autoencoder has tied weights: its atoms $a_1,\dots,a_{1024}\in\R^{100}$ serve as both encoder and decoder, with encoder biases $b^{\mathrm{enc}}_1,\dots,b^{\mathrm{enc}}_{1024}$ and no pre-bias. The reconstruction is $\mathbf x^{\mathrm{rec}}=\sum_{m=1}^{1024}a_m\ReLU(\mathrm{pre}_m)$ with $\mathrm{pre}_m=\langle a_m,\mathbf x\rangle+b^{\mathrm{enc}}_m$. The dictionary $A\in\cA_{1024}$ ($M=1024$) has columns $a_1,\dots,a_{1024}$.
 \item \emph{Initialisation.} The atoms are i.i.d.\ uniform on the unit sphere (normalised standard normal vectors), and all biases are $0$.
 \item \emph{Optimiser and loss.} AdamW on the atoms only, learning rate $10^{-3}$ (constant), moment coefficients $0.9$ and $0.999$, stabilising constant $10^{-8}$, weight decay $0.01$. The loss on a batch is the batch mean of $1-\cos(\mathbf x^{\mathrm{rec}},\mathbf x)$ plus $10^{-6}$ times the batch mean of $\sum_m\|a_m\|\ReLU(\mathrm{pre}_m)$. After every step each atom is replaced by $a_m/(\|a_m\|+10^{-8})$.
 \item \emph{Group bias adaptation (after Chen et al.\ \cite{chen2026}).} The biases are not trained by gradient. Atom $m$ belongs to group $\lfloor10(m-1)/1024\rfloor+1\in[10]$; group $k$ has target frequency $0.5\cdot(0.002)^{(k-1)/9}$. The adaptation interval is 50 steps (25\,600 data points). Over each interval, the firing fraction of atom $m$ is the fraction of the interval's data points with $\mathrm{pre}_m\ge10^{-6}$, and its peak is $\max(0,\max\mathrm{pre}_m)$, computed in the forward pass of each step. At the end of each interval an atom is \emph{never activated} if its firing fraction is below $1/25\,600$ and \emph{too often activated} if it exceeds its group's target frequency. In each group, every never-activated atom's bias is increased by $0.01$ times the mean peak of the group's atoms that are not never-activated (by $0.01\cdot10^{-3}$ if there are none); every too-often-activated atom's bias is decreased by $0.01$ times its own peak. Then every bias is clamped to $[-1,0]$.
 \item \emph{Read-out.} Dictionaries are read after 5000 and 10000 steps of the same run, for each $\mathfrak s\in\{0,1,2\}$.
 \end{itemize}
\item \emph{Synthetic dictionaries $\mathcal S_{\mathrm{syn}}$.} With the same $V$ and a NumPy generator seeded with $20261027$: $A\in\cA_{1024}$ consists of noisy copies of 346 of the 384 features (each $v_i$ plus an independent Gaussian vector with covariance $\epsilon_{\mathrm c}^2d^{-1}I_d$, where the copy noise scale $\epsilon_{\mathrm c}$ is drawn independently for each copy from $U[0,0.9]$, then normalised; the 38 remaining features have no copy), 20 bisector atoms $\hat v_i\pm\hat v_j$ of the 20 closest line pairs (sign chosen as in Remark~\ref{rem:T45}), and 658 random unit atoms; the atoms are then multiplied by random signs and by scales in $[0.5,2]$ and put in random order. For a noise level $\epsilon_{\mathrm{nl}}\in\{0,0.001,0.003,0.01,0.03,0.1,0.3\}$, $A'$ is obtained by adding to each normalised atom an independent Gaussian vector with covariance $(\epsilon_{\mathrm{nl}}^2/d)I_d$ and normalising; an eighth variant uses $\epsilon_{\mathrm{nl}}=0.03$, drops 64 atoms and adds 32 random unit atoms ($M'=992$).
\end{itemize}
\end{definition}

\subsection{The sum rule on the toy-model sweep}\label{ssec:T5sum}

\begin{empirical}[the earlier thresholded measurement]\label{emp:poc2}\MkMeas
\emph{Measured} (in our earlier document \cite{poc2doc}; not repeated here): the sum rule $\Dtot^{(\tau)}(W)$ against $d$ on its toy-model sweep, one model per inverse density $1/p_{\mathrm f}$ (called $x$ there). \emph{Expected:} $|\Dtot^{(0.2)}-d|/d\le0.15$. \emph{Observed:} with $\tau=0.2$ the thresholded sum $\Dtot^{(0.2)}$ fell $24.8\%$ short of $d$ at $1/p_{\mathrm f}=1.75$ ($\Dtot^{(0.2)}/d=0.7516$), while without the threshold $\Dtot\ge0.997\,d$ at every density. \emph{Status:} FAIL for the thresholded sum (script \texttt{el03\_uniform\_superposition.py}, sum-rule test row, $0.248$ against the tolerance $0.15$); INFO for the unthresholded sum (script \texttt{el10\_frame\_dimensionality.py} on the saved weights of that sweep).
\end{empirical}

\begin{empirical}[C13(b): the regularised sum rule gives one verdict over $\Lambda(W)$]\label{emp:C13b}\MkMeas
\emph{Measured:} $\Dttot(W;\lambda)/d$ for each of the 12 models of $\mathcal E$, at 21 log-spaced values $\lambda=f\,\omega(W)$, $f\in[10^{-3},10^{-1}]$ (both ends included), i.e.\ on a grid of $\Lambda(W)$ (D6). \emph{Expected} (our prediction C13(b), stated before the run): $\Dttot(W;\lambda)\ge0.85\,d$ for every model and every $\lambda\in\Lambda(W)$, so that the verdict ``$\Dttot\ge0.85\,d$'' is constant on $\Lambda(W)$ and equal to the verdict of the unthresholded test $\Dtot^{(0)}(W)\ge0.85\,d$. \emph{Observed:} minimum $0.90359$; verdict constant on the grid and equal to the unthresholded verdict for all 12 models. \emph{Status:} PASS.

\emph{INFO, same run.} $\Dtot^{(0)}/d\in[0.9974,0.9998]$; the norm-thresholded $\Dtot^{(0.2)}/d$ ranges from $0.7516$ ($1/p_{\mathrm f}=1.75$) and $0.7638$ ($1/p_{\mathrm f}=1.5$) up to $0.9987$, so the thresholded verdict ``$\ge0.85\,d$'' fails for two models. The minimum over $\Lambda(W)$ of $\Dttot/d$ lies in $[0.9036,0.9090]$ for every model. Outside $\Lambda(W)$, at $\lambda=0.3\,\omega$ and $\lambda=\omega$, $\Dttot/d$ falls to about $0.77$ and $0.50$, as the ridge term of \eqref{eq:C6c} predicts.
\end{empirical}

\begin{empirical}[C13(c) and checks of C6]\label{emp:C13c}\MkMeas
\emph{Measured:} on the same grid, the eigenvector term of \eqref{eq:C6c} divided by $d$; the residual of the identity \eqref{eq:C6c}; the smallest of its three terms; the largest violation of Theorem~\ref{thm:C6a}; and $\max_i(\Dt_i-\ell_i)$. \emph{Expected:} eigenvector term $\le0.01\,d$ (our prediction C13(c)); residual $\le10^{-9}d$, terms $\ge-10^{-9}$, violation $\le10^{-9}$ and $\max(\Dt_i-\ell_i)\le10^{-9}$ (Theorems~\ref{thm:C6a} and \ref{thm:C6bd}, up to rounding). \emph{Observed:} $0.0025608$; $5.6\cdot10^{-16}$; $0$; $0$; $0$. \emph{Status:} PASS (all five).
\end{empirical}

\subsection{Lipschitz constants: searches for counterexamples}\label{ssec:T5lip}

\begin{empirical}[no pair exceeds the constants of Theorems~\ref{thm:C4b} and \ref{thm:C5}]\label{emp:lip}\MkMeas
\emph{Measured:} the largest scale-free ratios $\sqrt\lambda\,|\ell_i(W;\lambda)-\ell_i(W';\lambda)|/\|W-W'\|_{\op}$ and $\sqrt\lambda\,|\deff(W;\lambda)-\deff(W';\lambda)|/\|W-W'\|_*$, over random pairs and gradient ascent on the ratio, in shapes $1\times1$, $2\times2$, $2\times3$, $3\times2$, $5\times5$, $10\times20$ and $30\times400$, and from the 12 models of $\mathcal E$. \emph{Expected:} $\le3\sqrt3/8=0.6495190528$ (plus $10^{-9}$) for both (Theorems~\ref{thm:C4b} and \ref{thm:C5}); a $1\times1$ search should reach at least $0.99\cdot3\sqrt3/8$ (Proposition~\ref{prop:scalar}). \emph{Observed:} $0.6495189310535826$ and $0.6495190507646841$; the $1\times1$ search reached $0.6495189$. \emph{Status:} PASS (all four). \emph{INFO:} every shape reached $0.64952$ and none exceeded it; in $1\times1$ and $2\times2$ the maximum sits at $\rho\approx0.578\approx1/\sqrt3$, in larger shapes at larger $\rho$, as Remark~\ref{rem:A4tight} explains (the scalar extremum embeds with orthogonal other columns).
\end{empirical}

\begin{empirical}[the profile of $\Dt_i$]\label{emp:C7}\MkMeas
\emph{Measured} (two runs): (1) in the first run, the slope of the family of Proposition~\ref{prop:C7b} by automatic differentiation for $R_0\in\{1.5,2,4,10,30,100\}$ ($\lambda=1$), and over all pairs found, $\min(2+\rho-\sqrt\lambda\cdot\text{ratio})$; (2) in the second run, fixed-$R$ gradient ascent on $\sqrt\lambda\,|\Dt_1(W;\lambda)-\Dt_1(W';\lambda)|/\|W-W'\|_{\op}$ at $\rho\in\{0.5,1,2,4,8,16,32,64,128,256\}$ in shapes $2\times2$, $3\times2$, $2\times3$, $3\times3$, started from \textsf{F2}, from the pairs saved in run (1) and from random points (286 ascent runs, 701 pairs filed), together with the \textsf{F1}/\textsf{F2} slopes by automatic differentiation. \emph{Expected:} slope equal to $\frac{2R_0}{9\lambda}\sqrt{1-\lambda/R_0^2}$ to relative $10^{-8}$ (Proposition~\ref{prop:C7b}); slack $\ge0$ (Corollary~\ref{cor:A77}(c)); \textsf{F1}/\textsf{F2} slopes equal to Corollary~\ref{cor:special}(3)--(4) to relative $10^{-8}$; $\max(\sqrt\lambda\cdot\text{ratio}-\rho/4)\le\sqrt{31}/8=0.69597$ and $\le2$ (the conjectured forms (ii) and (ii-w) of C7$'$, stated before the run; Theorem~\ref{thm:A76} later gave $3\sqrt3/8=0.6495$). \emph{Observed:} slopes equal in all six cases; slack $1.7938$; \textsf{F1}/\textsf{F2} maximal relative error $4.2\cdot10^{-16}$; $\max(\sqrt\lambda\cdot\text{ratio}-\rho/4)=0.515005$, which is also below $3\sqrt3/8$. \emph{Status:} PASS (all).

\emph{INFO.} The largest excess $0.515005$ occurs at $\rho=0.5016$ for the scalar pair $d=n=1$, the value $\chi^\star_s$ of Derivation~\ref{der:chis}; no pair in any shape exceeded it (consistent with Conjecture~\ref{conj:C7pp}). $\hat\Phi(\rho)/\rho=0.25275$, $0.25138$, $0.25069$ at $\rho=64,128,256$; $\hat\chi(\rho)-1/(4\sqrt2)=0.01756$, $-0.00196$, $-0.00098$, $-0.00049$, $-0.00025$ at $\rho=16,32,64,128,256$ (hats denote the best values found), consistent with Theorem~\ref{thm:A78} and Proposition~\ref{prop:profile}(iii). At $\rho\ge32$ the best pair found is the \textsf{F2} pair itself.
\end{empirical}

\subsection{Certificates on dictionaries}\label{ssec:T5cert}

\begin{empirical}[the setting]\label{emp:setting}\MkMeas
\emph{Measured:} for the $V$ of D9, $\theta_V=1.1500$, $\cos(\theta_V/2)=0.839182$ and $\talign(V)=0.927418$. Since $\talign(V)>\cos(\theta_V/2)$, Theorems~\ref{thm:C9a} and \ref{thm:C11} apply at $t=\talign(V)$. Code checks (PASS): $V$ rebuilt from seed $3000$ is bit-identical to the $V$ used; the run with $\mathfrak s=2$ reproduces an earlier trace of the same configuration ($\FRR_{0.8}=0.9974$ and $\FRR_{\talign}=0.65625$ at 5000 steps; $0.97396$ and $0.21875$ at 10000 steps; tolerance $2/384$); GBA biases stay in $[-1,0]$ and atom norms equal $1$ to $1.8\cdot10^{-7}$.
\end{empirical}

\begin{empirical}[checks of C8 and C9(a)]\label{emp:C8C9}\MkMeas
\emph{Measured:} on $\mathcal S_{\mathrm{syn}}$, $\max_i|c_i(A)-c_i(\tilde A)|$ for a permuted, signed and rescaled copy $\tilde A$; over the eight variants, $\max_i|c_i(A)-c_i(A')|-H(A,A')$ and $|\bar c(A)-\bar c(A')|-H(A,A')$; on $\mathcal G$, the same two quantities over all pairs of dictionaries; and, on $\mathcal S_{\mathrm{syn}}$ (96 dictionary--threshold pairs, including $t=\cos(\theta_V/2)+10^{-6}$; the 20 bisector atoms make the test sharp) and on every trained dictionary, the number of atoms with $|\cos(a_m,v_i)|\ge t$ for two features at a valid $t$. \emph{Expected:} $\le10^{-12}$; $\le10^{-9}$; $\le10^{-9}$; $0$ (Theorems~\ref{thm:C8} and \ref{thm:C9a}). \emph{Observed:} $4.4\cdot10^{-16}$; $1.7\cdot10^{-16}$ and $0$ (synthetic), $-0.992$ and $-1.188$ (trained); $0$ and $0$. \emph{Status:} PASS (all).
\end{empirical}

\begin{empirical}[C11 is never wrong but certifies nothing across seeds]\label{emp:C11}\MkMeas
\emph{Measured:} for every valid $t$ (on $\mathcal S_{\mathrm{syn}}$: $0.85$, $0.9$, $\talign(V)$, $0.95$, $0.98$, $\cos(\theta_V/2)+10^{-6}$) and a grid of $\gamma$, the inequality \eqref{eq:C11main}, the two bounds \eqref{eq:C11markov} and the two certificates of Theorem~\ref{thm:C11}, on the eight synthetic pairs and on the trained pairs across seeds. \emph{Expected:} no violation and no wrong certificate (Theorem~\ref{thm:C11}). \emph{Observed:} synthetic: 0 violations among 864 checks (minimum slack $0$: the bound is attained), 0 violations of the Markov bounds, 0 wrong among 43\,838 certificates; trained: 0 violations among 432 checks (minimum slack $28$), 0 violations of the Markov bounds, 0 wrong certificates, among \textbf{0} certified features. \emph{Status:} PASS.

\emph{INFO.} Between independently trained dictionaries $H(A,A')=1.2143$, $1.2082$, $1.2130$ (seed pairs $(0,1)$, $(0,2)$, $(1,2)$, 5000 steps) and $1.1995$, $1.1966$, $1.1938$ (10000 steps), close to the maximum $\sqrt2$: each dictionary has unaligned atoms with no close partner in the other. All 384 features are uncertified for every pair and every $t$. At $t=\talign(V)$ and 10000 steps, $\min_\gamma(N_\gamma+\max(X_\gamma,Y_\gamma))$ is $156/150$, $107/114$ and $150/156$ (the two directions) against actual changes $n|\Delta\FRR_t|=31$, $12$, $43$; over all valid $t$ the bound ranges from $32$ to $540$ against actual $0$ to $43$. On $\mathcal S_{\mathrm{syn}}$ at $\epsilon_{\mathrm{nl}}=0.1$, $H=0.1205$ and the ``recovered'' certificate fires for no feature at $t=\talign(V)$.
\end{empirical}

\begin{empirical}[C11-loc: valid everywhere, informative only on synthetic pairs]\label{emp:C11loc}\MkMeas
\emph{Measured:} for every certificate kind of Theorem~\ref{thm:C11loc}(a) and (c) (chord and angular; $t\in\{0.85,0.9,\talign(V),0.95\}$; $\kappa\in\{t,t-0.02,t-0.05,\cos(\theta_V/2)+10^{-6}\}$ with $\cos(\theta_V/2)<\kappa\le t$), the number of firings and of wrong firings; the inclusions (b) per feature; the chain \eqref{eq:chain}; and \eqref{eq:aggr2} for 136 cases and 9 values of $\gamma$; on the trained pairs (across seeds at equal steps, and the same seed at 5000 and 10000 steps, both directions), on the eight synthetic pairs, on 6000 random instances and in an adversarial search of $16\times150$ steps per certificate kind. \emph{Expected:} 0 wrong firings, 0 violations (Theorem~\ref{thm:C11loc}). \emph{Observed:} 0 wrong among 24\,112 (R), 78\,500 (NR), 29\,772 (R$\angle$), 78\,500 (NR$\angle$), 24\,112, 58\,330, 29\,379 and 59\,675 (the four a priori forms) firings; 0 violations of (b), of \eqref{eq:chain} and of \eqref{eq:aggr2}; 0 in the random and adversarial searches; the C11 counts agree with an independent implementation. \emph{Status:} PASS (all).

\emph{INFO} (mean over ordered pairs, $t=\talign(V)$): features certified recovered by (R), by (R$\angle$), by the a priori angular form with a nonempty near set (for $\kappa$ from $t$ down to $\cos(\theta_V/2)+10^{-6}$), and by (C11-R); and the number of status changes.
\begin{center}
\begin{tabular}{@{}lrrrrr@{}}
\toprule
Pair type & (R) & (R$\angle$) & a priori $\angle$ & (C11-R) & changes\\
\midrule
across seeds, 5000 steps & 0 & 0 & 0 & 0 & 162.7\\
across seeds, 10000 steps & 0 & 0 & 0 & 0 & 142.0\\
same seed, 5000 vs 10000 steps & 0 & 0 & 0 & 0 & 190.7\\
synthetic, $\epsilon_{\mathrm{nl}}=0$ & 141 & 141 & 141--253 & 141 & 0\\
synthetic, $\epsilon_{\mathrm{nl}}=0.001$ & 139 & 140 & 140--252 & 139 & 0\\
synthetic, $\epsilon_{\mathrm{nl}}=0.003$ & 135 & 139 & 139--250 & 135 & 0\\
synthetic, $\epsilon_{\mathrm{nl}}=0.01$ & 127 & 134 & 134--239 & 127 & 2\\
synthetic, $\epsilon_{\mathrm{nl}}=0.03$ & 105.5 & 127 & 127--225 & 105.5 & 2\\
synthetic, $\epsilon_{\mathrm{nl}}=0.1$ & 0 & 96.5 & 96.5--157 & 0 & 5\\
synthetic, $\epsilon_{\mathrm{nl}}=0.3$ & 0 & 15 & 15 & 0 & 55\\
synthetic, $\epsilon_{\mathrm{nl}}=0.03$, drop 64, add 32 & 98.5 & 119 & 119--212.5 & 98.5 & 11\\
\bottomrule
\end{tabular}
\end{center}
So on synthetic perturbations the local certificates gain over C11 (at $\epsilon_{\mathrm{nl}}=0.1$, C11 certifies no feature recovered and (R$\angle$) certifies $96.5$), while on trained dictionaries they certify no feature non-trivially.
\end{empirical}

\subsection{Instability of $\FRR$ at $\talign$}\label{ssec:T5inst}

\begin{empirical}[seed instability]\label{emp:seed}\MkMeas
\emph{Measured:} on $\mathcal G$, $\FRR_t$ and $\bar c$ for each seed and checkpoint (float64 analysis), and, at $t=\talign(V)$ and 10000 steps, the fraction of the 384 features recovered by at least one seed's dictionary but not by all three, the largest pairwise difference of $\bar c$, and for each seed pair the number of features whose status differs. \emph{Expected:} none stated in advance; this is not a claim. \emph{Status:} INFO.
\begin{center}
\begin{tabular}{@{}cccccc@{}}
\toprule
$\mathfrak s$ & steps & $\FRR_{0.8}$ & $\FRR_{0.9}$ & $\FRR_{\talign}$ & $\bar c$\\
\midrule
0 & 5000 & 1.0000 & 0.8359 & 0.6745 & 0.9348\\
0 & 10000 & 0.9766 & 0.6406 & 0.2500 & 0.9017\\
1 & 5000 & 0.9948 & 0.8464 & 0.6875 & 0.9339\\
1 & 10000 & 0.9818 & 0.6875 & 0.3307 & 0.9067\\
2 & 5000 & 0.9974 & 0.8490 & 0.6562 & 0.9325\\
2 & 10000 & 0.9740 & 0.6432 & 0.2188 & 0.9013\\
\bottomrule
\end{tabular}
\end{center}
At 10000 steps and $t=\talign(V)$: $55.47\%$ of the features are recovered by some seed but not by all, while $\bar c$ differs by at most $0.0054$ between seeds. The number of features whose status differs is $151$, $134$, $141$ for the seed pairs $(0,1)$, $(0,2)$, $(1,2)$ (35--39\% of the features), against net changes $n|\Delta\FRR_t|=31$, $12$, $43$; at 5000 steps it is $159$, $171$, $158$.
\end{empirical}

\begin{empirical}[instability along one training run]\label{emp:run}\MkMeas
\emph{Measured:} on $\mathcal G$, at $t=\talign(V)$, comparing each seed's dictionary at 5000 and at 10000 steps: the number of status changes, the net change, the local certificates of Theorem~\ref{thm:C11loc} for every $\kappa$ of the grid, $H(A,A')$, the median local angular distance $\delta^\angle_i$ at $\kappa=t$ and the median angular margin $|\angle_i(A)-\arccos t|$. \emph{Expected:} none stated in advance. \emph{Status:} INFO. \emph{Observed:} status changes $189$, $179$, $204$ for $\mathfrak s=0,1,2$ (mean $190.67$); net $n|\Delta\FRR_t|=163$, $137$, $168$; so about $176$, $158$, $186$ features lose their recovered status and $13$, $21$, $18$ gain it ((changes $\pm$ net)$/2$). (R), (R$\angle$) and the a priori recovered forms fire $0$ times; every (NR)/(NR$\angle$) firing ($112$, $99$, $114$) has an empty near set, so it only restates $c_i(A')<t$ (Remark~\ref{rem:T47}). $H(A,A')=1.186$, $1.176$, $1.192$; the median $\delta^\angle_i$ at $\kappa=t$ is $21.1^\circ$--$22.2^\circ$, against a median angular margin of $2.8^\circ$--$4.5^\circ$.
\end{empirical}

\begin{heuristic}[reading of the instability]\label{heu:reading}\MkSpec
The threshold metric $\FRR_{\talign}$ hides a per-feature turnover several times larger than its own change, while the threshold-free area $\bar c$ is stable to about $0.6\%$. This matches Theorem~\ref{thm:C8}(b) and Proposition~\ref{prop:area} qualitatively: $\bar c$ is Lipschitz in $H$, $\FRR_t$ is not continuous. Along one run, most of the change between the two checkpoints is a systematic drop ($\FRR_{\talign}$ falls from $0.66$--$0.69$ to $0.22$--$0.33$), with a smaller turnover in the other direction; the local certificates cannot see either, because the atoms near each feature move by about $20^\circ$ between checkpoints. Seed dependence of recovered features was reported before: Paulo and Belrose \cite{paulo2026} (real LLMs, Hungarian matching), Gerasimov et al.\ \cite{gerasimov2026} (Preprint (arXiv:2606.12138)), and Chen et al.\ \cite{chen2026}, \S4 (consistency across runs, on neurons of Qwen2.5-1.5B). We claim no novelty for the phenomenon.
\end{heuristic}

\section{Contributions and open problems}\label{sec:open}

This section lists, part by part, what each statement of this document adds, how far it has been checked and how it stands against the literature; it closes with the open problems. Each row gives (a) what is new, in one line; (b) the label of the statement; (c) the status: ``proved, blind-checked'' (marker $\checkmark$), ``proved, check pending'' (marker $\circ$), ``conjecture'' or ``measured'', with the markers explained in Appendix~\ref{app:labels}; and (d) the novelty verdict of a literature search. The verdict is \emph{solved} if a source (peer-reviewed, or a preprint labelled as such) states and proves the claim, or a statement that implies it under hypotheses that include ours; \emph{at risk} if a source states a close result (a special case, the same claim under stronger hypotheses, or the same mechanism for another metric), so that what is left is routine, or a source states the claim without proving it, or a relevant source's full text was not seen; \emph{open} if none of these holds after the recorded searches, which means ``not found'', not ``proved absent''. ``Routine'' qualifies a verdict; it does not by itself change it. A verdict is evidence about our search, not a proof of priority. Where a row says ``not audited'', the statement is a classical fact, an explicit computation or a measurement of our own code, and was not sent to the search.

\newcommand{\stC}{proved, blind-checked ($\checkmark$)}
\newcommand{\stP}{proved, check pending ($\circ$)}
\newcommand{\stCR}{proved, blind-checked ($\checkmark$); statement from a preprint}
\newcommand{\stNA}{not sent to the search (classical facts and explicit computations; no novelty claimed)}
\footnotesize
\setlength{\LTpre}{4pt}\setlength{\LTpost}{4pt}
\begin{longtable}{@{}p{4.5cm}p{2.5cm}p{2.3cm}p{5.2cm}@{}}
\toprule
{}(a) What is new & (b) Label & (c) Status & (d) Novelty verdict\\
\midrule
\endfirsthead
\toprule
{}(a) What is new & (b) Label & (c) Status & (d) Novelty verdict\\
\midrule
\endhead
\bottomrule
\endfoot
\multicolumn{4}{@{}l}{\textbf{Preliminaries (Sections~\ref{sec:prelim} and \ref{ssec:angle})}}\\
Elementary facts P1--P4; the slopes of the $2\times2$ family and their specialisations; the line-angle lemmas, including the triangle inequality for the line angle. & Lemmas~\ref{lem:P1}--\ref{lem:P4}, \ref{lem:family}; Cor.~\ref{cor:special}; Lemmas~\ref{lem:T41}, \ref{lem:triangle}--\ref{lem:T44} & \stC & \stNA.\\
\midrule
\multicolumn{4}{@{}l}{\textbf{T1: where feature dimensionality is discontinuous (Section~\ref{sec:T1})}}\\
Exact discontinuity set of $D_i$, with jump $1$ of $D_i$ and of the sum rule; local Lipschitz modulus of order $R/\|W_i\|^2$, and the order is needed. & Thm~\ref{thm:C1} & \stC & Discontinuity (a), (b): \emph{at risk}. Scherlis et al.\ \cite{scherlis2022} remark informally that a vanishing column gives $D_i\approx0$; the exact set is routine once stated. Lipschitz bound (c) and lower bound (d): \emph{open}; no perturbation bound for $D_i$ found.\\
The best coefficient $L_c^{\mathrm{best}}$ of the local modulus lies in $[1/2,4]$. & Rem.~\ref{rem:C1c} & \stC & \emph{Open}; no perturbation bound for $D_i$ found.\\
The precise implication that fails in an informal remark of Scherlis et al.\ \cite{scherlis2022}. & Rem.~\ref{rem:scherlis} & \stP & Not separately audited; it restates Thm~\ref{thm:C1}(b) (\emph{at risk}, as above).\\
\midrule
\multicolumn{4}{@{}l}{\textbf{T2: stable metrics (Section~\ref{sec:T2})}}\\
Ridge leverage lies in $[0,1)$ and vanishes only on a zero column. & Prop.~\ref{prop:C4a} & \stC & \emph{At risk}: a routine textbook fact about the ridge hat matrix, whose trace El Alaoui and Mahoney \cite{elalaoui2015} take as the effective dimension. No source found that states it with proof.\\
The scalar slope $3\sqrt3/(8\sqrt\lambda)$ is the least constant for $1\times1$ matrices and no pair attains it. & Prop.~\ref{prop:scalar} & \stP & \emph{Open} (routine calculus: the maximum of one derivative); no source found.\\
Sharp, shape-free Lipschitz constant $3\sqrt3/(8\sqrt\lambda)$ of ridge leverage in the operator norm, with two independent proofs. & Thm~\ref{thm:C4b}, Thm~\ref{thm:B4} & \stC & \emph{Open}. We found no source with a shape-free constant for ridge leverage (medium confidence: the literature on leverage scores is large).\\
Where the bound of Thm~\ref{thm:C4b} is attained. & Rem.~\ref{rem:A4tight} & \stC & \emph{Open}; nothing found.\\
Lipschitz bound $1/\sqrt\lambda$ for ridge leverage, as a corollary. & Cor.~\ref{cor:C4a} & \stC & \emph{Open}; weaker than Thm~\ref{thm:C4b}, nothing found.\\
One-sided leverage bounds and when the comparison function of the variational proof reaches the constant. & Rem.~\ref{rem:B4} & \stP & \emph{Open}; the first identity is routine, the sharp one-sided bound was not found.\\
The effective dimension is Lipschitz in the nuclear norm with the same sharp constant. & Thm~\ref{thm:C5} & \stC & \emph{At risk}: a routine consequence of the scalar constant and Mirsky's inequality \cite{mirsky1960}.\\
Elementary properties of the regularised dimensionality $\Dt_i$ (sandwich with $D_i$, monotone in $\lambda$, continuous, rotation invariant). & Thm~\ref{thm:C6a} & \stC & \emph{Open} (routine); $\Dt_i$ is our definition.\\
$\Dt_i\le\ell_i$ with its equality case, and the regularised defect identity for the sum rule. & Thm~\ref{thm:C6bd} & \stCR & \emph{At risk}: the inequality is Cauchy--Schwarz, and the identity is a corollary of Ivanov et al.\ \cite{ivanov2026}, Thm~8; the case $\lambda=0$ of the sum bound is in \cite{scherlis2022}. We claim no novelty.\\
Scale invariance of the profile $\Phi$ of the Lipschitz constant of $\Dt_i$, and lower bounds on $\Phi$ (coefficient $1/4$, additive constant $1/(4\sqrt2)$, scalar case). & Prop.~\ref{prop:profile} & \stC & \emph{Open}; nothing found (the scaling part is routine).\\
The dependence on $R$ cannot be removed: the slope of an explicit $2\times2$ family. & Prop.~\ref{prop:C7b} & \stP & \emph{Open}; an explicit computation in a toy family, nothing found.\\
Reduction to $\lambda=1$, a first-order bound at a point, the passage to a segment, and the coefficient bounds. & Lemmas~\ref{lem:A71}--\ref{lem:A75} & \stC & \emph{Open}; nothing found (reduction and segment lemmas are routine; the coefficient lemmas are routine calculus).\\
Lipschitz bound $\frac{3\sqrt3}{8\sqrt\lambda}+\frac R{4\lambda}$ for $\Dt_i$ with the coefficient $1/4$ optimal, and the asymptotic additive constant $1/(4\sqrt2)$. & Thm~\ref{thm:A76}, Cor.~\ref{cor:A77}, Thm~\ref{thm:A78} & \stC & \emph{Open}; no Lipschitz result found for $D_i$ or for any regularised version.\\
Combined bound and the structure of $\mathsf{F2}$; $\mathsf{F1}$ and $\mathsf{F2}$ are asymptotically extremal. & Rem.~\ref{rem:A7loss}, Rem.~\ref{rem:F1F2} & \stC & \emph{Open}; nothing found.\\
Comparison of the two proofs of the coefficient $1/4$. & Rem.~\ref{rem:A7vsB7} & \stP & \emph{Open}; nothing found.\\
Second proof of the upper half of the coefficient $1/4$. & Thm~\ref{thm:B7} & \stP & \emph{Open}; nothing found.\\
The scalar value $\chi^\star_s$ of the additive constant: root of a cubic, numerical inequalities, comparison lemma. & Der.~\ref{der:chis}, Lemmas~\ref{lem:A80}, \ref{lem:A81} & \stC & \emph{Open} (routine calculus and arithmetic); no content of their own.\\
The bound with the scalar additive constant $\chi^\star_s$ holds when $d=1$ or $n=1$, for small radius and for large radius. & Thms~\ref{thm:A82}, \ref{thm:A83}, \ref{thm:A84} & \stC & \emph{Open}; nothing found.\\
What is proved of the additive constant, and that none of the three upper bounds listed there gives C7$''$ on $(\rho_1,\rho_2)$. & Cor.~\ref{cor:C7pp} & \stP & \emph{Open}; nothing found.\\
\midrule
\multicolumn{4}{@{}l}{\textbf{T4: certificates for the feature recovery rate (Section~\ref{sec:T4})}}\\
Feature scores and the area $\bar c$ are Lipschitz in the line-Hausdorff distance; the smoothed rate is $1/\beta$-Lipschitz. & Thm~\ref{thm:C8} & \stC & \emph{At risk} (routine): the same mechanism as the 1-Lipschitz property of a distance to a set; the one-sided distance is that of Pali and Schnass \cite{pali2023}. No source found that states the bound for the smoothed rate or $\bar c$.\\
Above $\cos(\theta_V/2)$ an atom is aligned with at most one feature. & Thm~\ref{thm:C9a} & \stC & \emph{At risk} (routine): the triangle inequality for line angles; Chen et al.\ \cite{chen2026} state a uniqueness sentence above their threshold $\talign$, without proof. Under the absolute-coherence reading their threshold is $\cos(\theta_V/3)$, above ours (Remark~\ref{rem:chen}).\\
The threshold $\cos(\theta_V/2)$ is sharp for every $V$. & Rem.~\ref{rem:T45} & \stC & \emph{At risk} (routine): the equality case of the triangle inequality; no source found that states sharpness.\\
The uniqueness sentence of Chen et al.\ \cite{chen2026} can fail when signed and absolute cosines are mixed; it holds when $\talign(V)>\cos(\theta_V/2)$. & Rem.~\ref{rem:chen} & \stP & Counterexample: \emph{open}; no source notes the mismatch (the final published text of \cite{chen2026} was not seen by the search). Proof of the sentence under the stated condition: \emph{at risk}, since \cite{chen2026} states it without proof.\\
Global certificate: $n|\Delta\FRR_t|\le N_\gamma+\max(X_\gamma,Y_\gamma)$, with Markov bounds and per-feature certificates. & Thm~\ref{thm:C11} & \stC & \emph{Open}; no source found that bounds the change of a thresholded recovery count between two dictionaries.\\
Strict and non-strict inequalities, and the split form of the global certificate. & Rem.~\ref{rem:C11} & \stC & \emph{Open}; nothing found.\\
Feature-local certificates (chord and angular), their implications, an a-priori form and an atom-count bound. & Thm~\ref{thm:C11loc} & \stC & \emph{Open}; nothing found (the mechanism is again the triangle inequality plus margins).\\
Empty near sets; edge cases of the local certificates, with a counterexample. & Rem.~\ref{rem:T47}, Rem.~\ref{rem:T48} & \stC & \emph{Open}; elementary properties of our own definitions, nothing found.\\
Comparison of the a-priori local certificate with the global one. & Rem.~\ref{rem:eC11} & \stC & \emph{Open}; a comparison of our own bounds.\\
The area $\bar c$ equals $\int_0^1\FRR_t\,\mathrm dt$, and is Lipschitz in $H$. & Prop.~\ref{prop:area} & \stC & \emph{Solved}, as a routine consequence of the tail-integral identity for a variable in $[0,1]$; the Lipschitz half is Thm~\ref{thm:C8}(b).\\
\midrule
\multicolumn{4}{@{}l}{\textbf{T5: measurements (Section~\ref{sec:T5})}}\\
The regularised sum rule gives one verdict over the ridge range on 12 trained toy models. & Findings~\ref{emp:C13b}, \ref{emp:C13c} & measured & Not audited (measurement of our own code).\\
No pair exceeds the proved Lipschitz constants; slopes of the families agree with the formulas. & Findings~\ref{emp:lip}, \ref{emp:C7} & measured & Not audited (measurement of our own code).\\
The setting satisfies the hypothesis of Thm~\ref{thm:C9a}; checks of Thms~\ref{thm:C8} and \ref{thm:C9a}. & Findings~\ref{emp:setting}, \ref{emp:C8C9} & measured & Not audited (measurement of our own code).\\
The certificates were never wrong, and certify no feature non-trivially across seeds or checkpoints. & Findings~\ref{emp:C11}, \ref{emp:C11loc} & measured & Not audited (measurement of our own code).\\
Per-feature turnover of $\FRR$ at $\talign$ across seeds and checkpoints, against a stable $\bar c$. & Findings~\ref{emp:seed}, \ref{emp:run} & measured & Not audited. We claim no novelty for the phenomenon (Heuristic~\ref{heu:reading}).\\
\midrule
\multicolumn{4}{@{}l}{\textbf{Conjectures}}\\
The additive constant of $\Dt_i$ is set by the scalar case. & Conj.~\ref{conj:C7pp} & conjecture & \emph{Open}; nothing found that proves or refutes it.\\
The same on the open range $(\rho_1,\rho_2)$ for $d,n\ge2$; equivalent to the previous one given Cor.~\ref{cor:C7pp}. & Conj.~\ref{conj:C7ppopen} & conjecture & \emph{Open}; nothing found.\\
A sufficient two-variable inequality, a candidate for a computer-assisted check. & Conj.~\ref{conj:twovar} & conjecture & \emph{Open}; nothing found.\\
\end{longtable}
\normalsize

\subsection{Limitations}\label{ssec:limits}

\begin{itemize}[leftmargin=*]
\item \emph{Setting.} All measurements are on toy models and synthetic dictionaries with known true features (Section~\ref{sec:T5}).
\item \emph{Trained dictionaries.} The trained dictionaries are small, with three seeds each, and the toy-model sweep has one model per density and one seed (Definition~\ref{def:D9}).
\item \emph{Seed dependence.} At $t=\talign(V)$, $55.47\%$ of the features are recovered by some seed but not by all three (Finding~\ref{emp:seed}).
\item \emph{No language model.} No real language model was tested (Definition~\ref{def:D9}).
\item \emph{The certificates need the second dictionary.} The local distances in the certificates of Theorems~\ref{thm:C11} and \ref{thm:C11loc} are computed from both dictionaries, so the certificates check a given change of dictionary but do not predict one (Remark~\ref{rem:oracle}).
\item \emph{No certificate across independent training runs.} On our trained dictionaries the certificates certify no feature non-trivially across independently trained dictionaries, nor between two checkpoints of one run (Findings~\ref{emp:C11}, \ref{emp:C11loc} and \ref{emp:run}).
\item \emph{The open range of C7$''$.} The conjectured additive constant is proved for $\rho\in(0,\rho_1]\cup[\rho_2,\infty)$ and for $d=1$ or $n=1$; the range $(\rho_1,\rho_2)$ with $d,n\ge2$ is open (Theorems~\ref{thm:A82}--\ref{thm:A84}, Corollary~\ref{cor:C7pp}, Conjecture~\ref{conj:C7ppopen}).
\end{itemize}

\subsection{Open problems}\label{ssec:open}

The three conjectures above are the first open problem below. These are the open problems we identified.

\begin{enumerate}[leftmargin=*]
\item \emph{The additive constant of $\Dt_i$ on $(\rho_1,\rho_2)$.} Conjectures~\ref{conj:C7ppopen} and \ref{conj:twovar}. The two-variable inequality of Conjecture~\ref{conj:twovar} is a candidate for a computer-assisted check by interval arithmetic; if it fails, some of the losses of Heuristic~\ref{heu:A7loss} must be removed.
\item \emph{The optimal coefficient in the local modulus of $D_i$.} The infimum $L_c^{\mathrm{best}}$ lies in $[1/2,4]$ (Remark~\ref{rem:C1c}).
\item \emph{C5 without Mirsky.} Theorem~\ref{thm:C5} rests on the tool \eqref{eq:mirsky}, whose source we have not read; a self-contained proof (for instance by differentiating $\lambda\tr(S_\lambda^{-1})$ along the segment) would remove that dependence.
\item \emph{A priori bounds on the local distances.} The certificates of Theorems~\ref{thm:C11} and \ref{thm:C11loc} need $A'$. A bound on $\delta_i(A,A';\kappa)$ from properties of training alone would turn them into predictions (Remark~\ref{rem:oracle}).
\item \emph{Informative certificates on trained dictionaries.} On our trained dictionaries the atoms near a feature move by about $20^\circ$ across seeds and checkpoints, against margins of a few degrees (Findings~\ref{emp:C11loc} and \ref{emp:run}). Whether a certificate based on another invariant (for instance subspaces rather than single atoms) can certify features in this regime is open.
\item \emph{Certified bounds on $D_i$ via $\Dt_i$.} A two-sided certified bound on the discontinuous $D_i$ and on the sum rule over a rotation-aligned ball, using $\Dt_i$ as a stable minorant, was formulated but not pursued here.
\end{enumerate}

\appendix
\newcommand{\lab}[1]{\textsf{[#1]}}

\section{The label system}\label{app:labels}

Every statement carries exactly one marker, a boxed word after its title. The markers map to the project's labels as follows. The check mark $\checkmark$ means that an independent reviewer confirmed the proof written here. The circle $\circ$ means that an independent reviewer checked the proof and raised a point, not yet resolved, on the wording of the statement or an edge case; until it is resolved, the statement may claim slightly more than the proof shows.

\begin{center}
\begin{tabular}{@{}p{0.33\textwidth}p{0.58\textwidth}@{}}
\toprule
Marker & Project label\\
\midrule
\mk{definition (ours)} & \lab{Definition: program}\\
\mk{definition (source \dots)} & a definition taken from the named source; any part marked ``ours'' is \lab{Definition: program}. Elhage et al.\ \cite{elhage2022} is a preprint and lab report; Cohen et al.\ \cite{cohen2017}, El Alaoui and Mahoney \cite{elalaoui2015} and Chen et al.\ \cite{chen2026} are peer-reviewed.\\
\mk{ours $\checkmark$} & \lab{Proof: program, checked} (statement and proof are ours)\\
\mk{source (unreviewed) $\checkmark$} & \lab{Reported} for the statement, which follows from a preprint, with \lab{Proof: program, checked} for our proof\\
\mk{ours $\circ$} & \lab{Proof: program, pending critic} (statement and proof are ours; a reviewer's point is open)\\
\mk{source (unreviewed) $\circ$} & the same for a statement that follows from a preprint: \lab{Reported} for the statement, \lab{Proof: program, pending critic} for our proof\\
\mk{source (unreviewed)} & \lab{Reported}, no proof here\\
\mk{tool} & \lab{tool, not proved here}\\
\mk{conjecture (ours)} & \lab{Conjecture}, ours\\
\mk{speculation} & \lab{Speculation}\\
\mk{measured} & \lab{Measured: program run}\\
\bottomrule
\end{tabular}
\end{center}

\begin{itemize}[itemsep=1pt]
\item \lab{Reported}: stated in unreviewed work (a preprint or a lab report); not checked by peer review.
\item \lab{Proof: program, checked}: proved here, in numbered steps, and confirmed by an independent reviewer. \lab{Proof: program, pending critic}: proved here, in numbered steps, with a point raised by the reviewer still open. Classical facts that we prove for lack of a qualifying source (P1--P4, Lemma~\ref{lem:triangle}) carry one of these labels too.
\item \lab{tool, not proved here}: a standard fact used but not proved here, with a pointer where we have one; ``not verified'' means that the section number was not checked.
\item \lab{Conjecture}: a precise statement we believe true and have not proved.
\item \lab{Measured: program run}: a result of our own unreviewed run of our own code. Each such finding states the quantity measured, the method, the expected value and its source, and PASS/FAIL or INFO.
\item \lab{Speculation}: an informal reading or a research direction, not a precise claim.
\end{itemize}

\noindent Statement types: Definition; Theorem (including Lemma, Proposition, Corollary, Tool, and Remark with a proof); Derivation; Empirical finding; Heuristic; Conjecture. A ``credit'' sentence inside a statement names a source for a mechanism or a related statement; it is not a separate statement.

\section{A second proof of Theorem~\ref{thm:C4b}}\label{app:B4}

This proof is independent of the one in \S\ref{ssec:C4b}. It writes $\ell_i$ through the Gram matrix, turns it into a minimum over test vectors, and needs a single test vector to compare $W$ with $W'$; the comparison reduces to a function of one variable.

\begin{definition}[notation for the second proof]\label{def:B4not}\MkDef
For $\lambda>0$, $\mathcal V_i(W;\lambda):=\min\{\|Wz\|^2+\lambda\|z\|^2:\ z\in\R^n,\ z_i=1\}$ (Step~2 shows that the minimum exists). $K_\lambda=W^\top W+\lambda I_n$ (D1) and $K'_\lambda:=W'^\top W'+\lambda I_n$; $E$ and $\mathsf h$ are as in D1, and $\Fsc$ is as in Definition~\ref{def:scalar}.
\end{definition}

\begin{theorem}[sharp Lipschitz constant of the ridge leverage, second proof]\label{thm:B4}\MkOursChk
For all $d,n\ge1$, $\lambda>0$, $i\in[n]$ and $W,W'\in\R^{d\times n}$,
\begin{equation}\label{eq:B4main}
|\ell_i(W;\lambda)-\ell_i(W';\lambda)|\le\frac{3\sqrt3}{8\sqrt\lambda}\,\|W-W'\|_{\op}.
\end{equation}
No smaller constant works: the least constant $L^\star(\lambda)$ of Theorem~\ref{thm:C4b} equals $3\sqrt3/(8\sqrt\lambda)$.
\end{theorem}

\begin{proof}
Fix $d,n,\lambda,i,W,W'$. Tools used: Tools~\ref{tool:pd}, \ref{tool:cs}, \ref{tool:op}, \ref{tool:calc}.
\begin{steps}
\item \emph{Gram representation.} We claim
\begin{equation}\label{eq:B4gram}
\ell_i(W;\lambda)=1-\lambda\,\big[K_\lambda^{-1}\big]_{ii}.
\end{equation}
Both $S_\lambda=WW^\top+\lambda I_d$ and $K_\lambda=W^\top W+\lambda I_n$ are invertible by Tool~\ref{tool:pd}, since $WW^\top$ and $W^\top W$ are positive semidefinite. Multiplying out, $W^\top S_\lambda=W^\top WW^\top+\lambda W^\top=K_\lambda W^\top$. Multiply this identity on the left by $K_\lambda^{-1}$ and on the right by $S_\lambda^{-1}$: $K_\lambda^{-1}W^\top=W^\top S_\lambda^{-1}$. Hence, using $W_i=W\mathsf e_i$ and \eqref{eq:D4},
\[
\ell_i(W;\lambda)=\mathsf e_i^\top W^\top S_\lambda^{-1}W\mathsf e_i=\mathsf e_i^\top K_\lambda^{-1}W^\top W\mathsf e_i=\mathsf e_i^\top K_\lambda^{-1}(K_\lambda-\lambda I_n)\mathsf e_i=1-\lambda\,\mathsf e_i^\top K_\lambda^{-1}\mathsf e_i ,
\]
which is \eqref{eq:B4gram}.
\item \emph{Variational form; reduction to one test vector.} Write $\mathsf d_i=[K_\lambda^{-1}]_{ii}$; then $\mathsf d_i>0$ by Tool~\ref{tool:pd}. We claim that the minimum defining $\mathcal V_i(W;\lambda)$ exists, equals $1/\mathsf d_i$, and is attained at $z_\star=K_\lambda^{-1}\mathsf e_i/\mathsf d_i$. So
\begin{equation}\label{eq:B4var}
\ell_i(W;\lambda)=1-\frac{\lambda}{\mathcal V_i(W;\lambda)} .
\end{equation}
First, for every $z\in\R^n$, $z^\top K_\lambda z=\|Wz\|^2+\lambda\|z\|^2$. Let $z_i=1$. With the inner product $\langle\cdot,\cdot\rangle_{K_\lambda}$ of Tool~\ref{tool:pd}, $1=\mathsf e_i^\top z=\langle K_\lambda^{-1}\mathsf e_i,z\rangle_{K_\lambda}$, and Tool~\ref{tool:cs} gives $1\le(\mathsf e_i^\top K_\lambda^{-1}K_\lambda K_\lambda^{-1}\mathsf e_i)(z^\top K_\lambda z)=\mathsf d_i\,z^\top K_\lambda z$. So $\|Wz\|^2+\lambda\|z\|^2\ge1/\mathsf d_i$ for every admissible $z$. The vector $z_\star$ is admissible, since $(z_\star)_i=\mathsf e_i^\top K_\lambda^{-1}\mathsf e_i/\mathsf d_i=1$. It attains the bound: $z_\star^\top K_\lambda z_\star=\mathsf e_i^\top K_\lambda^{-1}K_\lambda K_\lambda^{-1}\mathsf e_i/\mathsf d_i^2=1/\mathsf d_i$. Therefore $\mathcal V_i(W;\lambda)=1/\mathsf d_i$, and \eqref{eq:B4var} follows from \eqref{eq:B4gram}.

What this step preserves: $\ell_i(W;\lambda)$ is an increasing function of the single number $\mathcal V_i(W;\lambda)$, and $\mathcal V_i$ is a minimum over test vectors. An upper bound on $\mathcal V_i(W';\lambda)$ therefore needs only one test vector.
\item \emph{The two reduced variables.} Let $z_\star$ be the minimiser of Step~2 for $W$, and put $\mathsf u=\|Wz_\star\|\ge0$ and $\mathsf v=\|z_\star\|$. Since $(z_\star)_i=1$, we have $\mathsf v\ge|(z_\star)_i|=1$. By Step~2,
\begin{equation}\label{eq:B4PsiW}
\mathcal V_i(W;\lambda)=\mathsf u^2+\lambda\mathsf v^2 .
\end{equation}
\item \emph{Upper bound on $\mathcal V_i(W';\lambda)$ in the reduced variables.} The vector $z_\star$ is admissible for $\mathcal V_i(W';\lambda)$. By Tool~\ref{tool:op}, $\|W'z_\star\|\le\|Wz_\star\|+\|Ez_\star\|\le\mathsf u+\mathsf h\mathsf v$. Both sides are $\ge0$, so squaring preserves the inequality. Hence
\begin{equation}\label{eq:B4PsiWp}
0<\mathcal V_i(W';\lambda)\le\|W'z_\star\|^2+\lambda\mathsf v^2\le(\mathsf u+\mathsf h\mathsf v)^2+\lambda\mathsf v^2 .
\end{equation}
(The left inequality holds because $\mathcal V_i(W';\lambda)=1/[K_\lambda'^{-1}]_{ii}>0$ by Step~2 applied to $W'$.)
\item \emph{One-sided bound by a scalar function.} Define $\mathcal L:[0,\mathsf h]\to\R$ by $\mathcal L(u)=-\lambda/\big((\mathsf u+u\mathsf v)^2+\lambda\mathsf v^2\big)$. Its denominator is $\ge\lambda\mathsf v^2\ge\lambda>0$, so $\mathcal L$ is continuously differentiable. By \eqref{eq:B4var} and \eqref{eq:B4PsiW}, $\ell_i(W;\lambda)=1+\mathcal L(0)$. The map $y\mapsto-\lambda/y$ is increasing on $(0,\infty)$. So \eqref{eq:B4var} and \eqref{eq:B4PsiWp} give $\ell_i(W';\lambda)\le1+\mathcal L(\mathsf h)$. Hence
\begin{equation}\label{eq:B4onesided}
\ell_i(W';\lambda)-\ell_i(W;\lambda)\le\mathcal L(\mathsf h)-\mathcal L(0).
\end{equation}
\item \emph{Exact maximisation of $\Fsc$.} For $u\ge0$, $\Fsc'(u)=2(1+u^2)^{-2}-8u^2(1+u^2)^{-3}=2(1-3u^2)/(1+u^2)^3$. So $\Fsc$ is increasing on $[0,1/\sqrt3]$ and decreasing on $[1/\sqrt3,\infty)$, and
\begin{equation}\label{eq:B4Theta}
\max_{u\ge0}\Fsc(u)=\Fsc(1/\sqrt3)=\frac{2/\sqrt3}{(4/3)^2}=\frac{9}{8\sqrt3}=\frac{3\sqrt3}{8}.
\end{equation}
\item \emph{Derivative of $\mathcal L$ in terms of $\Fsc$.} For $u\in[0,\mathsf h]$, differentiating the quotient gives $\mathcal L'(u)=2\lambda\mathsf v(\mathsf u+u\mathsf v)/\big((\mathsf u+u\mathsf v)^2+\lambda\mathsf v^2\big)^2$. Put $y=(\mathsf u+u\mathsf v)/(\sqrt\lambda\,\mathsf v)\ge0$. Then $(\mathsf u+u\mathsf v)^2+\lambda\mathsf v^2=\lambda\mathsf v^2(1+y^2)$ and $2\lambda\mathsf v(\mathsf u+u\mathsf v)=2\lambda^{3/2}\mathsf v^2y$, so
\[
\mathcal L'(u)=\frac{2\lambda^{3/2}\mathsf v^2y}{\lambda^2\mathsf v^4(1+y^2)^2}=\frac{\Fsc(y)}{\sqrt\lambda\,\mathsf v^2}\le\frac{3\sqrt3}{8\sqrt\lambda}.
\]
The inequality uses \eqref{eq:B4Theta} and $\mathsf v\ge1$ (Step~3).
\item \emph{One-sided bound.} If $\mathsf h=0$, then $W=W'$ and there is nothing to prove. If $\mathsf h>0$, then by Tool~\ref{tool:calc}(c) on $[0,\mathsf h]$ and Step~7, $\mathcal L(\mathsf h)-\mathcal L(0)\le\frac{3\sqrt3}{8\sqrt\lambda}\mathsf h$. With \eqref{eq:B4onesided}, $\ell_i(W';\lambda)-\ell_i(W;\lambda)\le\frac{3\sqrt3}{8\sqrt\lambda}\|W-W'\|_{\op}$.
\item \emph{Symmetry.} Steps~2--8 used nothing about $W$ and $W'$ beyond $\|W'-W\|_{\op}=\mathsf h$. Exchanging the roles of $W$ and $W'$ (with the minimiser of Step~2 now taken for $W'$) gives $\ell_i(W;\lambda)-\ell_i(W';\lambda)\le\frac{3\sqrt3}{8\sqrt\lambda}\|W-W'\|_{\op}$. The two one-sided bounds together give \eqref{eq:B4main}.
\item \emph{Sharpness.} Take $d=n=1$ and $W=(w)$. Then $\ell_1(W;\lambda)=w^2/(w^2+\lambda)$, and $\frac{\mathrm d}{\mathrm dw}\frac{w^2}{w^2+\lambda}=\frac{2w(w^2+\lambda)-2w^3}{(w^2+\lambda)^2}=\frac{2\lambda w}{(w^2+\lambda)^2}=\frac{\Fsc(w/\sqrt\lambda)}{\sqrt\lambda}$. At $w=\sqrt{\lambda/3}$ this equals $3\sqrt3/(8\sqrt\lambda)$ by \eqref{eq:B4Theta}. Here $\|W-W'\|_{\op}=|w-w'|$. So the difference quotient $|\ell_1(w)-\ell_1(w')|/|w-w'|$ tends to $3\sqrt3/(8\sqrt\lambda)$ as $w'\to w=\sqrt{\lambda/3}$. Hence for every $L<3\sqrt3/(8\sqrt\lambda)$ there is a pair $W,W'$ that violates the inequality with constant $L$. Together with Step~9, $L^\star(\lambda)=3\sqrt3/(8\sqrt\lambda)$.
\end{steps}
\end{proof}

\begin{remark}[slightly stronger facts]\label{rem:B4}\MkOurs
Let $\mathsf v_W$ and $\mathsf v_{W'}$ be the norms $\mathsf v$ of Step~3 for $W$ and for $W'$ (the minimiser of Step~2 taken for $W$, respectively $W'$). Then:
(1) \eqref{eq:B4main} implies the inequality of Corollary~\ref{cor:C4a}.
(2) $\|z_\star-\mathsf e_i\|^2=\mathsf v^2-1$; so $\mathsf v=1$ iff $z_\star=\mathsf e_i$, iff $\langle W_j,W_i\rangle=0$ for every $j\ne i$.
(3) $\ell_i(W';\lambda)-\ell_i(W;\lambda)\le\frac{3\sqrt3}{8\sqrt\lambda\,\mathsf v_W^2}\|W-W'\|_{\op}$, and $|\ell_i(W;\lambda)-\ell_i(W';\lambda)|\le\frac{3\sqrt3}{8\sqrt\lambda\,\min(\mathsf v_W,\mathsf v_{W'})^2}\|W-W'\|_{\op}$.
(4) $\mathcal L'(0)=\frac{3\sqrt3}{8\sqrt\lambda}$ iff $\langle W_j,W_i\rangle=0$ for every $j\ne i$ and $\|W_i\|^2=\lambda/3$, in any $d$ and $n$ (compare Remark~\ref{rem:A4tight}).
(5) For $W\ne W'$ the inequality \eqref{eq:B4main} is strict. So $3\sqrt3/(8\sqrt\lambda)$ is the supremum of the ratios $|\ell_i(W;\lambda)-\ell_i(W';\lambda)|/\|W-W'\|_{\op}$ (Step~10), approached but attained by no pair.
\end{remark}
\begin{proof}
\begin{steps}
\item (1): $3\sqrt3/8<1$ (Corollary~\ref{cor:C4a}, Step~1).
\item (2): $(z_\star)_i=1$ (Step~2), so $\|z_\star-\mathsf e_i\|^2=\|z_\star\|^2-2(z_\star)_i+1=\mathsf v^2-1$. Hence $\mathsf v=1$ iff $z_\star=\mathsf e_i$.
\item (2), continued: $z_\star=\mathsf e_i$ iff $K_\lambda^{-1}\mathsf e_i=\mathsf d_i\mathsf e_i$ (Step~2). If so, $K_\lambda\mathsf e_i=\mathsf e_i/\mathsf d_i$. Conversely, if $K_\lambda\mathsf e_i=c\,\mathsf e_i$, then $c=\mathsf e_i^\top K_\lambda\mathsf e_i>0$ (Tool~\ref{tool:pd}), $K_\lambda^{-1}\mathsf e_i=\mathsf e_i/c$, $\mathsf d_i=1/c$ and $z_\star=\mathsf e_i$. And $K_\lambda\mathsf e_i$ is a multiple of $\mathsf e_i$ iff $W^\top W\mathsf e_i$ is, iff its entries $\langle W_j,W_i\rangle$, $j\ne i$, vanish.
\item (3): Steps~7--8 with the bound $\mathcal L'(u)=\Fsc(y)/(\sqrt\lambda\,\mathsf v^2)\le\frac{3\sqrt3}{8\sqrt\lambda\,\mathsf v^2}$ (without using $\mathsf v\ge1$) give the first inequality with $\mathsf v=\mathsf v_W$. Exchanging $W$ and $W'$ as in Step~9 gives $\ell_i(W;\lambda)-\ell_i(W';\lambda)\le\frac{3\sqrt3}{8\sqrt\lambda\,\mathsf v_{W'}^2}\|W-W'\|_{\op}$. Both right sides are at most the one with $\min(\mathsf v_W,\mathsf v_{W'})^2$.
\item (4): with $y(u):=(\mathsf u+u\mathsf v)/(\sqrt\lambda\,\mathsf v)$ as in Step~7, $\mathcal L'(0)=\Fsc(y(0))/(\sqrt\lambda\,\mathsf v^2)$ and $y(0)=\mathsf u/(\sqrt\lambda\,\mathsf v)$. As $\Fsc\le3\sqrt3/8$ with equality only at $1/\sqrt3$ (Step~6, with the strict monotonicity of Tool~\ref{tool:calc}(d)) and $\mathsf v\ge1$, equality holds iff $\mathsf v=1$ and $y(0)=1/\sqrt3$. By (2), $\mathsf v=1$ iff $W_i$ is orthogonal to the other columns, and then $z_\star=\mathsf e_i$, $\mathsf u=\|W\mathsf e_i\|=\|W_i\|$, so $y(0)=\|W_i\|/\sqrt\lambda$; this equals $1/\sqrt3$ iff $\|W_i\|^2=\lambda/3$.
\item (5), one side: let $\mathsf h>0$. For $u\in[0,\mathsf h]$, $y(u)=(\mathsf u+u\mathsf v)/(\sqrt\lambda\,\mathsf v)$ is strictly increasing in $u$ ($\mathsf v>0$), so $y(u)=1/\sqrt3$ for at most one $u$. The function $\mathsf f(u):=\frac{3\sqrt3}{8\sqrt\lambda}-\mathcal L'(u)$ is continuous, $\ge0$ (Step~7), and $>0$ wherever $y(u)\ne1/\sqrt3$ (Step~6 and $\mathsf v\ge1$), hence at some point of $[0,\mathsf h]$. By Tool~\ref{tool:calc}(c),(f),(g), $\mathcal L(\mathsf h)-\mathcal L(0)=\frac{3\sqrt3}{8\sqrt\lambda}\mathsf h-\int_0^{\mathsf h}\mathsf f<\frac{3\sqrt3}{8\sqrt\lambda}\mathsf h$. With \eqref{eq:B4onesided}, $\ell_i(W';\lambda)-\ell_i(W;\lambda)<\frac{3\sqrt3}{8\sqrt\lambda}\|W-W'\|_{\op}$.
\item (5), other side and conclusion: exchanging $W$ and $W'$ as in Step~9 gives the strict bound for $\ell_i(W;\lambda)-\ell_i(W';\lambda)$, so \eqref{eq:B4main} is strict when $\mathsf h>0$, i.e.\ $W\ne W'$. The supremum statement is Step~10.
\end{steps}
\end{proof}

\section{A second proof of the coefficient $1/4$}\label{app:B7}

This proof is independent of the one in \S\ref{ssec:C7p}. After scaling to $\lambda=1$, $\Dt_i$ depends on $W$ only through the two numbers $\|W_i\|$ and $\|W_{-i}^\top W_i\|$; the perturbation moves each of them by an explicit amount, and the suprema of the two partial derivatives are computed in closed form. The mean value theorem is applied only along coordinate segments of the reduced plane.

\begin{definition}[notation for the second proof]\label{def:B7not}\MkDef
$J_i\in\R^{n\times(n-1)}$ is the matrix that inserts a zero at position $i$: $J_iy$ has entry $0$ at position $i$ and the entries of $y$, in order, elsewhere. So $W_{-i}=WJ_i$, and $\|J_iy\|=\|y\|$. With $\mathsf a=\|W_i\|$ and $\mathsf a'=\|W'_i\|$ (D1), put $\mathsf x=\|W_{-i}^\top W_i\|=\big(\sum_{j\ne i}\langle W_j,W_i\rangle^2\big)^{1/2}$ and $\mathsf x'=\|W_{-i}'^\top W'_i\|$. Let $\mathcal D:[0,\infty)^2\to[0,\infty)$, $\mathcal D(y_1,y_2)=y_1^4/(y_1^2+y_1^4+y_2^2)$ for $y_1>0$ and $\mathcal D(0,y_2)=0$, and $\mathsf m(y_1)=y_1^2+y_1^4$.
\end{definition}

\begin{theorem}[upper half of C7$'$(i), second proof]\label{thm:B7}\MkOurs
For all $d,n\ge1$, $\lambda>0$, $i\in[n]$ and $W,W'\in\R^{d\times n}$,
\begin{equation}\label{eq:B7main}
|\Dt_i(W;\lambda)-\Dt_i(W';\lambda)|\le\Big(\frac{1/\sqrt2+1/4}{\sqrt\lambda}+\frac{R_{\min}}{4\lambda}\Big)\|W-W'\|_{\op} .
\end{equation}
Since $R_{\min}\le R$, the same bound holds with $R$ in place of $R_{\min}$. Consequently, for every $\rho>0$,
\begin{equation}\label{eq:B7Phi}
\Phi(\rho)\le\frac1{\sqrt2}+\frac14+\frac\rho4,\qquad\text{so}\qquad\limsup_{\rho\to\infty}\frac{\Phi(\rho)}{\rho}\le\frac14 .
\end{equation}
\end{theorem}

\begin{proof}
Tools used: Tools~\ref{tool:op} and \ref{tool:calc}(b).
\begin{steps}
\item \emph{Reduction to $\lambda=1$ by scaling.} Let $c>0$. For $W_i\ne0$, D5 gives
\[
\Dt_i(cW;c^2\lambda)=\frac{c^4\|W_i\|^4}{(cW_i)^\top(c^2S)(cW_i)+c^2\lambda\,c^2\|W_i\|^2}=\frac{c^4\|W_i\|^4}{c^4\big(W_i^\top SW_i+\lambda\|W_i\|^2\big)}=\Dt_i(W;\lambda),
\]
and for $W_i=0$ both sides are $0$. Also $\|cW-cW'\|_{\op}=c\|W-W'\|_{\op}$ and $\|cW\|_{\op}=c\|W\|_{\op}$. Take $c=\lambda^{-1/2}$. Then \eqref{eq:B7main} for $(W,W';\lambda)$ is the same inequality as \eqref{eq:B7main} for $(cW,cW';1)$: the left sides are equal, and the right side for $(cW,cW';1)$ is $\big(1/\sqrt2+1/4+cR_{\min}/4\big)c\mathsf h$, which equals the right side for $(W,W';\lambda)$. So it suffices to prove \eqref{eq:B7main} for $\lambda=1$, which we assume from now on.
\item \emph{Change of representation: $\Dt_i$ depends only on $(\mathsf a,\mathsf x)$.} The entries of $W^\top W_i$ are $\langle W_j,W_i\rangle$, $j\in[n]$, and the $i$-th one is $\|W_i\|^2$. Hence $W_i^\top SW_i=\|W^\top W_i\|^2=\|W_i\|^4+\sum_{j\ne i}\langle W_j,W_i\rangle^2=\mathsf a^4+\mathsf x^2$ (for $n=1$ the sum is empty and $\mathsf x=0$). By D5 with $\lambda=1$ and the definition of $\mathcal D$,
\begin{equation}\label{eq:B7reduced}
\Dt_i(W;1)=\mathcal D(\mathsf a,\mathsf x)\qquad\text{for every }W,
\end{equation}
including $W_i=0$, where both sides are $0$. What is preserved: \eqref{eq:B7reduced} is an identity, so the difference $\Dt_i(W;1)-\Dt_i(W';1)$ is exactly $\mathcal D(\mathsf a,\mathsf x)-\mathcal D(\mathsf a',\mathsf x')$. All dependence on $d$, on $n$ and on the other columns is now carried by the two numbers $\mathsf x,\mathsf x'$.
\item \emph{Continuity of $\mathcal D$ in $y_1$ at $y_1=0$.} For $y_1>0$ and $y_2\ge0$, $0\le\mathcal D(y_1,y_2)\le y_1^4/y_1^2=y_1^2$. Hence for each fixed $y_2\ge0$, the map $y_1\mapsto\mathcal D(y_1,y_2)$ is continuous on $[0,\infty)$; it is also differentiable on $(0,\infty)$, where its denominator is $\ge y_1^2>0$. For fixed $y_1>0$, $y_2\mapsto\mathcal D(y_1,y_2)$ is differentiable on $[0,\infty)$. For $y_1=0$, $y_2\mapsto\mathcal D(0,y_2)$ is identically $0$.
\item \emph{How the reduced variables move.} Here $E_i=E\mathsf e_i=W'_i-W_i$ and $E_{-i}=W'_{-i}-W_{-i}=EJ_i$. By Tool~\ref{tool:op} and $\|J_iy\|=\|y\|$, $\|E_i\|\le\mathsf h$, $\|E_{-i}\|_{\op}\le\mathsf h$, and $\|W_{-i}\|_{\op}\le\|W\|_{\op}$, $\|W'_{-i}\|_{\op}\le\|W'\|_{\op}$. Then:
\begin{enumerate}[label=(\alph*)]
\item $|\mathsf a'-\mathsf a|=|\|W'_i\|-\|W_i\||\le\|E_i\|\le\mathsf h$, by the reverse triangle inequality.
\item $W'^\top_{-i}W'_i-W_{-i}^\top W_i=W_{-i}^\top(W'_i-W_i)+(W'_{-i}-W_{-i})^\top W'_i=W_{-i}^\top E_i+E_{-i}^\top W'_i$. By the reverse triangle inequality and Tool~\ref{tool:op}, $|\mathsf x'-\mathsf x|\le\|W\|_{\op}\mathsf h+\mathsf h\mathsf a'=\mathsf h(\|W\|_{\op}+\mathsf a')$.
\item In the same way, $W'^\top_{-i}W'_i-W_{-i}^\top W_i=W'^\top_{-i}E_i+E_{-i}^\top W_i$, so $|\mathsf x'-\mathsf x|\le\mathsf h(\|W'\|_{\op}+\mathsf a)$.
\end{enumerate}
(For $n=1$, $\mathsf x=\mathsf x'=0$ and (b), (c) hold trivially.)
\item \emph{Partial derivatives of $\mathcal D$.} On $(0,\infty)\times[0,\infty)$, $\mathsf m(y_1)>0$, and the quotient rule gives
\[
\partial_1\mathcal D=\frac{4y_1^3(y_1^2+y_1^4+y_2^2)-y_1^4(2y_1+4y_1^3)}{(y_1^2+y_1^4+y_2^2)^2}=\frac{2y_1^5+4y_1^3y_2^2}{(\mathsf m(y_1)+y_2^2)^2}\ \ge0,
\qquad
\partial_2\mathcal D=-\frac{2y_2y_1^4}{(\mathsf m(y_1)+y_2^2)^2}.
\]
\item \emph{Exact supremum of $\partial_1\mathcal D$.} Fix $y_1>0$, and put $y=y_2^2\ge0$ and $\mathcal F(y)=(2y_1^5+4y_1^3y)/(\mathsf m+y)^2$ with $\mathsf m=\mathsf m(y_1)$. Then
\[
\mathcal F'(y)=\frac{4y_1^3(\mathsf m+y)-2(2y_1^5+4y_1^3y)}{(\mathsf m+y)^3}=\frac{4y_1^3(\mathsf m-y_1^2-y)}{(\mathsf m+y)^3}=\frac{4y_1^3(y_1^4-y)}{(\mathsf m+y)^3}.
\]
So $\mathcal F$ increases on $[0,y_1^4]$ and decreases on $[y_1^4,\infty)$. Since $\mathsf m+y_1^4=y_1^2(1+2y_1^2)$ and $2y_1^5+4y_1^7=2y_1^5(1+2y_1^2)$, $\max_y\mathcal F=\mathcal F(y_1^4)=\frac{2y_1^5(1+2y_1^2)}{y_1^4(1+2y_1^2)^2}=\frac{2y_1}{1+2y_1^2}$. Next, $\frac{\mathrm d}{\mathrm dy_1}\frac{2y_1}{1+2y_1^2}=\frac{2(1-2y_1^2)}{(1+2y_1^2)^2}$, so the maximum over $y_1>0$ is at $y_1=1/\sqrt2$, with value $\sqrt2/2=1/\sqrt2$. Hence
\begin{equation}\label{eq:B7dalpha}
0\le\partial_1\mathcal D(y_1,y_2)\le\frac1{\sqrt2}\qquad(y_1>0,\ y_2\ge0).
\end{equation}
\item \emph{Exact supremum of $|\partial_2\mathcal D|$ over $y_2$, for fixed $y_1$.} Fix $y_1>0$ and $\mathsf m=\mathsf m(y_1)$. For $y_2\ge0$, $\frac{\mathrm d}{\mathrm dy_2}\frac{y_2}{(\mathsf m+y_2^2)^2}=\frac{(\mathsf m+y_2^2)-4y_2^2}{(\mathsf m+y_2^2)^3}=\frac{\mathsf m-3y_2^2}{(\mathsf m+y_2^2)^3}$. So the maximum over $y_2\ge0$ is at $y_2^2=\mathsf m/3$, with value $\sqrt{\mathsf m/3}\,/(4\mathsf m/3)^2=\frac{9}{16\sqrt3}\mathsf m^{-3/2}$. Since $\mathsf m^{3/2}=y_1^3(1+y_1^2)^{3/2}$,
\begin{equation}\label{eq:B7dchi}
|\partial_2\mathcal D(y_1,y_2)|\le\frac{9}{8\sqrt3}\cdot\frac{y_1^4}{\mathsf m^{3/2}}=\frac{3\sqrt3}{8}\cdot\frac{y_1}{(1+y_1^2)^{3/2}}\qquad(y_2\ge0).
\end{equation}
\item \emph{Two one-variable maximisations.} For $y_1\ge0$:
\begin{itemize}
\item $\frac{\mathrm d}{\mathrm dy_1}\frac{y_1}{(1+y_1^2)^{3/2}}=\frac{1-2y_1^2}{(1+y_1^2)^{5/2}}$. So the maximum is at $y_1^2=1/2$, with value $\frac{1/\sqrt2}{(3/2)^{3/2}}=\frac{2}{3\sqrt3}$.
\item $\frac{\mathrm d}{\mathrm dy_1}\frac{y_1^2}{(1+y_1^2)^{3/2}}=\frac{y_1(2-y_1^2)}{(1+y_1^2)^{5/2}}$. So the maximum is at $y_1^2=2$, with value $\frac{2}{3^{3/2}}=\frac{2}{3\sqrt3}$.
\end{itemize}
Hence, for every $\mathsf r_0\ge0$ and $y_1\ge0$,
\begin{equation}\label{eq:B7comb}
\frac{3\sqrt3}{8}\cdot\frac{y_1(\mathsf r_0+y_1)}{(1+y_1^2)^{3/2}}\le\frac{3\sqrt3}{8}\Big(\mathsf r_0\cdot\frac{2}{3\sqrt3}+\frac{2}{3\sqrt3}\Big)=\frac{\mathsf r_0+1}{4}.
\end{equation}
\item \emph{Two-leg path in the reduced plane, first version.} Go from $(\mathsf a,\mathsf x)$ to $(\mathsf a',\mathsf x)$, then from $(\mathsf a',\mathsf x)$ to $(\mathsf a',\mathsf x')$.
\begin{itemize}
\item First leg: by Step~3, $\mathcal D(\cdot,\mathsf x)$ is continuous on the closed interval between $\mathsf a$ and $\mathsf a'$ and differentiable inside it, since the interior lies in $(0,\infty)$. By Tool~\ref{tool:calc}(b), \eqref{eq:B7dalpha} and Step~4(a), $|\mathcal D(\mathsf a,\mathsf x)-\mathcal D(\mathsf a',\mathsf x)|\le\mathsf h/\sqrt2$.
\item Second leg: if $\mathsf a'=0$, it contributes $0$ by Step~3. If $\mathsf a'>0$, then by Tool~\ref{tool:calc}(b), \eqref{eq:B7dchi}, Step~4(b) and \eqref{eq:B7comb} with $\mathsf r_0=\|W\|_{\op}$,
\[
|\mathcal D(\mathsf a',\mathsf x)-\mathcal D(\mathsf a',\mathsf x')|\le\frac{3\sqrt3}{8}\frac{\mathsf a'}{(1+\mathsf a'^2)^{3/2}}\,\mathsf h\big(\|W\|_{\op}+\mathsf a'\big)\le\frac{\|W\|_{\op}+1}{4}\,\mathsf h .
\]
\end{itemize}
By the triangle inequality and \eqref{eq:B7reduced}, $|\Dt_i(W;1)-\Dt_i(W';1)|\le\big(\tfrac1{\sqrt2}+\tfrac14+\tfrac{\|W\|_{\op}}4\big)\mathsf h$.
\item \emph{Second version, and conclusion for $\lambda=1$.} Go instead from $(\mathsf a,\mathsf x)$ to $(\mathsf a,\mathsf x')$, then to $(\mathsf a',\mathsf x')$. The argument of Step~9, with Step~4(c) in place of Step~4(b) and $\mathsf a$ in place of $\mathsf a'$ on the second-coordinate leg, gives (explicitly: the leg from $(\mathsf a,\mathsf x)$ to $(\mathsf a,\mathsf x')$ is $0$ if $\mathsf a=0$ by Step~3, and otherwise is at most $\frac{3\sqrt3}{8}\frac{\mathsf a}{(1+\mathsf a^2)^{3/2}}\mathsf h(\|W'\|_{\op}+\mathsf a)\le\frac{\|W'\|_{\op}+1}4\mathsf h$ by \eqref{eq:B7dchi}, Step~4(c) and \eqref{eq:B7comb} with $\mathsf r_0=\|W'\|_{\op}$; the leg from $(\mathsf a,\mathsf x')$ to $(\mathsf a',\mathsf x')$ is at most $\mathsf h/\sqrt2$ by \eqref{eq:B7dalpha} and Step~4(a)) $|\Dt_i(W;1)-\Dt_i(W';1)|\le\big(\tfrac1{\sqrt2}+\tfrac14+\tfrac{\|W'\|_{\op}}4\big)\mathsf h$. Taking the smaller of the two bounds gives \eqref{eq:B7main} for $\lambda=1$, and by Step~1 for every $\lambda>0$.
\item \emph{Consequence for $\Phi$.} Let $W\ne W'$ with $R\le\rho\sqrt\lambda$. Multiply \eqref{eq:B7main} by $\sqrt\lambda/\|W-W'\|_{\op}$ and use $R_{\min}\le R\le\rho\sqrt\lambda$: $\sqrt\lambda\,|\Dt_i(W;\lambda)-\Dt_i(W';\lambda)|/\|W-W'\|_{\op}\le\frac1{\sqrt2}+\frac14+\frac{R_{\min}}{4\sqrt\lambda}\le\frac1{\sqrt2}+\frac14+\frac\rho4$. Taking the supremum gives the first part of \eqref{eq:B7Phi}. Dividing by $\rho$ and letting $\rho\to\infty$ gives $\limsup_{\rho\to\infty}\Phi(\rho)/\rho\le\frac14$.
\end{steps}
\end{proof}

\begin{remark}[where the coefficient comes from]\label{rem:B7}\MkSpec
(i) Only the second-coordinate leg involves $\|W\|_{\op}$. Its coefficient $1/4$ is the exact supremum over $(y_1,y_2)$ of $|\partial_2\mathcal D|$ times the factor $\|W\|_{\op}$ of Step~4(b). It is approached at $y_1^2=1/2$ and $y_2^2=\mathsf m(y_1)/3=1/4$ (that is, $\|W_i\|^2=\lambda/2$ and $W_i^\top S_{-i}W_i=\lambda^2/4$ in the original scale), provided the perturbation of $W_i$ is aligned with a top right singular direction of $W_{-i}^\top$ and $W_{-i}^\top E_i$ is parallel to $W_{-i}^\top W_i$. We do not prove here that these alignments can be realised; the lower bound is Proposition~\ref{prop:profile}(ii). (ii) The constant $1/\sqrt2+1/4\approx0.957$ comes from bounding the two legs separately, each by its own supremum; we do not claim that it is optimal. (iii) \eqref{eq:B7main} improves the constants of Corollary~\ref{cor:A77}(c), $2/\sqrt\lambda$ and $R/\lambda$, to $(1/\sqrt2+1/4)/\sqrt\lambda$ and $R_{\min}/(4\lambda)$.
\end{remark}

\section{Glossary of symbols}\label{app:glossary}

\begin{longtable}{@{}p{0.27\textwidth}p{0.53\textwidth}p{0.14\textwidth}@{}}
\toprule
Symbol & Meaning & Where\\
\midrule
\endhead
$d,n,M,M'$; $[n]$ & dimensions, numbers of features and atoms; $\{1,\dots,n\}$ & D0, D7\\
$\langle\cdot,\cdot\rangle$, $\|\cdot\|$, $\hat x$, $\cos(x,y)$ & inner product, norm, normalisation, cosine & D0\\
$\|\cdot\|_{\op},\|\cdot\|_F,\|\cdot\|_*$; $\sigma_k(B)$ & operator, Frobenius, nuclear norm; singular values & D0\\
$\nu_{\min},\nu_{\max},\nu_k$ & smallest, largest, $k$-th eigenvalue of a symmetric matrix & D0\\
$I_p$, $O(d)$, $\#$, $\one[\cdot]$, $\mathsf e_j$ & identity, orthogonal group, cardinality, indicator, standard basis vector & D0\\
$u,x,y,z$; $B,B'$; $\mathsf f$; $\vartheta$ & dummy variables, matrices, functions, angles (local) & \S\ref{sec:notation}\\
$W,W_i,W_{-i}$; $W'$ & weight matrix, its columns, the other columns; a second matrix & D1\\
$S,S_{-i},S',r,\sigma_k,\omega(W)$ & frame operator, without column $i$, of $W'$; rank; singular values; mean eigenvalue & D1\\
$S_\lambda,S_{-i,\lambda},K_\lambda,K'_\lambda$ & $S+\lambda I_d$, $S_{-i}+\lambda I_d$, $W^\top W+\lambda I_n$, $W'^\top W'+\lambda I_n$ & D1, Def.~\ref{def:B4not}\\
$E,E_i,E_{-i},\mathsf h$ & $W'-W$, its columns, its other columns; $\|W'-W\|_{\op}$ & D1, Def.~\ref{def:C7pnot}\\
$W(\eta)$, $\eta$ & the segment $W+\eta E$ and its parameter & D1\\
$R,R_{\min},\Rb$; $\Rb_0$ & $\max$, $\min$ of $\|W\|_{\op},\|W'\|_{\op}$; any $\Rb\ge R$, $\Rb>0$; the value $(R_0^2+1/2)^{1/2}$ in Remark~\ref{rem:A7loss} & D1, Rem.~\ref{rem:A7loss}\\
$\mathsf a,\mathsf a',\mathsf b,\mathsf b',\mathsf q_i,\mathsf q_i'$ & $\|W_i\|$, $\|W'_i\|$, $\|W^\top\hat W_i\|$, $\|W'^\top\hat W'_i\|$, $W_i^\top SW_i$, $W_i'^\top S'W'_i$ & D1\\
$D_i,\Dtot$ & feature dimensionality and its total & D2\\
$\tau$, $D_i^{(\tau)},\Dtot^{(\tau)}$ & norm threshold; thresholded dimensionality & D3\\
$\lambda$; $\ell_i,\deff$ & ridge parameter; ridge leverage, effective dimension & D4\\
$\Dt_i,\Dttot$ & regularised dimensionality and its total & D5\\
$\Lambda(W)$ & ridge range $[10^{-3},10^{-1}]\,\omega(W)$ & D6\\
$V,v_i,\mu_{ij},\theta_V,\mu(V),\talign(V)$ & ground truth, features, cosines, line separation, signed coherence, alignment threshold & D7\\
$\cA_M$, $A,a_m$, $A',a'_{m'}$ & dictionaries, atoms & D7\\
$c_i,g_m,\bar c$ & per-feature score, per-atom score, area under the FRR curve & D7\\
$t$; $\FRR_t$; $\FRR_{t,\beta},\psi_{t,\beta},\beta$ & cosine threshold; recovery rate; smoothed rate, ramp, bandwidth & D7\\
$e_m(A\to A')$, $s$, $H(A,A')$ & one-sided line distance, sign, line-Hausdorff distance & D7\\
$\gamma$, $N_\gamma(A;t)$ & margin; margin count & D7\\
$Z_i$; $\nu_0$, $\varepsilon_0$ & discontinuity set of $D_i$; local quantities in its proof & Thm~\ref{thm:C1}\\
$L_a,L_c$, $L_c^+$, $L(\lambda)$, $L$ & constants of hypothetical bounds, $\max(L_c,0)$; a Lipschitz constant & Thm~\ref{thm:C1}, \S\ref{ssec:profile}\\
$\mathcal N$, $L_c^{\mathrm{best}}$ & class of pairs and best coefficient in Remark~\ref{rem:C1c} & Rem.~\ref{rem:C1c}\\
$\alpha_0,R_0,\phi,\nphi,\nphip$ & parameters and unit vectors of the $2\times2$ family & Def.~\ref{def:family}\\
$W^\star$, $E^{(\iota)}$, $\iota$, \textsf{F1}, \textsf{F2}; $W^\star(\eta)$ & base point, directions, their index, the two named directions; the curve $W^\star+\eta E^{(\iota)}$ & Def.~\ref{def:family}\\
$\pi_{12}$, $\mathcal C_\lambda$, $\Numfam,\Denfam$ & inner product, curve function, numerator and denominator of the family & Def.~\ref{def:family}\\
$\lsc,w,w',w_0,\Lsc,\Fsc$ & scalar leverage, its arguments, maximiser, maximal slope; $\Fsc(u)=2u/(1+u^2)^2$ & Def.~\ref{def:scalar}\\
$L^\star(\lambda)$ & least Lipschitz constant of $\ell_i$ & Thm~\ref{thm:C4b}\\
$\Theta,\Gamma_1,\Gamma_2,\mathcal T,\mathcal P$ & $\lambda K_\lambda^{-1}$ and quantities in the proof of Thm~\ref{thm:C4b} & \S\ref{ssec:C4b}\\
$\ell^0_j(B),D_j(B),r_B$ & unregularised leverage, dimensionality, rank, for a matrix $B$ & Prop.~\ref{prop:I8}\\
$\Waug$; $\nu$ & $[W,\sqrt\lambda I_d]$; an eigenvalue & \S\ref{ssec:C6}\\
$\rho$, $\Phi,\Phi_{\lambda=1},\chi,\chi^\star$; $c$; admissible & $R/\sqrt\lambda$; Lipschitz profile of $\Dt_i$ and its excess over $\rho/4$; a scaling constant; a tuple in the supremum & Def.~\ref{def:profile}\\
$\xi,\zeta,\qrad,\qtan,\qcpl,E_i^\perp,\Pcpl$ & reduced variables and parts of the perturbation & Def.~\ref{def:C7pnot}\\
$\arad,\atan,\acpl,\Psi_{\Rb},\alpha_\star,\upsilon,\upsilon_\star,\Omega,\varpi$ & coefficients and auxiliary functions for C7$'$ & Def.~\ref{def:C7pnot}\\
$\mathfrak a,\mathfrak b,\mathfrak h$; $W^{\mathrm s}$ & constants and function of Lemma~\ref{lem:A71}; scaled matrix & Lemma~\ref{lem:A71}\\
$\Numseg,\Denseg$, $\mathcal Z$, $\eta_\circ$ & numerator, denominator, zero set and zero along the segment & Lemmas~\ref{lem:A72}, \ref{lem:A73}\\
$\mathcal Q,y_{\mathrm v},\mathcal K$, $\Xi_{\Rb}$ & auxiliary functions in the proofs of C7$'$ & \S\ref{ssec:C7p}\\
$\chi_s,\chi^\star_s,\rho_1$; $\rho_2,\rho_c$; $y_0$ & scalar excess, its maximum, the small-radius bound; the large-radius thresholds; the root of the cubic & Def.~\ref{def:chis}, Der.~\ref{der:chis}\\
$\mathcal B,\mathcal Y$ & auxiliary functions in the proof of Thm~\ref{thm:A83} & Thm~\ref{thm:A83}\\
$\angle(x,y)$, $\ch$ & line angle, chord function & Def.~\ref{def:T4not}\\
$e^\angle_m,\angle_i,P_i,\kappa$ & angular distance, angular score, near set, cutoff & Def.~\ref{def:T4not}\\
$\delta_i,\delta^\angle_i$ & one-sided and symmetric local distances & Def.~\ref{def:T4not}\\
$\Rec_t$; maximiser $m_i$, $m'_i$ & recovered set; best atoms & Def.~\ref{def:T4not}, Thm~\ref{thm:C11}\\
$X_\gamma,Y_\gamma,N^\pm_\gamma,\Lost,\Gain$ & counts and sets for C11 & Def.~\ref{def:C11counts}\\
$N^\angle_t,N^{\mathrm{loc}}_t$, $\Delta$ & uncertified counts; a priori distance bound & Thm~\ref{thm:C11loc}\\
$\varsigma,\tilde A,\mathsf k,\mathsf k',\mathsf k_m$ & permutation, transformed dictionary, nonzero scalars & Lemma~\ref{lem:T41}, Thm~\ref{thm:C8}\\
$\vartheta_0$ & angle in the example of Remark~\ref{rem:chen} & Rem.~\ref{rem:chen}\\
$\mathcal I_1,\mathcal I_2,m^\dagger$ & index sets and witness map & Thm~\ref{thm:C11loc}\\
$\mathcal E,\mathcal G,\mathcal S_{\mathrm{syn}}$ & toy-model sweep, dictionary runs, synthetic dictionaries & D9\\
$p_{\mathrm f},\mathbf h,\mathbf b$ & feature probability, input, bias of the toy model & D9\\
$\mathbf x,\mathbf y,\mathbf x^{\mathrm{rec}},\mathrm{pre}_m,b^{\mathrm{enc}}_m,\mathfrak s$ & data point, its unnormalised form, reconstruction, pre-activation, bias, training seed & D9\\
$\epsilon_{\mathrm c},\epsilon_{\mathrm{nl}}$ & copy noise scale, noise level of the synthetic variants & D9\\
$\mathcal V_i,\mathsf d_i,z_\star,\mathsf u,\mathsf v,\mathcal L$; $\mathsf v_W,\mathsf v_{W'}$ & quantities in the second proof of Thm~\ref{thm:C4b}; $\mathsf v$ for $W$, for $W'$ & App.~\ref{app:B4}\\
$J_i,\mathsf x,\mathsf x',\mathcal D,\mathsf m,\mathcal F$ & quantities in the second proof of the coefficient $1/4$ & App.~\ref{app:B7}\\
\bottomrule
\end{longtable}

\noindent Local sans-serif abbreviations defined inside a single proof and not listed above: $\mathsf c'$ (Theorem~\ref{thm:C6bd}), $\mathsf w,\mathsf c_0$ (Lemma~\ref{lem:A74}), $\mathsf r_1,\mathsf r_2$ (Theorem~\ref{thm:A78}), $\mathsf p,\mathsf v_\zeta,\mathsf c_1$ (Theorem~\ref{thm:A83}), $\mathsf r_0$ (Theorem~\ref{thm:B7}). In Remark~\ref{rem:fadnavis}, $\varepsilon$ and $\sigma_p$ are the source's perturbation size and smallest singular value; they are used nowhere else.

\bibliographystyle{plain}
\bibliography{poc3}

\end{document}